\documentclass[5p,times]{elsarticle}

\usepackage{titlesec}
\usepackage[table,xcdraw,dvipsnames,svgnames,x11names]{xcolor}
\usepackage{amsmath,amsfonts}
\usepackage{algorithm}
\usepackage{algpseudocode}
\usepackage{array}
\usepackage[caption=false,font=normalsize,labelfont=sf,textfont=sf]{subfig}
\usepackage{textcomp}
\usepackage{stfloats}
\usepackage{url}
\usepackage{verbatim}
\usepackage{graphicx}
\usepackage{lineno,hyperref}
\usepackage{multirow}
\usepackage{makecell}
\usepackage{bbding}
\usepackage{amsthm}
\usepackage{tabularx}
\usepackage{rotating}
\usepackage{natbib}
\usepackage{microtype}
\usepackage{booktabs}
\usepackage{soul,xcolor}
\setul{0.5ex}{0.3ex} 
\definecolor{redline}{rgb}{1,0,0}

\usepackage[most]{tcolorbox}

\usepackage[commandnameprefix=always]{changes}
\definechangesauthor[name={Reviewer 1}, color=green]{R1}
\definechangesauthor[name={Reviewer 2}, color=red]{R2}
\definechangesauthor[name={Reviewer 3}, color=orange]{R3}

\titleclass{\subsubsubsection}{straight}[\subsection]
\newcounter{subsubsubsection}[subsubsection]
\renewcommand\thesubsubsubsection{\thesubsubsection.\arabic{subsubsubsection}}

\titleformat{\subsubsubsection}
  {\normalfont\normalsize\itshape}{\thesubsubsubsection}{1em}{}
\titlespacing*{\subsubsubsection}
{0pt}{0ex plus 0ex minus 0ex}{0ex plus 0ex}

\makeatletter
\def\toclevel@subsubsubsection{4}
\def\l@subsubsubsection{\@dottedtocline{4}{7em}{4em}}
\makeatother

\setcitestyle{numbers,square}
\newtheorem{defi}{Definition}

\journal{Knowledge Based System}

\begin{document}

\begin{frontmatter}


\title{A Comprehensive Survey on Automatic Text Summarization with Exploration of LLM-Based Methods}

\author[inst,email]{Yang Zhang}
\author[inst,email]{Hanlei Jin}
\author[inst,email]{Dan Meng}
\author[inst,email]{Jun Wang\thanks{*Corresponding author}}
\author[inst,email]{Jinghua Tan}

\affiliation[inst]{%
  organization={Southwestern University of Finance and Economics},
  addressline={Chengdu},
  country={China}
}

\affiliation[email]{%
  organization={Email Addresses},
  addressline={wangjun1987@swufe.edu.cn, 595915575@qq.com},
  country={}
}




\begin{abstract}
    The exponential growth of textual content on the internet, alongside vast archives of news articles, scientific papers, legal documents, and other domains, has made Automatic Text Summarization (ATS) increasingly important. ATS aims to create concise and accurate summaries, significantly reducing the effort required to process large volumes of text. Originating in the 1950s, ATS has evolved through several technical shifts, moving from statistical models to machine learning and deep learning approaches, and more recently to pre-trained models. Previous surveys have focused on conventional ATS methods, which are often constrained by the predefined generative paradigms. However, the advent of Large Language Models (LLMs) has introduced a paradigm-flexible approach to summarization. With their superior generative capabilities, in-context learning, and few-shot learning abilities, LLMs have demonstrated remarkable improvements in coherence, fluency, and overall summarization quality. In this survey, we provide a comprehensive review of both conventional ATS approaches and the latest advancements in LLM-based methods. Our contributions include: (1) offering an up-to-date survey of ATS; (2) reviewing the latest LLM-based summarization methods.
\end{abstract}



\begin{keyword}

Automatic Summarization, Large Language Models, Natural Language Processing



\end{keyword}

\end{frontmatter}



\section{Introduction}
The exponential growth of the World Wide Web has led to an overwhelming surge in textual data across various domains, including news articles, websites, user reviews, blogs, and social media. Additionally, vast text archives exist in specialized fields such as books, scholarly papers, legal documents, and biomedical records. This rapid expansion of data has far outpaced individuals' ability to search, read, and process all relevant information. To address this challenge, Automatic Text Summarization (ATS) has emerged, leveraging methods from Natural Language Processing (NLP) and Information Retrieval (IR) alongside advanced techniques such as Machine Learning (ML), Deep Learning (DL), and Large Language Models (LLMs). ATS enables users to grasp key ideas efficiently, significantly reducing the time and effort required for document reading and comprehension.

Historically, ATS has undergone several distinct eras of development. In the pre‑2000s, early approaches were primarily statistical, leveraging surface‑level features such as word frequency and sentence position to generate summaries \cite{SPARCKJONES1972,https://doi.org/10.1002/(SICI)1097-4571(199009)41:6<391::AID-ASI1>3.0.CO;2-9,Robertson1994OkapiAT}. From the 2000s to the 2010s, machine learning and shallow neural network–based methods emerged, automating the feature extraction process but still heavily relying on domain‑specific, handcrafted features \cite{1647766,10.1145/1014052.1014073,10.5555/1873781.1873855,10.1007/11775300_35,10.1145/1007568.1007621}. The introduction of deep learning in the 2010s—particularly encoder‑decoder architectures like RNNs \cite{chopra2016} and LSTMs \cite{Hanunggul2019TheIO}, enabled more nuanced, context‑aware summarization by modeling sequences of words and capturing richer semantic information. During this period, two main paradigms dominated: extractive summarization, which focused on identifying key sentences or phrases from the original text \cite{nazari2018,suleiman2020}, and abstractive summarization, which generated new sentences to paraphrase the main ideas \cite{gupta2019,lin2019}. However, these approaches were constrained by their inflexible generative paradigms, which were determined by the specific model architecture and training data. Moreover, abstractive summarization often faced challenges in maintaining content coherence and ensuring factual accuracy.

The advent of Large Language Models (LLMs), such as GPT‑3 and ChatGPT, represents a significant breakthrough in ATS by offering a paradigm‑flexible approach. Unlike earlier models restricted to extractive or abstractive frameworks, LLMs bring unprecedented flexibility by seamlessly integrating both paradigms. With their in‑context and few‑shot learning capabilities, LLMs can handle complex summarization tasks with minimal supervision \cite{liu2023,narayan2021}. This flexibility enables them to switch between extractive, abstractive, and hybrid summarization techniques without retraining or significant architectural changes. In addition to this generative flexibility, LLMs excel at producing coherent, semantically rich summaries by leveraging pre‑trained knowledge from vast text corpora. These advancements have not only improved summarization quality but also set new benchmarks for the field.

Despite these innovations, several surveys on ATS remain focused on conventional extractive and abstractive methods \cite{nazari2018,suleiman2020,gupta2019,lin2019}. These approaches are constrained by predefined generative paradigms and fixed architectures. In contrast, this survey introduces a new class of LLM‑based methods that offer more adaptive and flexible generative paradigms.


In this study our objective is to provide a comprehensive survey of ATS techniques, with particular focus on LLM-based methods. We have developed an automated retrieval algorithm that combines keyword-based searches with LLM-based prompting to efficiently collect and organize research papers on ATS topics. This algorithm streamlines the paper collection process and can be adapted for use in other fields. The source code for the algorithm is available on GitHub.\footnote[1]{https://github.com/JinHanLei/RetrievePapers} To the best of our knowledge, this work is among the first to review and analyze LLM-based ATS techniques in depth. We aim to make the following key contributions:

\begin{itemize}
    \item \textbf{An Up‑to‑Date Survey of ATS:} We provide a comprehensive review of the latest advancements in both conventional ATS approaches (extractive and abstractive summarization) and LLM-based methods that offer paradigm flexibility.
    \item \textbf{A Detailed Review of LLM‑Based Methods:} We present an in‑depth analysis of recent LLM-based summarization techniques, focusing on how in‑context learning, prompt engineering, and few‑shot learning have reshaped ATS.
\end{itemize}

The remainder of this paper is organized as follows: Section \ref{sec:background} provides background on Automatic Text Summarization (ATS), including its categorization and prior surveys. Section \ref{sec:Definition} defines key concepts in ATS and outlines the paper collection methodology. Section \ref{data} details the data acquisition methods used in the ATS pipeline. Conventional ATS approaches are discussed in Section \ref{sec:conventional_ats}. Section \ref{sec:llm_ats} explores the recent LLM-based summarization approaches. Section \ref{sec:evaluation} describes evaluation metrics for summarization, and Section \ref{sec:applications} highlights ATS-based applications. Finally, Section \ref{sec:future_direction} outlines future directions in ATS.

\section{Background of Automatic Text Summarization}
\label{sec:background}
\subsection{History of Automatic Text Summarization}

The field of Automatic Text Summarization (ATS) has evolved significantly over the past several decades. We can structure its development into distinct eras, each marked by advancements in technology and methods.

\textbf{Early Statistical Era (Pre‑2000s).} During this period, summarization methods were predominantly based on statistical text features such as word counts and term frequency measures. One of the foundational approaches was introduced by Hans Peter Luhn in 1958 \cite{luhn1958automatic}. He utilized hand‑crafted features like cue word lists, title words, and term frequencies to calculate a ``significance'' score for each word. Sentences containing the highest‑scoring words were then extracted to form an abstract, representing one of the earliest extractive summarization models.

Building upon Luhn's methodology, later research during this era incorporated more advanced statistical models such as TF‑IDF (Term Frequency‑Inverse Document Frequency) \cite{SPARCKJONES1972}, Latent Semantic Analysis (LSA) \cite{https://doi.org/10.1002/(SICI)1097-4571(199009)41:6<391::AID-ASI1>3.0.CO;2-9}, and BM25 (Best Matching 25) \cite{Robertson1994OkapiAT}. Researchers constructed sentence‑relationship maps to compute sentence similarity scores and then applied path traversal methods to select key sentences, ensuring the overall quality of generated summaries \cite{10.5555/1102022,SALTON1988513,SALTON1995483,doi:10.1126/science.264.5164.1421,SALTON1997193,10.3115/1119089.1119121,mitra-etal-1997-automatic}. However, these approaches struggled to capture deeper context, resulting in summaries that often lacked coherence for semantically rich texts \cite{10.1145/321510.321519}.

In contrast to purely statistical methods, another line of research during this era focused on structure‑based approaches for abstractive summarization. These techniques used templates, rules, trees, and ontologies to identify and reformulate the most important information. For example, \cite{gerani-etal-2014-abstractive} converted texts into discourse trees, selected the most significant nodes, and applied hand‑crafted templates to generate abstractive summaries. Similarly, ontology‑based methods treated summarization as extracting key entities and relationships. Researchers such as Li \cite{li-2015-abstractive} applied ontologies to extract essential phrases, which were then combined with template‑based techniques to produce coherent abstracts. These structure‑based approaches, however, depended heavily on hand‑crafted features and were limited in capturing deep semantics, constraining their ability to generate accurate and coherent summaries \cite{10.1145/321510.321519}.

\textbf{Machine Learning and Early Neural Networks Era (2000s–2010s).} Machine learning–based approaches learned from data \cite{Gambhir2017}, alleviating the need for manually defined rules used in statistical methods. These methods could be supervised, unsupervised, or semi‑supervised. In supervised learning, algorithms such as SVM (Support Vector Machines) \cite{1647766,10.5555/1613715.1613813,Fattah2014}, Naïve Bayes Classification \cite{10.1145/1014052.1014073,10.1162/089120102762671936,10.3115/1117575.1117580}, and CRF (Conditional Random Fields) \cite{10.5555/1873781.1873855,10.5555/1625275.1625736,ALIGULIYEV20097764} were trained on labeled data to classify sentences as ``summary'' or ``non‑summary.'' On the other hand, unsupervised techniques used clustering algorithms such as K‑Means \cite{10.1145/1526709.1526728,10.1007/3-540-47922-8_22}, Hierarchical Clustering \cite{10.1007/11775300_35,10.1145/900051.900097,10.1007/978-3-642-03697-2_33}, and DBSCAN (Density‑Based Spatial Clustering of Applications with Noise) \cite{10.1145/1007568.1007621,10.5555/1557769.1557823}. Semi‑supervised methods bridged the gap by requiring less labeled data to train an effective classifier. Nevertheless, these methods still relied heavily on feature engineering, and their performance could be highly dependent on feature selection, limiting their generalizability and the quality of generated summaries.

\textbf{Deep Learning and Transformer Era (2010s–2020s).} With the success of deep learning techniques in Natural Language Processing (NLP), various neural networks achieved remarkable results in tasks such as machine translation, voice recognition, and dialogue systems by offering a deeper understanding of language semantics. One of the most important categories of neural networks for summarization is Sequence‑to‑Sequence (Seq2Seq) models, which processed input text as sequential tokens, extracted semantic information through recurrent hidden neurons, and recursively generated sequential tokens as summaries. Examples include Recurrent Neural Networks (RNNs) \cite{nallapati-etal-2016-abstractive,nallapati-etal-2016-sequence,ma-etal-2017-improving}, Long Short‑Term Memory (LSTM) \cite{Song2019,cheng-lapata-2016-neural,10.1007/978-3-319-46478-7_47}, and Gated Recurrent Units (GRU) \cite{zhou-etal-2018-neural-document,Cao_Wei_Li_Li_2018}. Despite their improvements, RNN‑based models faced challenges in maintaining coherence over long sequences due to difficulty in capturing long‑range dependencies, which limited their ability to produce high‑quality summaries for longer documents.

With the introduction of the Transformer neural network, the NLP landscape shifted from a ``task‑specific'' paradigm to a ``pre‑trained'' paradigm. Unlike previous models that required specific design adjustments for different NLP tasks, transformer based language models are pre‑trained on extensive text datasets and then fine‑tuned for various objectives. At the architectural level, transformers feature sophisticated layers such as multi‑head attention and multiple neural layers, enabling them to capture deeper semantic meanings in texts. Summarization tasks are no exception: transformer‑based neural models have become the backbone for current methods, including BERT \cite{liu-lapata-2019-text,huang-etal-2021-efficient,Zhong_Liu_Xu_Zhu_Zeng_2022,cachola-etal-2020-tldr,liu2019finetunebertextractivesummarization,zhang-etal-2019-hibert}, GPT \cite{10.5555/3524938.3525989,yang-etal-2020-ted,yu-etal-2021-adaptsum}, Sentence‑BERT \cite{zhong-etal-2020-extractive,zhang-etal-2021-leveraging-pretrained}, RoBERTa \cite{10.1162/tacl_a_00362,10.1016/j.eswa.2021.116292,hartl-kruschwitz-2022-applying}, and XLNet \cite{xiao-carenini-2019-extractive,10.1145/3419106,10.1145/3404835.3462938}. However, the major limitations of these pre‑trained methods are: (1) they require task‑specific fine‑tuning, which limits scalability across datasets; and (2) they generally lack flexibility in generation.

\textbf{Large Language Model (LLM) Era (2020s–Present).}  
The emergence of Large Language Models (LLMs), such as GPT-3, GPT-3.5, and LLaMA, marks a significant milestone in the evolution of Automatic Text Summarization (ATS). These models, pre-trained on vast and diverse corpora, possess the ability to perform a wide range of tasks, including summarization. LLMs excel in \textbf{few-shot} and \textbf{zero-shot learning}, enabling them to generate high-quality summaries with minimal or no task-specific examples. This represents a major advancement over previous models, which often required extensive labeled data for training.

Unlike earlier approaches constrained to a single summarization paradigm—either \textbf{extractive} or \textbf{abstractive}, LLMs offer unprecedented flexibility in their generative capabilities. Thanks to their robust text generation abilities, LLMs can seamlessly adapt between extractive and abstractive paradigms or even blend both approaches within a single task. Moreover, LLM-generated summaries tend to exhibit superior coherence, fluency, and overall writing quality compared to traditional methods.

LLM-based summarization methods can be broadly categorized into four key approaches:

\begin{enumerate}
    \item \textbf{Prompt Engineering–Based Methods}  
    These methods involve designing effective prompts or templates to guide the LLMs in generating high-quality summaries without modifying the model's internal parameters.
    \item \textbf{Retrieval-Augmented Generation–Based Methods}  
    These methods enhance the summarization process by incorporating externally retrieved documents to supplement the model's internal knowledge.    
    \item \textbf{Fine-Tuning–Based Methods}  
    In this approach, LLMs are fine-tuned on dedicated summarization datasets. By adjusting the model's internal parameters, fine-tuning allows LLMs to specialize in summarization, further improving their accuracy and relevance for this task.
    \item \textbf{Knowledge Distillation–Based Methods}  
    These approaches extract knowledge from LLMs to create smaller, task-specific models. By distilling summarization capabilities into compact models, it is possible to retain high performance while improving efficiency for specific summarization tasks.
\end{enumerate}

The field of Automatic Text Summarization (ATS) has followed a development path through several distinct eras, from early statistical approaches to the advent of Large Language Models (LLMs). Conventional ATS approaches are constrained by a fixed summarization paradigm, either extractive or abstractive. The emergence of LLMs represents a significant breakthrough, offering a flexible, paradigm-agnostic approach that surpasses previous limitations in both extractive and abstractive summarization. As the field continues to evolve, we anticipate that LLM-based methods will not only enhance summarization capabilities but also pave the way for more adaptive, scalable, and efficient solutions. Future research will likely focus on improving the interpretability, scalability, and domain adaptability of these models, cementing their role as the foundation for next-generation ATS systems.

\subsection{Categorization of Automatic Text Summarization}
\label{sec:cat}

Automatic Text Summarization (ATS) systems are conventionally classified as ``Extractive'', ``Abstractive'' and ``Hybrid'' based on their generation paradigms. Extractive methods focus on selecting original sentences or phrases from the input text to create summaries, while abstractive approaches generate new sentences that convey the same meaning but differ from the original. Hybrid methods combine both techniques. Conventional methods are inflexible, as they are specifically designed and trained to perform either extraction or abstraction.

By contrast, LLM-based approaches unify summarization within a generative framework, enabling both extractive and abstractive outcomes through prompting. Examples include LLMs extracting key information from the original text \cite{liSkillGPTRESTfulAPI2023,mishra-etal-2023-llm}, reading the text and generating a summary \cite{heZCodePretrainedLanguage,chuangSPeCSoftPromptBased2023}, and applying hybrid strategies \cite{ravautSummaRerankerMultiTaskMixtureExperts2022,ghadimiHybridMultidocumentSummarization2022}.

In this survey, we classify ATS methods into two major categories: \textbf{Conventional Methods} (extractive, abstractive and hybrid) and \textbf{LLM-based Methods}. The former adhere strictly to fixed paradigms, while the latter demonstrate adaptability across generation paradigms, flexibly producing coherent summaries without traditional constraints.


\begin{table*}[t]
\caption{Overview of past ATS surveys on their summarization domain, type and methods coverage}
\label{table:survey_overview}
\renewcommand\arraystretch{1.2}
\begin{center}
\setlength{\tabcolsep}{2mm}{
\resizebox{\textwidth}{!}{
\begin{tabular}{llllll}
\hline\hline
\textbf{Survey Ref.}                                     & \textbf{Domain} & \textbf{Type}       & \textbf{Methods Coverage}                                                                                                                                                           & \textbf{Citation} & \textbf{Year} \\
\hline
\cite{gholamrezazadeh2009comprehensive} & General         & Comprehensive       & Graph;Machine Learning;Rule-based;Statistical                                                                                                                                       & 130               & 2009          \\
\cite{gupta2010a}                       & General         & Extractive          & Concept-based;Fuzzy Logic;Graph;Neural Network                                                                                                                                      & 889               & 2017          \\
\cite{moratanch2017survey}              & General         & Extractive          & Graph;Machine Learning;Neural Network;Rule-based;Statistical                                                                                                                        & 243               & 2017          \\
\cite{nazari2018}                       & General         & Comprehensive       & Machine Learning;Neural Network;Rule-based;Statistical;Term Frequency                                                                                                               & 124               & 2018          \\
\cite{lin2019}                          & General         & Abstractive         & Deep Language Model;Graph;Reinforcement Learning;Rule-based                                                                                                                         & 120               & 2019          \\
\cite{10.1007/978-981-13-0589-4_7}     & General         & Hybrid              & Neural Network;Term Frenquency                                                                                                                                                      & 21                & 2019          \\
\cite{syed2021}                         & General         & Abstractive         & Deep Language Model;Neural Network;Word Embedding                                                                                                                                   & 51                & 2021          \\
\cite{el-kassas2021}                    & General         & Comprehensive       & \begin{tabular}[c]{@{}l@{}}Concept-based;Deep Language Model;Graph;Machine Learning;Neural Network;\\ Fuzzy Logic;Rule-based;Statistical;Term Frequency;Word Embedding\end{tabular} & 545               & 2021          \\
\cite{mridha2021}                       & General         & Comprehensive       & \begin{tabular}[c]{@{}l@{}}Concept-based;Deep Language Model;Graph;Machine Learning;Neural Network;\\ Fuzzy Logic;Rule-based;Statistical;Term Frequency;Word Embedding\end{tabular}                                            & 53                & 2021          \\
\cite{kohEmpiricalSurveyLong2023}       & General         & Comprehensive       & Deep Language Model;Graph;Neural Network                                                                                                                                            & 44                & 2023          \\
\cite{10.1145/3166054.3166058}   & Domain-Specific & Dialogue            & Deep Learning;Knowledge Base;Machine Learning;Retrieval;Term Frequency                                                                                                              & 774               & 2017          \\
\cite{li2018}                           & Domain-Specific & Dialogue            & Machine Learning;Statistical                                                                                                                                                        & 98                & 2018          \\
\cite{Kanapala2019}                     & Domain-Specific & Legal               & Graph;Term Frequency;Statistical                                                                                                                                                    & 166               & 2019          \\
\cite{liu2019}                          & Domain-Specific & Graph-based Methods & Graph                                                                                                                                                                               & 348               & 2019          \\
\cite{IBRAHIMALTMAMI20221011}           & Domain-Specific & Scientific Article  & \textit{Do Not Apply}                                                                                                                                              & 73                & 2022          \\
\cite{feng2022}                         & Domain-Specific & Dialogue            & \textit{Do Not Apply}                                                                                                                                              & 69                & 2022          \\
\cite{10.1145/3529754}                  & Domain-Specific & Multi-Document      & Deep Language Model;Graph;Machine Learning;Neural Network                                                                                                                           & 93                & 2022          \\
\cite{jain2022survey}                   & Domain-Specific & Medical             & \textit{Do Not Apply}                                                                                                                                              & 284               & 2022          \\ \hline\hline
\end{tabular}
    }
    }
    \end{center}
\vspace{-2.0em}
\end{table*}

\subsubsection{Conventional ATS Methods}

ATS methods are conventionally categorized as extractive, abstractive, or hybrid. Extractive methods extract key phrases from the text, while abstractive methods rephrase or synthesize content based on semantic understanding. Hybrid methods combine these approaches. These methods are often inflexible, constrained by their paradigms, and require manual feature engineering and task-specific training.

\begin{itemize}
    \item \textbf{Extractive Methods} select important sentences or phrases directly from the original document to form a summary. It identifies key components in the text and uses them to create a coherent output \cite{7944061}. Techniques include unsupervised methods, such as calculating and ranking the statistical importance of words and sentences, and supervised methods, which train machine learning or deep learning models to classify text as ``summary'' or ``non-summary.''
    \item \textbf{Abstractive Methods} generate summaries by producing new sentences that may not appear in the original text. This can be achieved through structured methods, like tree or graph-based models, or generative approaches, such as sequence-to-sequence (seq2seq) architectures using RNNs \cite{mikolov2010recurrent} or Transformers \cite{Vaswani2017AttentionIA}. Pre-trained models often enhance these methods, enabling more flexible and human-like summary generation.
    \item \textbf{Hybrid Methods} combine extractive and abstractive approaches. Typically, a hybrid ATS system uses an extractive model to identify key sentences, followed by an abstractive model to refine or rewrite the extracted content. This process can take two forms: a simpler ``Extractive to Shallow Abstractive'' approach or a more sophisticated ``Extractive to Abstractive'' approach with a specifically trained abstractive model \cite{el-kassas2021}.
\end{itemize}

\subsubsection{LLM-based Methods}

Large Language Models (LLMs) differ from conventional ATS methods, which are constrained by pre-designed generation paradigms and limited training data. LLMs are pre-trained on vast datasets using extensive neural architectures, granting them superior generative capabilities. They can produce highly coherent and fluent text and, through in-context and few-shot learning, adaptively generate summaries in both extractive and abstractive styles based on task requirements \cite{chuangSPeCSoftPromptBased2023,zhengBIMGPTPromptBasedVirtual}.

LLM-based ATS methods face challenges such as accurately following summarization instructions, incorporating task-specific knowledge, and addressing issues like ``hallucinations.'' To improve their performance, three key research directions have emerged: (1) Prompt Engineering, which focuses on designing effective prompts, templates, and examples to guide LLMs in generating accurate and task-specific summaries; (2) Fine-tuning, which adapts LLMs with domain-specific data to enhance their understanding and the relevance of generated summaries; and (3) Knowledge Distillation, which extracts knowledge from LLMs to train smaller, specialized models for specific summarization tasks.





\subsection{Related Surveys on ATS} 


Several ATS surveys have been published over the years, generally adopting a technical categorization approach by classifying ATS methods as either ``extractive'' or ``abstractive'', which are summarized in Table \ref{table:survey_overview}. For instance, studies like \cite{gupta2010a,moratanch2017survey} focus on ``extractive'' methods, which select key sentences or paragraphs from the original documents to form concise summaries. These surveys cover techniques such as term frequency, statistical models, and supervised learning approaches. Similarly, \cite{suleiman2020} highlights extractive methods that rely on neural models. In contrast, research like \cite{gupta2019,lin2019} explores abstractive summarization methods, documenting the transition from early statistical and rule-based techniques to the latest advancements in neural language models. Additionally, \cite{el-kassas2021} provides a comprehensive review of both extractive and abstractive approaches.

Another line of surveys focuses on domain-specific ATS techniques, addressing summarization tasks for different content fields. For instance, survey works such as \cite{10.1007/978-3-030-15712-8_27,Kanapala2019,JAIN2021100388} review specialized methodologies designed for legal document summarization, while \cite{10.1145/3529754} offers a comprehensive review of methods for summarizing multiple documents. \cite{feng2022} focuses on summarizing dialogue and conversational texts, and \cite{10.1007/978-981-13-1498-8_76} provides an overview of methods for summarizing micro-blog content. Collectively, these surveys contribute to a nuanced understanding of ATS by illustrating how summarization techniques adapt to different content domains. 

In addition to summarizing conventional extractive and abstractive methods, this survey expands the scope by reviewing the latest advancements in LLM-based summarization techniques. By examining how large language models (LLMs) have transformed the field with their few-shot, in-context learning and superior generative abilities in text summarization.

\section{Definition and Literature Collection Methodology for Automatic Text Summarization}
\label{sec:Definition}

\subsection{Definitions for Automatic Text Summarization}

\begin{defi}[\textbf{Text Summarization}]
Text summarization can be defined as a mapping function 
\begin{equation}
f_{\theta}: T \rightarrow S,
\end{equation}
where \( f_{\theta}\) is the summarization method with trainable parameter \( \theta\), \( T \) is the set of tokens in the input text and \( S \subseteq T \) (for extractive summarization) or \( S \) is a newly generated sequence (for abstractive summarization). The objective is to minimize the information loss

\begin{equation}
\mathcal{L}(\theta) = \text{dist}(T, f_{\theta}(T)),
\end{equation}

\noindent subject to the constraint that \( S \) is significantly smaller in length than \( T \), while preserving the semantic and syntactic integrity of the original text.
\end{defi}

Considering that Automatic Text Summarization (ATS) techniques involve a series of intermediate steps to achieve the final objective, i.e., generating a concise and informative summary. We define the ``ATS Process'' as follows:

\begin{defi}[\textbf{Automatic Text Summarization (ATS) Process}]
The process of automatic text summarization can be formalized as a sequence of operations on the input text, denoted by a tuple 
\[
P = (D, M, G, E),
\]
where:
\begin{itemize}
    \item \( D \) represents \textbf{data preprocessing}, which applies transformations \( \phi(T) \) such as tokenization or cleaning on the input \( T \).
    \item \( M \) is the \textbf{modeling} step, where a function \( f_\theta(T) \) is trained or applied to produce a compressed representation \( h(T) \), where \( h(T) \ll T \).
    \item \( G \) is the \textbf{generation} step, applying the function \( g(h(T)) \) to produce the summary \( S \), either by extracting or generating a subset of \( T \).
    \item \( E \) is the \textbf{evaluation} phase, using metrics \( E_m(S, S^*) \) to measure how well the generated summary \( S \) matches the ground truth \( S^* \), with commonly used metrics being ROUGE and BLEU.
\end{itemize}
\end{defi}

\begin{figure*}[t]
    \centering
	\includegraphics[width=1\textwidth]{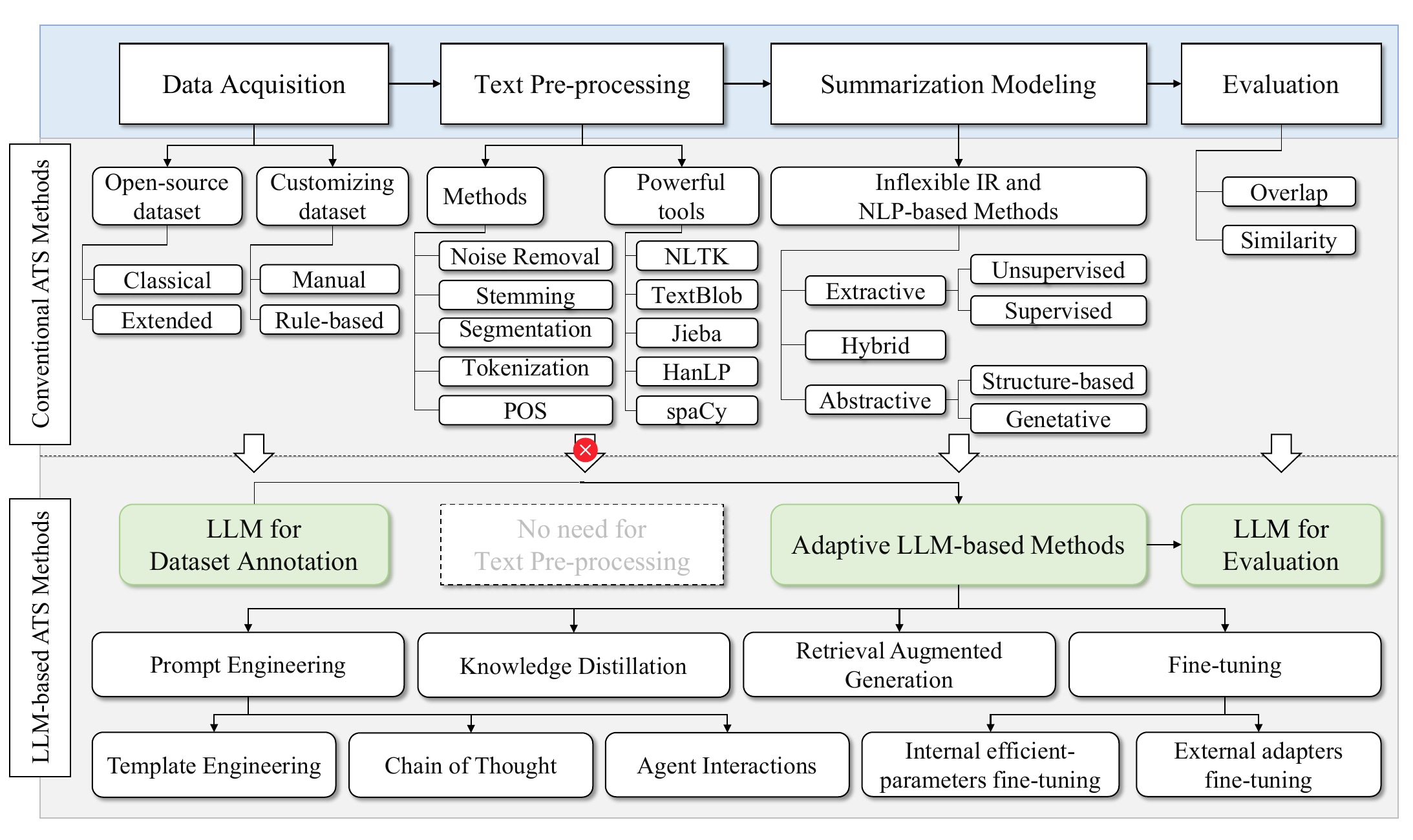}
	\caption{Overview of the ATS (Automatic Text Summarization) framework, comparing conventional and LLM-based methods. Traditional approaches follow a staged pipeline, including data acquisition, pre-processing, modeling, and evaluation, using rule-based techniques and NLP tools. In contrast, LLM-based methods streamline the process via end-to-end summarization without explicit pre-processing.}
    \label{framework}
\vspace{-1.5em}
\end{figure*}

\subsection{Process of Automatic Text Summarization}

Building upon the definition of ATS, we delineate the intermediate steps necessary to achieve the goal of abstraction generation. The ATS process is illustrated in Figure \ref{framework}, and its constituent steps are defined as follows:

\begin{enumerate}
    \item \textbf{Data Acquisition.} 
    Data Acquisition, the initial step of the ATS process, involves obtaining datasets critical for the system. The specifics of this step are elaborated in Section \ref{data}, which provides a comprehensive overview of existing datasets for ATS, along with methodologies for constructing new datasets from scratch.
    
    \item \textbf{Text Pre-processing.}
    Pre-processing is a crucial step aimed at refining the collected texts by removing noise and transforming raw texts into a clean, structured format, as elucidated by \cite{gupta2010a}. This step predominantly employs linguistic techniques such as noise removal, stemming, and sentence/word segmentation to enhance the quality of the text data. For a detailed exploration of the pre-processing techniques and their applications in ATS, refer to Section \ref{pre-process}.

     \item \textbf{Summarization Modeling.} 
    Modeling, the cornerstone of an ATS system, is dedicated to developing versatile language models that can interpret and distill language data into concise summaries. This is achieved through rule-based, statistical, or deep learning approaches which extract patterns from the language data. The process of modeling in ATS is inherently an NLP task, typically commencing with language modeling and subsequently progressing to summarization modeling. The conventional ATS methods are bifurcated into language modeling and summarization modeling. For an in-depth discussion and categorization of the various ATS models, refer to Section \ref{sec:conventional_ats} for conventional approaches and Section \ref{sec:llm_ats} for methods based on Large Language Models (LLMs).

     \item \textbf{Evaluation Metrics.}
    Evaluation metrics of ATS to judge how well ATS works. 
    Objective, comprehensive, and accurate evaluation metrics 
    can lead to the recognition and acceptance of research on ATS. Refer to Section \ref{sec:evaluation} for details.
\end{enumerate}

Based on the framework, the core distinction lies in the autonomous closed-loop capability of LLM for ATS, which fundamentally differs from traditional methods. Conventional ATS systems rely on fragmented pipepline regquried manual interventions, such as preprocessing of the input texts and sequential stages of language modeling followed by summarization modeling; whereas LLMs operate as unified end-to-end systems that bypass these intermediate steps entirely, as illustrated in Figure~\ref{framework}. This is evident in the visual contrast between the top (conventional) and bottom (LLM-based) halves of the figure: LLMs directly process raw text inputs through self-supervised learning mechanisms, eliminating the need for explicit text normalization or modular separation. Furthermore, LLMs inherently support self-contained data annotation via techniques like prompt engineering and retrieval-augmented generation, enabling dynamic synthesis of training data. The evaluation phase is similarly integrated, as LLMs can self-assess generated summaries through internal consistency checks or adversarial sampling, forming a closed-loop optimization system. This holistic capability allows LLMs to autonomously traverse data construction, modeling, and evaluation phases with minimal external intervention.

Based on the framework, the core distinction is the autonomous closed‑loop capability of LLMs, which fundamentally differs from traditional methods. Conventional ATS systems rely on fragmented pipelines requiring manual interventions, i.e., preprocessing input texts and training separate language‑modeling and summarization stages; whereas LLMs operate as unified end‑to‑end systems that bypass these steps entirely (Figure~\ref{framework}). This contrast is evident in the visual comparison of the top (conventional) and bottom (LLM‑based) sections: LLMs directly process raw text via self‑supervised learning, eliminating the need for explicit text normalization and modular separation. Moreover, LLMs inherently support self‑contained data annotation through techniques like prompt engineering and retrieval‑augmented generation, enabling dynamic synthesis of training data. The evaluation phase is likewise integrated—LLMs can self‑assess generated summaries via internal consistency checks or adversarial sampling, completing a closed‑loop optimization. This holistic capability enables LLMs to traverse data collection, modeling, and evaluation autonomously with minimal external intervention.

However, LLM-based methods also present challenges: while they generate fluent, contextually rich summaries, they occasionally produce factually inconsistent outputs compared to traditional statistical models. These trade-offs are discussed in detail in Section \ref{model:LLM} alongside mitigation strategies like Retrieval Augmented Generation for ATS.

\begin{algorithm}[ht]
\setlength{\textfloatsep}{5pt} 
\setlength{\floatsep}{5pt}     
\caption{Algorithm for Crawling ATS Relevant Papers} 
\label{algo:base} 
\begin{algorithmic}
\State Initialize an empty database \( D \)
\State Initialize web crawler \( C \)
\State Initialize the LLM for classification \( LLM \)
\State Define search keywords \( K = \{\texttt{summarization}, \dots\} \)

\ForAll{\( k \in K \)}
    \State \( results \gets C.search(k) \) \Comment{Perform search using keyword \( k \)}
    \ForAll{\( r \in results \)}
        \State Extract \( paper = \{title, pub\_url\} \) from \( r \)
        \If{\( paper \notin D \)} 
            \State Add \( paper \) to \( D \) 
        \EndIf
    \EndFor
\EndFor

\ForAll{\( paper \in D \)}
    \State \( page \gets C.crawl(paper.pub\_url) \)
    \If{\( paper.pub\_url \) ends with \texttt{.pdf}}
        \State Extract \( abstract \) from \( page.FILE \)
    \Else
        \State Extract \( abstract \) from \( page.HTML \)
    \EndIf
    \State Clean and store \( abstract \) in \( D \)
\EndFor

\State Define few-shot examples \( SHOTs = \{\{title, abstract, label\}\} \)

\ForAll{\( paper \in D \)}
    \State \( result \gets LLM(SHOTs, paper.title, paper.abstract) \)
    \State Add \( result \) to categorized dataset \( R \)
\EndFor

\State \Return \( R \)
\end{algorithmic}
\end{algorithm}

\subsection{Methodology for Crawling ATS Papers}
\label{sec:paper_crawling}
We programmed an automated crawling algorithm (Algorithm \ref{algo:base}) to efficiently collect papers relevant to ATS, addressing the limitations of manual keyword-based searches. By integrating web crawling and Large Language Model (LLM)-based filtering, the algorithm enhances both the accuracy and comprehensiveness of the retrieval process.

Manual selection of ATS papers from Google Scholar poses two key challenges. First, relevant papers may use diverse and non-standard keywords such as ``summarization,'' ``text condensation,'' or ``summary generation,'' making manual searches time-consuming. Second, some papers relevant to ATS may lack explicit keywords in their titles but include them in their abstracts, as seen in examples like ``ChatGPT Chemistry Assistant for Text Mining and the Prediction of MOF Synthesis'' \cite{zhengChatGPTChemistryAssistant}, where ``summarization'' appears only in the abstract. Similar issues occur with other works \cite{zhaoSLiCHFSequenceLikelihood2023, xu-etal-2023-inheritsumm}.

To this end, we designed a three-stage algorithm that automates the collection, filtering, and categorization of ATS papers. The process leverages synonym-enhanced keyword searches and analyzes both titles and abstracts for relevance, which are as follows:

\begin{enumerate}
    \item \textbf{Paper Searching:} Crawl papers from Google Scholar using a diverse set of synonym-enhanced keywords (e.g., ``summarization,'' ``text condensation''). The results are deduplicated and preprocessed to create an initial dataset \( D \).
    
    \item \textbf{Relevance Filtering:} Retrieve abstracts to filter out irrelevant papers. Since Google Scholar often provides incomplete abstracts, the algorithm additionally retrieves their publication URLs (\( pub\_url \)) to scrape full abstracts. For URLs pointing to PDFs, text is extracted directly from the files using PyMuPDF\footnote[1]{https://github.com/pymupdf/PyMuPDF}.
    
    \item \textbf{Categorization Using LLMs:} Leverage LLaMA-3-Instruct 8B model with specifically designed few-shot examples to classify the crawled papers based on their titles and abstracts into predefined categories (e.g., datasets, methodologies, applications). The outputs are manually reviewed to ensure accuracy.
\end{enumerate}

This pipeline ensures scalable and precise retrieval of ATS-related papers, addressing the limitations of keyword-based searches. The algorithm is outlined in Algorithm \ref{algo:base}.

\begin{table*}[t]
\centering
\caption{Evaluation Results of the Proposed Retrieval Algorithm for LLM-Based ATS Papers}
\label{t:cls_res}
\renewcommand\arraystretch{1.2}
\setlength{\tabcolsep}{6pt}
\begin{center}
    \setlength{\tabcolsep}{2mm}{
    \resizebox{\textwidth}{!}{
\begin{tabularx}{\textwidth}{lXccc}
\hline\hline  
    \textbf{Task} & \textbf{Task Description} & \textbf{Accuracy} & \textbf{Macro-F1} & \textbf{Micro-F1} \\  
    \hline  
    \texttt{is\_LLM} & Determine whether the article discusses LLM-based ATS methods. & 94.2 & N/A & N/A \\
    \hline      
    \texttt{is\_dataset} & Identify whether the article introduces or utilizes a dataset for ATS tasks. & 93.4 & N/A & N/A \\
    \hline  
    \texttt{Type} & Classify the article's contribution type (e.g., survey, novel method, evaluation). & 85.6 & 71.9 & 85.6 \\
    \hline  
    \texttt{Methodology} & Identify the methodological approach used (e.g., prompting, fine-tuning, distillation). & 86.0 & 85.9 & 86.1 \\
    \hline  
    \texttt{Domain} & Determine the application domain of the article; mark as general if unspecified. & 92.0 & 83.6 & 92.0 \\
    \hline\hline  
\end{tabularx}
}
    }
    \end{center}
\vspace{-2em}
\end{table*}

\subsection{Evaluation Methodology and Results of LLM-based ATS Retrieval Algorithm \ref{algo:base}}

To evaluate the effectiveness of the retrieval algorithm in identifying ATS-related papers, we manually annotated 1,405 papers retrieved using Algorithm~\ref{algo:base}. Each paper was labeled across five dimensions: whether it involves LLM-based ATS methods (is\_LLM), whether it introduces or uses a dataset for ATS tasks (is\_dataset), the type of contribution (e.g., survey, novel method, evaluation; Type), the employed methodology (e.g., prompting, fine-tuning; Methodology), and the application domain within ATS (Domain). For each classification task, we report Accuracy, Macro-F1, and Micro-F1 scores. A summary of the evaluation results is presented in Table~\ref{t:cls_res}.

To establish ground-truth labels for evaluation, we employed a structured manual annotation protocol. Three PhD students with research expertise in natural language processing and data mining were recruited to independently annotate each of the 1,405 papers retrieved by Algorithm~\ref{algo:base}. The annotation covered five dimensions as outlined previously.

To ensure labeling quality and reduce individual bias, we adopted a majority voting strategy: for each instance, the final label was determined if at least two of the three annotators agreed. Disagreements without a majority consensus were discussed and resolved through joint review. All annotators received detailed annotation guidelines and conducted a pilot annotation round to calibrate their understanding before the full-scale task. For more detailed
information, please refer to \ref{app:1}.

\subsubsection{Retrieval Algorithm Result Analysis}

The evaluation results, summarized in Table~\ref{t:cls_res}, demonstrate that the proposed retrieval algorithm enables highly effective classification across five annotation dimensions. The binary classification tasks (is\_LLM and is\_dataset) achieved strong performance, with accuracy scores of 94.2\% and 93.4\%, respectively, indicating high precision in identifying whether articles pertain to LLM-based methods or ATS datasets.

For the multi-class tasks, the model also performed robustly. The Type task achieved an accuracy of 85.6\%, with a macro-F1 score of 71.9 and a micro-F1 of 85.6, suggesting that while the model performed well on common categories, there was room for improvement in handling less frequent types. The Methodology task yielded a macro-F1 of 85.9 and a micro-F1 of 86.1, reflecting consistent performance across both dominant and rare techniques. The Domain classification task also showed strong results, with 92.0\% accuracy, 83.6 macro-F1, and 90.0 micro-F1, confirming the model's ability to distinguish between general and domain-specific ATS applications.

\subsubsection{Error Analysis and Implications of Retrieval Algorithm}

Two primary sources of classification error were identified:

\begin{enumerate}
    \item Terminological Ambiguity: Classification performance, particularly for the \textit{is\_LLM} and \textit{Methodology} dimensions, was affected by implicit or ambiguous phrasing in the original texts. For example, references to models such as ``Transformer'' or ``pretrained language model,'' without explicitly identifying them as LLMs, led to misclassification. Additionally, overlapping or loosely defined methodological terms hindered accurate and consistent categorization.

    \item Long-Tail Distribution and Class Imbalance: In multi-class tasks such as \textit{Type} and \textit{Methodology}, a small number of dominant categories (e.g., model-centric studies) appeared much more frequently than others. This skewed distribution contributed to performance disparities, as reflected in the divergence between macro-F1 and micro-F1 scores. While high-frequency classes inflated the micro-F1, the macro-F1 revealed reduced effectiveness in identifying less-represented categories.
\end{enumerate}

Despite the two errors, the overall annotation and classification framework demonstrated strong performance across all dimensions, indicating the reliability of the retrieval algorithm for ATS-related literature. In addition, the algorithm shows potential generalizability, which could potentially be expanded to other types of papers collection by altering the prompting words. Future improvements may include integrating context-aware disambiguation mechanisms, leveraging domain-specific lexicons, and applying data augmentation techniques to mitigate long-tail effects and enhance the representation of underrepresented categories.

\section{ATS Datasets}
\label{data}
This section introduces two types of datasets for ATS: (1) open-source datasets commonly used in existing research and (2) techniques to build new ATS datasets.

Open-source datasets are mostly adopted for general ATS tasks but often lack suitability for domain-specific applications, as noted in studies \cite{feng2020dialogue,liu2022clts+,asi2022end}. To address these gaps, we outline methodologies for creating new custom datasets, focusing on leveraging Large Language Models (LLMs). LLMs enable scalable and efficient generation of domain-specific data, reducing manual effort and enhancing dataset diversity.

\subsection{Open-source Datasets}

We collected datasets in the ATS domain that are publicly accessible online or via email, referred to as \textit{Open-Source Datasets}. We adopted the automated retrieval algorithm illustrated in Sec. \ref{sec:paper_crawling} to retrieve the ATS datasets by using ATS dataset-related queries. This process also consists of three main steps:

\begin{enumerate}
    \item \textbf{Dataset Searching:} Using the keyword ``summarization dataset,'' we searched Google Scholar for papers and associated metadata, sorted by relevance.
    
    \item \textbf{Relevance Filtering:} To exclude irrelevant articles, we employed an LLM for few-shot learning classification. Titles and abstracts were analyzed and categorized into two groups: articles introducing new datasets, labeled as ``dataset,'' and those unrelated, labeled as ``non-dataset''. Manual screening was then conducted to ensure the accuracy of the classification, resulting in a definitive list of relevant datasets.
    
    \item \textbf{Categorization:} Since articles on datasets tend to be structured and straightforward, we manually categorized the collected datasets based on their citation impact and relevance to ATS research.
\end{enumerate}

Given the large number of ATS datasets available, we organized them into two groups for clarity and ease of presentation:

\begin{itemize}
    \item \textbf{Conventional Datasets:} These are well-established datasets that have been widely adopted in ATS research and have each received over 100 citations. They are considered foundational benchmarks for model training and evaluation. For each dataset, we provide a brief contextual overview in the text and summarize key attributes, such as publication year, size, domain, language, and accessibility, in Table~\ref{table:datasets}. 
    
    \item \textbf{Supplementary and Emerging Datasets:} This category includes datasets with fewer than 100 citations, consisting of both recently released and domain-specific resources that are still gaining traction. While these datasets are less established, which are summarized in Table~\ref{table:datasets-ex}.

\end{itemize}

\begin{table*}[t]
    \caption{Conventional Datasets with more than 100 citations in the ATS. The table is arranged in ascending order by publication year and provides details on the publish year, size (number of total pairs including train, validation, and test), domain, language, along with their publicly available URLs for download.}
    \label{table:datasets}
    \renewcommand\arraystretch{1.2}
\begin{center}
    \setlength{\tabcolsep}{2mm}{
    \resizebox{\textwidth}{!}{
    \begin{tabular}{lllllll}
    \hline\hline
Name              & Year & Size            & Domain         & Language     & URL                                                        \\ \hline
Gigaword          & 2003 & 9,876,086       & News           & English      & https://github.com/harvardnlp/sent-summary                 \\
DUC 2001-2007     & 2004 & 250-1600        & News           & English      & https://duc.nist.gov/data.html                             \\
CNN \& Daily Mail & 2016 & 312,084         & News           & English      & https://github.com/abisee/cnn-dailymail                    \\
LCSTS             & 2016 & 2,400,591       & Blogs          & Chinese      & http://icrc.hitsz.edu.cn/Article/show/139.html             \\
ArXiv      & 2018 & 215,000 & Academic paper & English      & https://github.com/armancohan/long-summarization           \\
PubMed              & 2018 & 133,000         & Academic paper           & English      & https://github.com/armancohan/long-summarization  \\
XSum              & 2018 & 226,711         & News           & English      & https://github.com/EdinburghNLP/XSum                       \\
NEWSROOM          & 2018 & 1.3M            & News           & English      & https://lil.nlp.cornell.edu/newsroom/                      \\
WikiHow           & 2018 & 230,843         & Knowledge Base & English      & https://github.com/mahnazkoupaee/WikiHow-Dataset           \\
Multi-News        & 2019 & 56,216          & News           & English      & https://github.com/Alex-Fabbri/Multi-News                  \\
SAMSum            & 2019 & 16,369          & Dialogue       & English      & https://arxiv.org/src/1911.12237v2/anc/corpus.7z           \\
BIGPATENT         & 2019 & 1.3M            & Patent         & English      & https://evasharma.github.io/bigpatent/                     \\
Scisumm           & 2019 & 1,000           & Academic paper & English      & https://cs.stanford.edu/$\sim$myasu/projects/scisumm\_net/ \\
BillSum           & 2019 & 22,218          & Bill           & English      & https://github.com/FiscalNote/BillSum                      \\
WikiLingua        & 2020 & 141,457         & Knowledge Base & Multilingual & https://github.com/esdurmus/Wikilingua                     \\
MLSUM             & 2020 & 1.5M            & News           & Multilingual & https://github.com/ThomasScialom/MLSUM                     \\
XL-sum            & 2021 & 1.35M           & News           & Multilingual & https://github.com/csebuetnlp/xl-sum                       \\
DialogSum         & 2021 & 13,460          & Dialogue       & English      & https://github.com/cylnlp/DialogSum                        \\
MediaSum          & 2021 & 463,600         & Interview      & English      & https://github.com/zcgzcgzcg1/MediaSum/                    \\
Booksum           & 2021 & 146,532         & Literature     & English      & https://github.com/salesforce/booksum                      \\
Summscreen        & 2021 & 26,900          & TV series      & English      & https://github.com/mingdachen/SummScreen                   \\ \hline\hline
\end{tabular}}
    }
    \end{center}
\vspace{-2em}
\end{table*}

\subsubsection{Conventional Datasets for ATS}
This subsection presents a collection of widely adopted and historically influential datasets commonly used in Automatic Text Summarization (ATS) research. These datasets, many of which have accumulated over 100 citations, serve as conventional benchmarks for training and evaluating summarization models. Table~\ref{table:datasets} provides key information including publication year, dataset size (number of pairs across training, validation, and test splits), domain, language, and public access links. Below, we offer a brief overview of each dataset to contextualize its relevance and usage.

\textbf{Gigaword}\cite{graff2003} is a news summarization dataset from the Linguistic Data Consortium (LDC), sourced from seven media outlets, including \textit{Agence France-Presse}, \textit{Associated Press}, \textit{Bloomberg}, and \textit{Xinhua News Agency}. With nearly 10 million English news documents and headline-based summaries, it is well-suited for training deep neural networks. The dataset directly uses headlines as summaries\cite{el-kassas2021}.

\textbf{DUC}\cite{harman2004effects} is a series of high-quality datasets from the Document Understanding Conferences (2001–2007), containing 1600 news document-summary pairs. Summaries are available in three forms: manually created, automatically generated baselines, and algorithm-generated submissions. Due to their small size and quality, DUC datasets are commonly used for testing\cite{lin2019}.

\textbf{CNN \& Daily Mail}\cite{hermann2015teaching} is a news dataset extracted from \textit{CNN} and \textit{Daily Mail}, featuring 286,817 training, 13,368 validation, and 11,487 test pairs. It includes news body contents paired with editor-created highlights as summaries. Training documents average 766 words across 29.74 sentences, with summaries averaging 53 words and 3.72 sentences. The dataset has two versions: non-anonymous (real entity names) and anonymous (entity names removed).

\textbf{LCSTS}\cite{hu2016} contains more than 2 million Chinese microblogs from domains including politics, economics, and entertainment, sourced from \textit{Sina Weibo's} official accounts such as People's Daily, the Ministry of National Defense, etc. Summaries are manually annotated, with rules applied to ensure quality (e.g., accounts with over 1 million followers). The dataset is divided into three parts: a large master dataset, a high-quality subset with manual scoring, and a refined test set.

\textbf{ArXiv, PubMed}\cite{cohan2018} are academic datasets with over 300,000 papers from \textit{arXiv.org} and \textit{PubMed.com}, using abstracts as summaries. Pre-processing includes filtering out excessively long or short documents, removing figures and tables, normalizing math formulas and citations, and retaining only relevant sections. For arXiv, LATEX files are converted to plain text with Pandoc to preserve discourse structure.

\textbf{XSum}\cite{narayan2018a} is a dataset of 226,711 \textit{BBC} news articles (2010– 2017) with single-sentence professionally written summaries, often authored by the article's writer. The dataset spans diverse domains, including politics, sports, business, technology, and entertainment.

\textbf{NEWSROOM}\cite{gruskyNewsroomDataset132020} contains 1.3 million articles and human-written summaries from 38 major news outlets (1998–2017). It features diverse summarization styles, combining abstractive and extractive approaches, crafted by newsroom authors and editors across various domains like news, sports, and finance.

\textbf{WikiHow}\cite{Koupaee2018WikiHowAL} comprises over 230,000 article-summary pairs from the \textit{WikiHow} knowledge base, covering procedural tasks across various topics. Summaries are formed by concatenating bold introductory lines for each step, while detailed descriptions form the source article. Articles are categorized as single-method tasks or multi-method tasks.

\textbf{Multi-News}\cite{fabbri2019} contains 56,216 article-summary pairs from \textit{newser.com}, with professionally written summaries that include links to cited articles. The summaries are longer (averaging 260 words), have a lower compression rate, and less variability in copied words, aiding models in generating fluent and coherent text.

\textbf{SAMSum}\cite{gliwaSAMSumCorpusHumanannotated2019} is a dataset of over 16,000 human-created messenger-style conversations with annotated summaries. Designed by linguists to reflect real-life chat interactions, it emphasizes the complexity of dialogue summarization compared to news articles, requiring specialized models and evaluation metrics.

\textbf{BIGPATENT}\cite{sharmaBIGPATENTLargeScaleDataset2019} is a dataset of 1.3 million U.S. patent documents post-1971 across nine technological domains, featuring human-written abstractive summaries. Built from Google Patents Public Datasets, it addresses dataset limitations with richer discourse, balanced content, and higher compression ratios, supporting research in abstractive summarization.

\textbf{Scisumm}\cite{yasunaga2019} includes the 1,000 most cited academic papers (21,928 citations) from the \textit{ACL Anthology Network}, with expert-annotated summaries and citation information. The dataset construction involved sampling citations over time, filtering inappropriate sentences, and a pilot study with PhD students to validate annotations. Final summaries were created by annotators using abstracts and citing sentences.

\textbf{BillSum}\cite{kornilovaBillSumCorpusAutomatic2019} is the first dataset for summarizing US Congressional and California state bills, comprising 22,218 US bills (1993–2018) and 1,237 California bills (2015–2016), each with human-written summaries. The bills, mid-length (5,000–20,000 characters), were sourced from Govinfo and the California legislature's website to ensure diversity. The dataset supports research in legislative summarization and model transferability to new legislatures.

\textbf{WikiLingua}\cite{ladhakWikiLinguaNewBenchmark2020} is a cross-lingual abstractive summarization dataset with 141,457 English articles and 600k total articles in 17 languages, aligned via illustrative images. Sourced from WikiHow, it features high-quality, collaboratively written how-to content, supporting research into cross-lingual summarization using synthetic data and pre-trained Neural Machine Translation.

\textbf{MLSUM}\cite{scialomMLSUMMultilingualSummarization2020} is a large-scale multilingual summarization dataset with over 1.5 million article-summary pairs in French, German, Spanish, Russian, and Turkish. Sourced from online newspapers, it uses articles and human-written highlights as pairs. The dataset supports cross-lingual research, with detailed statistics and plans for expansion to more languages.

\textbf{XL-sum}\cite{hasanXLSumLargeScaleMultilingual2021} is a multilingual abstractive summarization dataset with 1 million article-summary pairs in 44 languages, sourced from the BBC. It focuses on diverse and low-resource languages, offering high-quality, concise summaries. Fine-tuned mT5 models achieve competitive results, making XL-sum the largest dataset for multilingual summarization in both samples and language coverage.

\textbf{DialogSum}\cite{chenDialogSumRealLifeScenario2021} is a large-scale dialogue summarization dataset with 13,460 real-life dialogues on topics like education, work, and healthcare. It contains 1.8 million tokens and averages 131 tokens per dialogue. Created through careful cleaning, bi-turn formatting, and annotation, it addresses challenges like complex discourse, coreferences, and ellipsis, supporting research in dialogue summarization.

\textbf{MediaSum}\cite{zhuMediaSumLargescaleMedia2021} is a large-scale dialogue summarization dataset with 463.6K transcripts and abstractive summaries from NPR and CNN interviews. Covering diverse domains and multi-party conversations, it includes segmented summaries for multi-topic interviews. The dataset's size and positional biases improve model performance in transfer learning for dialogue summarization tasks.

\textbf{Booksum}\cite{kryscinskiBookSumCollectionDatasets2022} is a long-form narrative summarization dataset featuring novels, plays, and stories with human-written abstractive summaries. It challenges models with lengthy texts, causal/temporal dependencies, and highly abstractive summaries. Structured into paragraph, chapter, and book levels, it includes 217 titles, 6,327 chapters, and 146,532 paragraph-level examples. Sourced from Project Gutenberg and Web Archive, the dataset is cleaned, aligned, and split for training, validation, and testing, with provided baselines for extractive and abstractive models.

\textbf{Summscreen}\cite{chenSummScreenDatasetAbstractive2022} is a screenplay summarization dataset with 29,186 TV episode transcripts and human-written recaps from various genres. It includes 6,683 episodes from ForeverDreaming and 22,503 from TVMegaSite. Challenges include indirect plot expression and non-plot content. The dataset was filtered for character overlap and transcript length, split into train/dev/test sets, and features entity-centric evaluation metrics to reflect characters' key role in TV series.

\subsubsection{Supplementary and Emerging Datasets}
This subsection includes summarization datasets cited fewer than 100 times, encompassing both recently released resources and domain-specific datasets that have not yet achieved widespread adoption. Despite their lower citation counts, these datasets offer valuable diversity in domains, formats, and summarization objectives, making them useful for exploring underrepresented challenges in ATS. Table~\ref{table:datasets-ex} summarizes their characteristics, including size, domain, and availability.

\begin{table*}[h!t]
    \setlength\abovecaptionskip{0cm}
    \caption{Supplementary datasets with fewer than 100 citations in ATS. The table is formatted in the same manner as the Conventional Dataset Table, exhibiting diversity and serving as a valuable complement to ATS datasets.}
    \label{table:datasets-ex}
    \renewcommand\arraystretch{1.2}
\begin{center}
    \setlength{\tabcolsep}{2mm}{
    \resizebox{\textwidth}{!}{
    \begin{tabular}{lllllll}
    \hline\hline
Ref. & Name           & Year & Size      & Domain            & Language              & URL                                                             \\ \hline
\cite{caoTGSumBuildTweet}& TGSum          & 2016 & 1,114     & News \& Relation  & English               & http://www4.comp.polyu.edu.hk/                                  \\
\cite{nguyen-etal-2016-vsolscsum}    & VSoLSCSum      & 2016 & 2,448     & News              & Vietnamese            & https://github.com/nguyenlab/VSoLSCSum-Dataset                  \\
\cite{nguyenSoLSCSumLinkedSentenceComment}    & Solscsum       & 2016 & 5,858     & News \& Comment   & English               & http://150.65.242.101:9292/yahoo-news.zip                       \\
\cite{liReaderAwareMultiDocumentSummarization2017}    & RA-MDS         & 2017 & 19,000    & News \& Comment   & English               & http://www.se.cuhk.edu.hk/$\sim$textmine/dataset/ra-mds/        \\
\cite{kurniawanIndoSumNewBenchmark2019}    & Indosum        & 2018 & 19,000    & News              & Indonesian            & https://github.com/kata-ai/indosum                              \\
\cite{devargasfeijoRulingBRSummarizationDataset2018}    & Rulingbr       & 2018 & 10,000    & Ruling            & Portuguese            & https://github.com/diego-feijo/rulingbr                         \\
\cite{strakaSumeCzechLargeCzech}    & SumeCzech      & 2018 & 1,001,593 & News              & Czech                 & https://lindat.mff.cuni.cz/repository/xmlui/handle/11234/1-2615 \\
\cite{levTalkSummDatasetScalable2019a}    & Talksumm       & 2019 & 1,716     & Scientific papers & English               & https://github.com/levguy/talksumm                              \\
\cite{nguyenVNDSVietnameseDataset}    & Vnds           & 2019 & 150,704   & News              & Vietnamese            & https://github.com/ThanhChinhBK/vietnews                        \\
\cite{ghalandariLargeScaleMultiDocumentSummarization2020}    & WCEP           & 2020 & 2,390,000 & News              & English               & https://github.com/complementizer/wcep-mds-dataset              \\
\cite{luMultiXScienceLargescaleDataset2020a}    & Multi-XScience & 2020 & 40,528    & Scientific papers & English               & https://github.com/yaolu/Multi-XScience                         \\
\cite{kulkarniAQuaMuSeAutomaticallyGenerating2020}    & Aquamuse       & 2020 & 5,519     & Query             & English               & https://github.com/google-research-datasets/aquamuse            \\
\cite{gusevDatasetAutomaticSummarization}    & Gazeta         & 2020 & 63,435    & News              & Russian               & https://github.com/IlyaGusev/gazeta                             \\
\cite{kotoLiputan6LargescaleIndonesian2020}    & Liputan6       & 2020 & 215,827   & News              & Indonesian            & https://github.com/fajri91/sum\_liputan6                        \\
\cite{roushDebateSumLargescaleArgument}    & DebateSum      & 2020 & 187,386   & Debate            & English               & https://github.com/Hellisotherpeople/DebateSum                  \\
\cite{huangGeneratingSportsNews}    & SPORTSSUM      & 2020 & 5,428     & Sports            & Chinese               & https://github.com/ej0cl6/SportsSum                             \\
\cite{liuCLTSNewChinese2020}    & CLTS           & 2020 & 180,000   & Long articles     & Chinese               & https://github.com/lxj5957/CLTS-Dataset                         \\
\cite{antogniniGameWikiSumNovelLarge2020}    & GameWikiSum    & 2020 & 14,652    & Video game        & English               & https://github.com/Diego999/GameWikiSum                         \\
\cite{hayashiWikiAspDatasetMultidomain2020}    & WikiAsp        & 2021 & 320,272   & Multi-domain      & English               & http://github.com/neulab/wikiasp                                \\
\cite{mengBringingStructureSummaries2021}    & FacetSum       & 2021 & 60,532    & Scientific papers & English               & https://github.com/hfthair/emerald\_crawler                     \\
\cite{guptaSumPubMedSummarizationDataset2021}    & SumPubMed      & 2021 & 33,772         & Scientific papers & English               & https://github.com/vgupta123/sumpubmed                          \\
\cite{feigenblatTWEETSUMMDialogSummarization2021}    & TWEETSUMM      & 2021 & 6,500     & Dialogue          & English               & https://github.com/guyfe/Tweetsumm                              \\
\cite{linCSDSFineGrainedChinese2021}    & CSDS           & 2021 & 10,701    & Dialogue          & Chinese               & https://github.com/xiaolinAndy/CSDS                             \\
\cite{bhattacharjeeCrossSumEnglishCentricCrossLingual2023}    & CrossSum       & 2021 & 1,680,000 & News              & Multilingual          & https://github.com/csebuetnlp/CrossSum                          \\
\cite{fuMMAVSFullScaleDataset}    & MM-AVS         & 2021 & 2,260         & Multimodal        & English               & https://github.com/xiyan524/MM-AVS                              \\
\cite{parikhLawSumWeaklySupervised2021}    & Lawsum         & 2021 & 10,000    & Legal             & Indonesian            & https://github.com/mtkh0602/LegalSummarization                  \\
\cite{khalmanForumSumMultiSpeakerConversation2021}    & ForumSum       & 2021 & 4,058     & Dialogue          & English               & https://www.tensorflow.org/datasets/catalog/forumsum            \\
\cite{pasqualiTLSCovid19NewAnnotated2021}    & Tls-covid19    & 2021 & 100,399   & Timeline          & Portuguese \& English & https://github.com/LIAAD/tls-covid19                            \\
\cite{10.1145/3412841.3441946}    & ISSumSet       & 2021 &  136,263         & Tweet             & English               & https://github.com/AlexisDusart/ISSumSet                        \\
\cite{zhaoQBSUMLargeScaleQueryBased2021}    & QBSUM          & 2021 & 49,000    & Query             & Chinese               & https://github.com/yanshengli/QBSUM                             \\
\cite{wangClidSumBenchmarkDataset2022}    & Clidsum        & 2022 & 112,000   & Dialogue          & Multilingual          & https://github.com/krystalan/ClidSum                            \\
\cite{wangSQuALITYBuildingLongDocument2022}    & Squality       & 2022 & 625       & Story             & English               & https://github.com/nyu-mll/SQuALITY                             \\
\cite{aumillerEURLexSumMultiCrosslingual2022}    & EUR-lex-sum    & 2022 &    24 languages        & Legal             & Multilingual          & https://github.com/achouhan93/eur-lex-sum                       \\
\cite{maddelaENTSUMDataSet}    & EntSUM         & 2022 & 2788      & Entity            & English               & https://zenodo.org/record/6359875                               \\
\cite{aumillerKlexikonGermanDataset2022}    & Klexikon       & 2022 & 2,898     & Knowledge Base    & German                & https://github.com/dennlinger/klexikon                          \\
\cite{poddarCAVESDatasetFacilitate2022}    & Caves          & 2022 & 10,000    & Tweet             & English               & https://github.com/sohampoddar26/caves-data                     \\
\cite{mukherjeeECTSumNewBenchmark2022}    & Ectsum         & 2022 & 2,425     & ECTs              & English               & https://github.com/rajdeep345/ECTSum                            \\
\cite{bahrainianNEWTSCorpusNews2022}    & NEWTS          & 2022 & 312,084   & News              & English               & https://github.com/OpenNMS/newts                                \\
\cite{chen-etal-2023-toward}    & MIMIC-RRS      & 2022 & 207,782   & Radiology report  & English               & https://github.com/jbdel/vilmedic                               \\
\cite{clarkSEAHORSEMultilingualMultifaceted2023}    & SEAHORSE       & 2023 & 96,000    & Multifaceted      & Multilingual          & https://goo.gle/seahorse                                        \\
\cite{huMeetingBankBenchmarkDataset2023}    & MeetingBank    & 2023 & 6,892     & Meeting           & English               & https://meetingbank.github.io/                                  \\
\cite{linVideoXumCrossmodalVisual2024}    & Videoxum       & 2023 & 140,000   & Multimodal        & English               & https://github.com/jylins/videoxum                   \\ \hline\hline
\end{tabular}
}
    }
    \end{center}
\vspace{-2em}
\end{table*}

As summarized in Tables \ref{table:datasets} and \ref{table:datasets-ex}, current summarization datasets offer several strengths: they include large-scale datasets for training deep neural networks and smaller, refined datasets for evaluation, are predominantly open source and easily accessible, and primarily focus on the news domain. However, there is a gap in high-quality datasets for other domains, limiting research and practical applications in specialized areas such as financial earnings releases \cite{el-kassas2021}. To address this need, we propose methodologies for constructing customized datasets, which are detailed in the following section.

\subsection{Techniques to Build New Datasets}
Building new summarization datasets involves two main steps: 1) crawling or fetching texts, and 2) obtaining summaries. While manual annotation is reliable, it is time-consuming and labor-intensive\cite{harman2004effects,fabbri2019,yasunaga2019}. Recently, automatic annotation techniques, including rule-based and LLM-based methods, have gained popularity for balancing accuracy and efficiency.

\textbf{Rule-based annotation}:
Rule-based annotation uses specific portions of the text, such as titles, headlines\cite{hermann2015teaching}, or the first few sentences (e.g., LEAD-3\cite{zhu2020make}), as summaries. News articles, structured in the journalistic style, often start with key information, making their opening sentences suitable for summaries\cite{narayan2018a}. Similarly, academic papers have abstract sections that naturally serve as summaries\cite{cohan2018,luMultiXScienceLargescaleDataset2020a}. However, rule-based summaries can be imprecise and overly condensed, and texts in other domains often lack a structured format, making rule-based annotation challenging. In such cases, leveraging LLMs alongside manual efforts offers greater efficiency.

\textbf{LLM-based Data Generation}: 
LLMs are effective for generating summaries. A straightforward approach is to input original texts along with carefully designed prompts, then fine-tune the model to better grasp the summarization task and produce high-quality outputs (see Section \ref{model:LLM}). Some studies focus on LLM-based annotation methods. For example, \cite{fengLanguageModelAnnotator2021} used DialoGPT\cite{zhang2019dialogpt} to transform annotation into tasks like keyword extraction and redundancy detection, improving informativeness and relevance. \cite{chintaguntaMedicallyAwareGPT32021} developed a GPT-3 algorithm for few-shot annotation of medical dialogues, outperforming human-trained models in accuracy and coherence. LLM-based data generation could also be incorporated with human annotators to further improve robustness. For instance, \cite{liu2022wanli} proposed a GPT-3-based scheme combining LLM generation with human evaluation for improved outcomes.

Generating summaries directly via LLMs can encounter hallucination issues, resulting in factually incorrect information\cite{ji2023survey}. Several studies have aimed to measure and mitigate such hallucinations. \cite{goyalAnnotatingModelingFinegrained2021} utilized synthetic and human-labeled data to train models for detecting word-, dependency-, and sentence-level factual errors in summarization. Their work showed that the best factuality detection models, including the sentence-factuality model\cite{kryscinski-etal-2020-evaluating} and the arc-factuality model\cite{goyal2020evaluating}, effectively identify non-factual tokens. Additionally, \cite{balachandran2022correcting} proposed improving factual consistency by generating challenging synthetic examples of non-factual summaries using infilling language models. 

\section{Text Pre-processing}\label{pre-process}
After collecting the data, the next step in the process is pre-processing. Pre-processing is the process of transforming raw text into structured format data. This section describes common methods and powerful tools.

\subsection{Pre-processing methods}
\textbf{Noise Removal}: eliminates unnecessary parts of the input text, such as HTML tags in crawled text, extra spaces, blank lines, unknown symbols, and gibberish, etc. In earlier methods, it was necessary to remove stop words from the text \cite{Vodolazova2013TheRO}. Stop words, commonly occurring words in the text such as articles, pronouns, prepositions, auxiliary verbs, and determiners, were deleted using an artificially designed stop-words file because they were not useful for the analyses \cite{Saif2014OnSF} and had no significant impact on the selection of summary results.

\textbf{Part-Of-Speech (POS)}: involves the assignment of POS tags, such as verbs, nouns, adjectives, etc., to each word in a sentence. This categorization helps in identifying words based on their syntactic roles and context within the sentence structure \cite{DBLP:journals/aim/Charniak97}.

\textbf{Stemming}: is the process of converting words with the same root or stem to a basic form by eliminating variable endings like ``es'' and ``ed'' \cite{Nuzumlali2014AnalyzingSA,Galiotou2013OnTE}. This process was initially employed in open vocabulary text mining, for similar reasons as removing stop words: to reduce computation time and enhance recall in information retrieval \cite{hickman2022}. \cite{TorresMoreno2012BeyondSA} proposed reducing each word to its initial letters, a method referred to as Ultra-stemming, which demonstrated significant improvements.

\textbf{Sentence Segmentation}: splits texts into sentences. The simplest method involves using end markers such as ``.'', ``?'', or ``!'', but it is prone to a lot of interference, including abbreviations like ``e.g.'' or ``i.e.''. To address this issue, simple heuristics and regular expressions are employed \cite{gupta2012multi}. Sentence segmentation is typically necessary for processing long texts \cite{Moro2022SemanticSF} or for specific model designs, such as using sentences as nodes in a graph \cite{Wang2021UnsupervisedGL}.

\textbf{Word Tokenization}: divides words into subwords. Byte Pair Encoding (BPE) \cite{Sennrich2015NeuralMT} stands out as a simple and highly effective method for subword segmentation. It recursively maps common byte pairs to new ones and subsequently reconstructs the original text using the mapping table. Wordpiece \cite{Wu2016GooglesNM} selects the new word unit from all possible options, choosing the one that maximally enhances the likelihood on the training data when incorporated into the model.

\subsection{Pre-process Toolkits}
The primary tools for English pre-processing with Python are NLTK\footnote[1]{https://www.nltk.org/}, spaCy\footnote[2]{https://spacy.io/} and TextBlob\footnote[3]{https://github.com/sloria/TextBlob}. NLTK (Natural Language Toolkit) stands out as a leading platform for developing Python programs that handle human language data. It offers a suite of text processing libraries covering classification, tokenization, stemming, tagging, parsing, and semantic reasoning. NLTK also provides wrappers for industrial-strength NLP libraries and maintains an active discussion forum. spaCy, a high-performance NLP library optimized for production use, complements these tools by offering efficient processing pipelines and pre-trained models for English and other languages. Its robust features, such as named entity recognition, dependency parsing, and sentence splitting, make it particularly valuable for tasks like summarization, where structured linguistic analysis is critical for extracting key information. TextBlob, built on NLTK, offers a simplified API for various NLP tasks, including part-of-speech tagging, noun phrase extraction, sentiment analysis, classification, translation, and more. While TextBlob is application-oriented and user-friendly, it sacrifices some flexibility compared to NLTK. For other languages, Jieba\footnote[4]{https://github.com/LiveMirror/jieba} and HanLP\footnote[5]{https://github.com/hankcs/HanLP} provide effective solutions for the Chinese language. Jieba is a relatively lightweight tool with its primary function being word segmentation, capable of fulfilling most Chinese word splitting needs. On the other hand, HanLP\cite{he-choi-2021-stem} serves as an NLP toolkit built on PyTorch and TensorFlow, contributing to advancing state-of-the-art deep learning techniques in both academia and industry. It has demonstrated notable results in entity segmentation.

Furthermore, diverse models require distinct pre-processing methods as a prerequisite. In earlier models, it was imperative to design and implement an extensive set of rules to eliminate text that posed processing challenges for the model. Technological advancements have led to a noticeable reduction in the labor intensity associated with pre-processing. However, the need for pre-processing remains an indispensable component of the process, albeit with diminished intensity.

\begin{table*}[h!t]
\centering
\caption{Pros and cons of the conventional methods.}
\label{tab:summarization_methods}
\setlength{\tabcolsep}{6pt}
\begin{center}
    \setlength{\tabcolsep}{2mm}{
    \resizebox{\textwidth}{!}{
\begin{tabularx}{\textwidth}{llXXl}
\hline\hline
& \textbf{Method} & \textbf{Pros} & \textbf{Cons} & \textbf{Notable Works} \\
\hline
\multirow[t]{5}{*}{\makecell{\\Extractive\\summarization}} 
& Statistical-based & Low resource usage, simple implementation. & Prone to irrelevant text and high-scoring unimportant sentences.
 & \cite{mihalcea2004,erkan2004,gillick2009} \\
& Cluster-based & Improves coherence, reduces unrelated text interference. & May fail to identify semantically equivalent sentences. & \cite{gunaratna2015,zhang2015,haider2020automatic} \\
& Topic-based & Captures semantic relationships, reduces dimensionality. & Sensitive to parameters and requires preprocessing. & \cite{yeh2005,haghighi2009exploring,10.1145/1526709.1526728,haghighi2009exploring} \\
& ML-based & Optimizes selection using labeled data. & Dependent on high-quality data, costly to obtain. & \cite{10.1007/978-3-642-01818-3_23,OUYANG2011227} \\
& DL-based & Provides accurate sentence construction and improved classification. & Computationally expensive and resource-intensive. & \cite{abdel2022performance,xie2022pre} \\
\hline
\multirow[t]{5}{*}{\makecell{\\Abstractive\\summarization}}
& Tree-based & Creates organized, concise summaries. & May miss contextual relationships in long contexts.
 & \cite{9463007,10.1145/3510003.3510224} \\
& Graph-based & Handles complex relationships effectively. & High complexity relationship are brings high complexity in implementation. & \cite{10.5555/1873781.1873820,liu2018abstractivesummarizationusingsemantic,10083216} \\
& Rule-based & Precise control over information extraction. & Labor-intensive and inflexible for new content. & \cite{gupta2019,genest-lapalme-2012-fully} \\
& DL-based & Generates coherent, advanced summaries. & Face challenges with long sequences, computational efficiency, and consistency of output quality. & \cite{nallapati2017,Hanunggul2019TheIO,grail2021globalizing,wang2020friendly} \\
& Pre-trained & Efficient fine-tuning, strong generalization. & Dependency on data, computation cost, domain specific limitation  & \cite{10.5555/3524938.3525989,Goodwin2020FlightOT,pagnoni2022socratic} \\
\hline
\multirow[t]{2}{*}{\makecell{\\Hybrid\\summarization}} 
& E2SA & Enhances readability, preserves key information & High complexity, computation cost. & \cite{LLORET2013164,6968629,10.1007/978-981-10-5687-1_56} \\
& E2A & Factual accuracy, domain adaptability. & Training complexity, limited abstractive capability  & \cite{8029339} \\
\hline\hline
\end{tabularx}
}
    }
    \end{center}
\vspace{-2em}
\end{table*}

\section{Conventional ATS Methods}
\label{sec:conventional_ats}
As shown in section \ref{sec:cat}, conventional ATS methods are typically classified into extractive, abstractive, and hybrid approaches. 

\subsection{Extractive Summarization}
Extractive summarization selects sentences from the original text through three main steps: calculating sentence importance, sorting sentences based on their importance, and selecting the top-k sentences to form the summary. Sentence importance can be calculated using unsupervised or supervised methods. \textbf{Extractive ATS based on unsupervised approaches} relies on algorithms such as statistical, clustering, and topic-based techniques, offering high performance with low resource requirements, which will be introduced in detail below:

\textbf{Statistical-based Models}: 
Statistical-based models define importance as ``most frequent'' or ``most likely to occur'' \cite{gupta2010a}, such as selecting sentences with the highest word frequency. Methods like TextRank \cite{mihalcea2004} extract keywords and sentences by estimating similarity between phrases based on shared lexical tokens, while LexRank \cite{erkan2004} uses a graph-based approach with intra-sentence cosine similarity to determine sentence importance. Another method \cite{gillick2009} scores sentences based on the number of concepts they contain. These models are computationally efficient and require no extra linguistic knowledge, making them widely used. However, they are prone to interference from irrelevant words, leading to high-scoring but unimportant sentences.

\textbf{Cluster-based Models}: 
Cluster-based models extract summaries by identifying central sentences that represent the important information within a cluster. \cite{gunaratna2015} proposed clustering at the word level, selecting sentences containing the most central words as the summary. \cite{zhang2015} suggested that sentences with more similar counterparts are more representative, converting sentences into vectors, clustering, and selecting those closest to the cluster center. \cite{haider2020automatic} utilized word2vec for feature extraction and K-Means for summarization. These models improve coherence and reduce irrelevant text interference but may struggle to detect semantically equivalent sentences.

\textbf{Topic-based Models}: 
Topic-based models assume that each document consists of a mix of topics, with each topic represented by a set of words. Text is often represented as a topic-word matrix for summarization. Latent Semantic Analysis (LSA) \cite{Landauer2008LatentSA} captures semantics from word co-occurrences, with \cite{yeh2005} using a word-sentence matrix for filtering sentences into summaries. Latent Dirichlet Allocation (LDA) \cite{blei2003latent} identifies latent topics, with \cite{haghighi2009exploring} using it to model topic hierarchies for state-of-the-art summarization. LSA and LDA capture deeper semantic relationships, reducing data dimensionality for efficient processing. However, their performance depends heavily on parameters like the number of topics, requiring careful tuning and limiting generalizability across datasets.

\textbf{Extractive ATS based on supervised learning} uses a classifier to categorize sentences as summaries or non-summaries. The process involves representing sentences, classifying them, and selecting those labeled as summaries. Classifiers from other tasks can be adapted for this purpose, with options including machine-learning-based (ML) and Pre-trained-based methods. While supervised approaches improve sentence representation and extraction accuracy using labeled data, they incur additional costs for annotation.

\textbf{ML-based Models}: 
Machine learning ML-based ATS leverages labeled data to optimize sentence selection, with models like SVM and regression enhancing summarization. \cite{10.1007/978-3-642-01818-3_23} proposed an SVM-based ensemble model using Cross-Validation Committees to improve multi-document summarization by correcting classifier errors through ensemble outputs. \cite{OUYANG2011227} applied Support Vector Regression (SVR) to query-focused summarization, estimating sentence importance with features and pseudo-training data derived from human summaries. However, ML-based methods depend on high-quality labeled data, which is costly to produce, leading many studies to explore deep learning for richer information extraction.

\textbf{Pre-trained-based Models}:
Pre-trained-based models offer rich textual representations that enhance sentence vector construction and improve summarization accuracy. \cite{abdel2022performance} introduced SqueezeBERTSum, fine-tuned with a SqueezeBERT encoder variant, while \cite{xie2022pre} explored generative and discriminative techniques to fuse domain knowledge into knowledge adapters, integrating them into pre-trained models for efficient fine-tuning. However, training these models is computationally intensive, requiring significant hardware resources.

\textbf{Pros and Cons:} Extractive summarization models excel at capturing precise terminologies and require less training data, making them accurate, cost-effective, and efficient. However, they differ from human-generated summaries in expressive quality, often producing outputs with redundancy, excessive length, or contextual inconsistencies, lacking the nuance of human-crafted summaries.

\subsection{Abstractive Summarization}
Abstractive summarization models generate summaries by producing sentences distinct from the original text. This can be achieved through text structuring and combining or via generative models trained to predict the next token. Mainstream approaches of abstractive summarization are discussed as the following:

\textbf{Tree-based Models}:
Tree-based models use syntactic trees to structure input text, identify key sentences, and integrate them into coherent summaries. \cite{9463007} proposed BASTS, which uses a dominator tree to split ASTs into blocks, modeled with Tree-LSTMs for improved code summarization. \cite{10.1145/3510003.3510224} introduced AST-Trans, leveraging ancestor-descendant and sibling relationships to apply tree-structured attention for efficient encoding. These methods reduce redundancy but may miss important semantic connections by not fully considering broader context.

\textbf{Graph-based Models}:
Graph structures represent complex relationships between elements, enabling improved information flow in summarization tasks. \cite{10.5555/1873781.1873820} introduced ``Opinosis,'' a graph-based abstractive summarizer with nodes representing words and edges reflecting sentence structure. \cite{liu2018abstractivesummarizationusingsemantic} proposed a framework using Abstract Meaning Representation (AMR) graphs, transforming source text into condensed semantic graphs for summary generation. \cite{10083216} developed GMQS, a graph-enhanced multihop summarizer, using semantic relation graphs for better multirelational aggregation. While graph-based methods offer flexibility, they are harder to implement and optimize, with less interpretable outputs than tree-based approaches.

\begin{sidewaystable*}[]
\centering
\caption{Table of LLM-based ATS literature, distinguishing between works with over 15 citations and those with fewer citations but published. Categories include the base model used (BaseModel), the methodology employed (Methodology), the domain of the study (Domain), and the metrics used to evaluate the ATS system (Metric).}
\label{table:LLM}
\begin{center}
    \setlength{\tabcolsep}{2.5mm}{
    \resizebox{\textwidth}{!}{
\begin{tabular}{|cc|ccccc|ccc|ccccc|ccccc|}
\hline
\multicolumn{2}{|c|}{}                                                                       & \multicolumn{5}{c|}{\cellcolor[HTML]{CBCEFB}BaseModel}                                                                                                                                                                                                                       & \multicolumn{3}{c|}{\cellcolor[HTML]{FFCCC9}Methodology}                                                                                               & \multicolumn{5}{c|}{\cellcolor[HTML]{98FB98}Domain}                                                                                                                                                                                                                         & \multicolumn{5}{c|}{\cellcolor[HTML]{C0C0C0}Metric}                                                                                                                                                                                                                               \\ \hline
\multicolumn{1}{|c|}{\rotatebox{-90}{Ref.}}                          & \rotatebox{-90}{Year} & \multicolumn{1}{c|}{\rotatebox{-90}{GPT-3}}             & \multicolumn{1}{c|}{\rotatebox{-90}{GPT-3.5}}           & \multicolumn{1}{c|}{\rotatebox{-90}{GPT-4(v)}}          & \multicolumn{1}{c|}{\rotatebox{-90}{LLaMA}}             & \rotatebox{-90}{Others}              & \multicolumn{1}{c|}{\rotatebox{-90}{Prompt}}            & \multicolumn{1}{c|}{\rotatebox{-90}{Fine-tune}}         & \rotatebox{-90}{Distillation}      & \multicolumn{1}{c|}{\rotatebox{-90}{General}}           & \multicolumn{1}{c|}{\rotatebox{-90}{Medical}}           & \multicolumn{1}{c|}{\rotatebox{-90}{Code}}              & \multicolumn{1}{c|}{\rotatebox{-90}{Dialogue}}          & \rotatebox{-90}{Others}             & \multicolumn{1}{c|}{\rotatebox{-90}{ROUGE}}             & \multicolumn{1}{c|}{\rotatebox{-90}{BLEU}}              & \multicolumn{1}{c|}{\rotatebox{-90}{METEOR}}            & \multicolumn{1}{c|}{\rotatebox{-90}{BERTScore}}         & \rotatebox{-90}{Others}                   \\ \hline
\multicolumn{1}{|c|}{\cite{maIterativeOptimizingFramework2024}}      & 2024                  & \multicolumn{1}{c|}{}                                   & \multicolumn{1}{c|}{\cellcolor[HTML]{CBCEFB}\checkmark} & \multicolumn{1}{c|}{}                                   & \multicolumn{1}{c|}{}                                   &                                      & \multicolumn{1}{c|}{\cellcolor[HTML]{FFCCC9}\checkmark} & \multicolumn{1}{c|}{}                                   &                                    & \multicolumn{1}{c|}{}                                   & \multicolumn{1}{c|}{\cellcolor[HTML]{98FB98}\checkmark} & \multicolumn{1}{c|}{}                                   & \multicolumn{1}{c|}{}                                   &                                     & \multicolumn{1}{c|}{\cellcolor[HTML]{C0C0C0}\checkmark} & \multicolumn{1}{c|}{}                                   & \multicolumn{1}{c|}{}                                   & \multicolumn{1}{c|}{\cellcolor[HTML]{C0C0C0}\checkmark} &                                           \\ \hline
\multicolumn{1}{|c|}{\cite{edgeLocalGlobalGraph2024}}                & 2024                  & \multicolumn{1}{c|}{}                                   & \multicolumn{1}{c|}{}                                   & \multicolumn{1}{c|}{}                                   & \multicolumn{1}{c|}{}                                   & \cellcolor[HTML]{CBCEFB}LLM          & \multicolumn{1}{c|}{\cellcolor[HTML]{FFCCC9}\checkmark} & \multicolumn{1}{c|}{\cellcolor[HTML]{FFCCC9}\checkmark} &                                    & \multicolumn{1}{c|}{\cellcolor[HTML]{98FB98}\checkmark} & \multicolumn{1}{c|}{}                                   & \multicolumn{1}{c|}{}                                   & \multicolumn{1}{c|}{}                                   &                                     & \multicolumn{1}{c|}{}                                   & \multicolumn{1}{c|}{}                                   & \multicolumn{1}{c|}{}                                   & \multicolumn{1}{c|}{}                                   & \cellcolor[HTML]{C0C0C0}LLM evaluator     \\ \hline
\multicolumn{1}{|c|}{\cite{ahmedAutomaticSemanticAugmentation2024}}  & 2024                  & \multicolumn{1}{c|}{}                                   & \multicolumn{1}{c|}{\cellcolor[HTML]{CBCEFB}\checkmark} & \multicolumn{1}{c|}{}                                   & \multicolumn{1}{c|}{}                                   &                                      & \multicolumn{1}{c|}{\cellcolor[HTML]{FFCCC9}\checkmark} & \multicolumn{1}{c|}{}                                   &                                    & \multicolumn{1}{c|}{}                                   & \multicolumn{1}{c|}{}                                   & \multicolumn{1}{c|}{\cellcolor[HTML]{98FB98}\checkmark} & \multicolumn{1}{c|}{}                                   &                                     & \multicolumn{1}{c|}{}                                   & \multicolumn{1}{c|}{\cellcolor[HTML]{C0C0C0}\checkmark} & \multicolumn{1}{c|}{}                                   & \multicolumn{1}{c|}{}                                   &                                           \\ \hline
\multicolumn{1}{|c|}{\cite{suDistilledGPTSource2024}}                & 2024                  & \multicolumn{1}{c|}{}                                   & \multicolumn{1}{c|}{\cellcolor[HTML]{CBCEFB}\checkmark} & \multicolumn{1}{c|}{}                                   & \multicolumn{1}{c|}{}                                   &                                      & \multicolumn{1}{c|}{}                                   & \multicolumn{1}{c|}{}                                   & \cellcolor[HTML]{FFCCC9}\checkmark & \multicolumn{1}{c|}{}                                   & \multicolumn{1}{c|}{}                                   & \multicolumn{1}{c|}{\cellcolor[HTML]{98FB98}\checkmark} & \multicolumn{1}{c|}{}                                   &                                     & \multicolumn{1}{c|}{}                                   & \multicolumn{1}{c|}{}                                   & \multicolumn{1}{c|}{\cellcolor[HTML]{C0C0C0}\checkmark} & \multicolumn{1}{c|}{}                                   &               \\ \hline
\multicolumn{1}{|c|}{\cite{chuangSPeCSoftPromptBased2023}}           & 2024                  & \multicolumn{1}{c|}{}                                   & \multicolumn{1}{c|}{\cellcolor[HTML]{CBCEFB}\checkmark} & \multicolumn{1}{c|}{}                                   & \multicolumn{1}{c|}{}                                   &                                      & \multicolumn{1}{c|}{\cellcolor[HTML]{FFCCC9}\checkmark} & \multicolumn{1}{c|}{}                                   &                                    & \multicolumn{1}{c|}{}                                   & \multicolumn{1}{c|}{\cellcolor[HTML]{98FB98}\checkmark} & \multicolumn{1}{c|}{}                                   & \multicolumn{1}{c|}{}                                   &                                     & \multicolumn{1}{c|}{\cellcolor[HTML]{C0C0C0}\checkmark} & \multicolumn{1}{c|}{}                                   & \multicolumn{1}{c|}{}                                   & \multicolumn{1}{c|}{}                                   &                                           \\ \hline
\multicolumn{1}{|c|}{\cite{ghoshCLIPSyntelCLIPLLM2023}}              & 2024                  & \multicolumn{1}{c|}{}                                   & \multicolumn{1}{c|}{\cellcolor[HTML]{CBCEFB}\checkmark} & \multicolumn{1}{c|}{}                                   & \multicolumn{1}{c|}{}                                   & \cellcolor[HTML]{CBCEFB}CLIP         & \multicolumn{1}{c|}{\cellcolor[HTML]{FFCCC9}\checkmark} & \multicolumn{1}{c|}{}                                   &                                    & \multicolumn{1}{c|}{}                                   & \multicolumn{1}{c|}{\cellcolor[HTML]{98FB98}\checkmark} & \multicolumn{1}{c|}{}                                   & \multicolumn{1}{c|}{}                                   &                                     & \multicolumn{1}{c|}{\cellcolor[HTML]{C0C0C0}\checkmark} & \multicolumn{1}{c|}{\cellcolor[HTML]{C0C0C0}\checkmark} & \multicolumn{1}{c|}{}                                   & \multicolumn{1}{c|}{\cellcolor[HTML]{C0C0C0}\checkmark} & \cellcolor[HTML]{C0C0C0}Human             \\ \hline
\multicolumn{1}{|c|}{\cite{10650513}}                                           & 2024                  & \multicolumn{1}{c|}{}                                   & \multicolumn{1}{c|}{}                                   & \multicolumn{1}{c|}{}                                   & \multicolumn{1}{c|}{}                                   & \cellcolor[HTML]{CBCEFB}Baichuan2-7B & \multicolumn{1}{c|}{}                                   & \multicolumn{1}{c|}{\cellcolor[HTML]{FFCCC9}\checkmark} &                                    & \multicolumn{1}{c|}{}                                   & \multicolumn{1}{c|}{}                                   & \multicolumn{1}{c|}{}                                   & \multicolumn{1}{c|}{\cellcolor[HTML]{98FB98}\checkmark} &                                     & \multicolumn{1}{c|}{\cellcolor[HTML]{C0C0C0}\checkmark} & \multicolumn{1}{c|}{\cellcolor[HTML]{C0C0C0}\checkmark} & \multicolumn{1}{c|}{}                                   & \multicolumn{1}{c|}{\cellcolor[HTML]{C0C0C0}\checkmark} &                                           \\ \hline
\multicolumn{1}{|c|}{\cite{jiang-etal-2024-trisum}}                                           & 2024                  & \multicolumn{1}{c|}{}                                   & \multicolumn{1}{c|}{\cellcolor[HTML]{CBCEFB}\checkmark} & \multicolumn{1}{c|}{}                                   & \multicolumn{1}{c|}{}                                   &                                      & \multicolumn{1}{c|}{}                                   & \multicolumn{1}{c|}{}                                   & \cellcolor[HTML]{FFCCC9}\checkmark & \multicolumn{1}{c|}{\cellcolor[HTML]{98FB98}\checkmark} & \multicolumn{1}{c|}{}                                   & \multicolumn{1}{c|}{}                                   & \multicolumn{1}{c|}{}                                   &                                     & \multicolumn{1}{c|}{\cellcolor[HTML]{C0C0C0}\checkmark} & \multicolumn{1}{c|}{}                                   & \multicolumn{1}{c|}{}                                   & \multicolumn{1}{c|}{\cellcolor[HTML]{C0C0C0}\checkmark} & \cellcolor[HTML]{C0C0C0}BARTScore         \\ \hline
\multicolumn{1}{|c|}{\cite{siledar-etal-2024-product}}                                           & 2024                  & \multicolumn{1}{c|}{}                                   & \multicolumn{1}{c|}{\cellcolor[HTML]{CBCEFB}\checkmark} & \multicolumn{1}{c|}{}                                   & \multicolumn{1}{c|}{}                                   &                                      & \multicolumn{1}{c|}{}                                   & \multicolumn{1}{c|}{}                                   & \cellcolor[HTML]{FFCCC9}\checkmark & \multicolumn{1}{c|}{}                                   & \multicolumn{1}{c|}{}                                   & \multicolumn{1}{c|}{}                                   & \multicolumn{1}{c|}{}                                   & \cellcolor[HTML]{98FB98}E-commerce  & \multicolumn{1}{c|}{\cellcolor[HTML]{C0C0C0}\checkmark} & \multicolumn{1}{c|}{}                                   & \multicolumn{1}{c|}{}                                   & \multicolumn{1}{c|}{}                                   &                                           \\ \hline
\multicolumn{1}{|c|}{\cite{wang-etal-2024-integrate}}                                           & 2024                  & \multicolumn{1}{c|}{}                                   & \multicolumn{1}{c|}{\cellcolor[HTML]{CBCEFB}\checkmark} & \multicolumn{1}{c|}{\cellcolor[HTML]{CBCEFB}\checkmark} & \multicolumn{1}{c|}{}                                   &                                      & \multicolumn{1}{c|}{\cellcolor[HTML]{FFCCC9}\checkmark} & \multicolumn{1}{c|}{}                                   &                                    & \multicolumn{1}{c|}{\cellcolor[HTML]{98FB98}\checkmark} & \multicolumn{1}{c|}{}                                   & \multicolumn{1}{c|}{}                                   & \multicolumn{1}{c|}{}                                   &                                     & \multicolumn{1}{c|}{\cellcolor[HTML]{C0C0C0}\checkmark} & \multicolumn{1}{c|}{}                                   & \multicolumn{1}{c|}{}                                   & \multicolumn{1}{c|}{\cellcolor[HTML]{C0C0C0}\checkmark} &                                           \\ \hline
\multicolumn{1}{|c|}{\cite{10.1145/3661167.3661210}}                                           & 2024                  & \multicolumn{1}{c|}{}                                   & \multicolumn{1}{c|}{}                                   & \multicolumn{1}{c|}{}                                   & \multicolumn{1}{c|}{\cellcolor[HTML]{CBCEFB}\checkmark} &                                      & \multicolumn{1}{c|}{}                                   & \multicolumn{1}{c|}{\cellcolor[HTML]{FFCCC9}\checkmark} &                                    & \multicolumn{1}{c|}{}                                   & \multicolumn{1}{c|}{}                                   & \multicolumn{1}{c|}{\cellcolor[HTML]{98FB98}\checkmark} & \multicolumn{1}{c|}{}                                   &                                     & \multicolumn{1}{c|}{\cellcolor[HTML]{C0C0C0}\checkmark} & \multicolumn{1}{c|}{\cellcolor[HTML]{C0C0C0}\checkmark} & \multicolumn{1}{c|}{\cellcolor[HTML]{C0C0C0}\checkmark} & \multicolumn{1}{c|}{}                                   &                                           \\ \hline
\multicolumn{1}{|c|}{\cite{10.1145/3625468.3652197}}                                           & 2024                  & \multicolumn{1}{c|}{}                                   & \multicolumn{1}{c|}{}                                   & \multicolumn{1}{c|}{\cellcolor[HTML]{CBCEFB}\checkmark} & \multicolumn{1}{c|}{}                                   &                                      & \multicolumn{1}{c|}{\cellcolor[HTML]{FFCCC9}\checkmark} & \multicolumn{1}{c|}{}                                   &                                    & \multicolumn{1}{c|}{}                                   & \multicolumn{1}{c|}{}                                   & \multicolumn{1}{c|}{}                                   & \multicolumn{1}{c|}{}                                   & \cellcolor[HTML]{98FB98}Soccer      & \multicolumn{1}{c|}{}                                   & \multicolumn{1}{c|}{}                                   & \multicolumn{1}{c|}{}                                   & \multicolumn{1}{c|}{}                                   &                                           \\ \hline
\multicolumn{1}{|c|}{\cite{li-etal-2024-improving-faithfulness}}                                           & 2024                  & \multicolumn{1}{c|}{}                                   & \multicolumn{1}{c|}{\cellcolor[HTML]{CBCEFB}\checkmark} & \multicolumn{1}{c|}{}                                   & \multicolumn{1}{c|}{\cellcolor[HTML]{CBCEFB}\checkmark} & \cellcolor[HTML]{CBCEFB}Claude-2     & \multicolumn{1}{c|}{\cellcolor[HTML]{FFCCC9}\checkmark} & \multicolumn{1}{c|}{}                                   &                                    & \multicolumn{1}{c|}{\cellcolor[HTML]{98FB98}\checkmark} & \multicolumn{1}{c|}{}                                   & \multicolumn{1}{c|}{}                                   & \multicolumn{1}{c|}{}                                   &                                     & \multicolumn{1}{c|}{\cellcolor[HTML]{C0C0C0}\checkmark} & \multicolumn{1}{c|}{}                                   & \multicolumn{1}{c|}{}                                   & \multicolumn{1}{c|}{\cellcolor[HTML]{C0C0C0}\checkmark} & \cellcolor[HTML]{C0C0C0}FactCC, SummaC    \\ \hline
\multicolumn{1}{|c|}{\cite{bhushan-etal-2024-unveiling}}                                           & 2024                  & \multicolumn{1}{c|}{}                                   & \multicolumn{1}{c|}{\cellcolor[HTML]{CBCEFB}\checkmark} & \multicolumn{1}{c|}{\cellcolor[HTML]{CBCEFB}\checkmark} & \multicolumn{1}{c|}{}                                   &                                      & \multicolumn{1}{c|}{\cellcolor[HTML]{FFCCC9}\checkmark} & \multicolumn{1}{c|}{}                                   &                                    & \multicolumn{1}{c|}{}                                   & \multicolumn{1}{c|}{}                                   & \multicolumn{1}{c|}{}                                   & \multicolumn{1}{c|}{}                                   & \cellcolor[HTML]{98FB98}Diagram     & \multicolumn{1}{c|}{\cellcolor[HTML]{C0C0C0}\checkmark} & \multicolumn{1}{c|}{\cellcolor[HTML]{C0C0C0}\checkmark} & \multicolumn{1}{c|}{}                                   & \multicolumn{1}{c|}{}                                   & \cellcolor[HTML]{C0C0C0}BLEURT, PPL       \\ \hline
\multicolumn{1}{|c|}{\cite{zhang-etal-2024-comprehensive}}                                           & 2024                  & \multicolumn{1}{c|}{}                                   & \multicolumn{1}{c|}{\cellcolor[HTML]{CBCEFB}\checkmark} & \multicolumn{1}{c|}{}                                   & \multicolumn{1}{c|}{}                                   &                                      & \multicolumn{1}{c|}{\cellcolor[HTML]{FFCCC9}\checkmark} & \multicolumn{1}{c|}{}                                   &                                    & \multicolumn{1}{c|}{}                                   & \multicolumn{1}{c|}{}                                   & \multicolumn{1}{c|}{}                                   & \multicolumn{1}{c|}{}                                   & \cellcolor[HTML]{98FB98}Comment     & \multicolumn{1}{c|}{\cellcolor[HTML]{C0C0C0}\checkmark} & \multicolumn{1}{c|}{}                                   & \multicolumn{1}{c|}{}                                   & \multicolumn{1}{c|}{}                                   &                                           \\ \hline
\multicolumn{1}{|c|}{\cite{jung-etal-2024-impossible}}                                           & 2024                  & \multicolumn{1}{c|}{}                                   & \multicolumn{1}{c|}{\cellcolor[HTML]{CBCEFB}\checkmark} & \multicolumn{1}{c|}{}                                   & \multicolumn{1}{c|}{}                                   &                                      & \multicolumn{1}{c|}{}                                   & \multicolumn{1}{c|}{}                                   & \cellcolor[HTML]{FFCCC9}\checkmark & \multicolumn{1}{c|}{\cellcolor[HTML]{98FB98}\checkmark} & \multicolumn{1}{c|}{}                                   & \multicolumn{1}{c|}{}                                   & \multicolumn{1}{c|}{}                                   &                                     & \multicolumn{1}{c|}{\cellcolor[HTML]{C0C0C0}\checkmark} & \multicolumn{1}{c|}{\cellcolor[HTML]{C0C0C0}\checkmark} & \multicolumn{1}{c|}{}                                   & \multicolumn{1}{c|}{}                                   & \cellcolor[HTML]{C0C0C0}BERT-iBLEU, Human \\ \hline
\multicolumn{1}{|c|}{\cite{zhengChatGPTChemistryAssistant}}          & 2023                  & \multicolumn{1}{c|}{}                                   & \multicolumn{1}{c|}{\cellcolor[HTML]{CBCEFB}\checkmark} & \multicolumn{1}{c|}{}                                   & \multicolumn{1}{c|}{}                                   &                                      & \multicolumn{1}{c|}{\cellcolor[HTML]{FFCCC9}\checkmark} & \multicolumn{1}{c|}{}                                   &                                    & \multicolumn{1}{c|}{}                                   & \multicolumn{1}{c|}{}                                   & \multicolumn{1}{c|}{}                                   & \multicolumn{1}{c|}{}                                   & \cellcolor[HTML]{98FB98}Chemistry   & \multicolumn{1}{c|}{}                                   & \multicolumn{1}{c|}{}                                   & \multicolumn{1}{c|}{}                                   & \multicolumn{1}{c|}{}                                   & \cellcolor[HTML]{C0C0C0}P, R, F1          \\ \hline
\multicolumn{1}{|c|}{\cite{thawakar-etal-2024-xraygpt}}      & 2023                  & \multicolumn{1}{c|}{}                                   & \multicolumn{1}{c|}{}                                   & \multicolumn{1}{c|}{}                                   & \multicolumn{1}{c|}{}                                   & \cellcolor[HTML]{CBCEFB}Vicuna       & \multicolumn{1}{c|}{\cellcolor[HTML]{FFCCC9}\checkmark} & \multicolumn{1}{c|}{\cellcolor[HTML]{FFCCC9}\checkmark} &                                    & \multicolumn{1}{c|}{}                                   & \multicolumn{1}{c|}{\cellcolor[HTML]{98FB98}\checkmark} & \multicolumn{1}{c|}{}                                   & \multicolumn{1}{c|}{}                                   &                                     & \multicolumn{1}{c|}{\cellcolor[HTML]{C0C0C0}\checkmark} & \multicolumn{1}{c|}{}                                   & \multicolumn{1}{c|}{}                                   & \multicolumn{1}{c|}{}                                   &                                           \\ \hline
\multicolumn{1}{|c|}{\cite{liuTCRALLMTokenCompression2023}}          & 2023                  & \multicolumn{1}{c|}{}                                   & \multicolumn{1}{c|}{\cellcolor[HTML]{CBCEFB}\checkmark} & \multicolumn{1}{c|}{}                                   & \multicolumn{1}{c|}{}                                   &                                      & \multicolumn{1}{c|}{}                                   & \multicolumn{1}{c|}{\cellcolor[HTML]{FFCCC9}\checkmark} &                                    & \multicolumn{1}{c|}{\cellcolor[HTML]{98FB98}\checkmark} & \multicolumn{1}{c|}{}                                   & \multicolumn{1}{c|}{}                                   & \multicolumn{1}{c|}{}                                   &                                     & \multicolumn{1}{c|}{}                                   & \multicolumn{1}{c|}{}                                   & \multicolumn{1}{c|}{}                                   & \multicolumn{1}{c|}{}                                   & \cellcolor[HTML]{C0C0C0}Accuracy          \\ \hline
\multicolumn{1}{|c|}{\cite{xuRECOMPImprovingRetrievalAugmented2023}} & 2023                  & \multicolumn{1}{c|}{}                                   & \multicolumn{1}{c|}{\cellcolor[HTML]{CBCEFB}\checkmark} & \multicolumn{1}{c|}{}                                   & \multicolumn{1}{c|}{}                                   &                                      & \multicolumn{1}{c|}{\cellcolor[HTML]{FFCCC9}\checkmark} & \multicolumn{1}{c|}{}                                   & \cellcolor[HTML]{FFCCC9}\checkmark & \multicolumn{1}{c|}{\cellcolor[HTML]{98FB98}\checkmark} & \multicolumn{1}{c|}{}                                   & \multicolumn{1}{c|}{}                                   & \multicolumn{1}{c|}{}                                   &                                     & \multicolumn{1}{c|}{}                                   & \multicolumn{1}{c|}{}                                   & \multicolumn{1}{c|}{}                                   & \multicolumn{1}{c|}{}                                   & \cellcolor[HTML]{C0C0C0}PPL               \\ \hline
\multicolumn{1}{|c|}{\cite{wang2023a}}                  & 2023                  & \multicolumn{1}{c|}{\cellcolor[HTML]{CBCEFB}\checkmark} & \multicolumn{1}{c|}{}                                   & \multicolumn{1}{c|}{}                                   & \multicolumn{1}{c|}{}                                   &                                      & \multicolumn{1}{c|}{\cellcolor[HTML]{FFCCC9}\checkmark} & \multicolumn{1}{c|}{\cellcolor[HTML]{FFCCC9}\checkmark} &                                    & \multicolumn{1}{c|}{}                                   & \multicolumn{1}{c|}{}                                   & \multicolumn{1}{c|}{}                                   & \multicolumn{1}{c|}{}                                   & \cellcolor[HTML]{98FB98}News        & \multicolumn{1}{c|}{\cellcolor[HTML]{C0C0C0}\checkmark} & \multicolumn{1}{c|}{}                                   & \multicolumn{1}{c|}{}                                   & \multicolumn{1}{c|}{\cellcolor[HTML]{C0C0C0}\checkmark} &                                           \\ \hline
\multicolumn{1}{|c|}{\cite{jonesTeachingLanguageModels2023}}         & 2023                  & \multicolumn{1}{c|}{}                                   & \multicolumn{1}{c|}{}                                   & \multicolumn{1}{c|}{}                                   & \multicolumn{1}{c|}{}                                   & \cellcolor[HTML]{CBCEFB}Vicuna, Orea & \multicolumn{1}{c|}{}                                   & \multicolumn{1}{c|}{\cellcolor[HTML]{FFCCC9}\checkmark} & \cellcolor[HTML]{FFCCC9}\checkmark & \multicolumn{1}{c|}{\cellcolor[HTML]{98FB98}\checkmark} & \multicolumn{1}{c|}{}                                   & \multicolumn{1}{c|}{}                                   & \multicolumn{1}{c|}{}                                   &                                     & \multicolumn{1}{c|}{\cellcolor[HTML]{C0C0C0}\checkmark} & \multicolumn{1}{c|}{\cellcolor[HTML]{C0C0C0}\checkmark} & \multicolumn{1}{c|}{}                                   & \multicolumn{1}{c|}{}                                   &      \\ \hline
\multicolumn{1}{|c|}{\cite{liSkillGPTRESTfulAPI2023}}                & 2023                  & \multicolumn{1}{c|}{}                                   & \multicolumn{1}{c|}{}                                   & \multicolumn{1}{c|}{}                                   & \multicolumn{1}{c|}{\cellcolor[HTML]{CBCEFB}\checkmark} &                                      & \multicolumn{1}{c|}{\cellcolor[HTML]{FFCCC9}\checkmark} & \multicolumn{1}{c|}{}                                   &                                    & \multicolumn{1}{c|}{\cellcolor[HTML]{98FB98}\checkmark} & \multicolumn{1}{c|}{}                                   & \multicolumn{1}{c|}{}                                   & \multicolumn{1}{c|}{}                                   &                                     & \multicolumn{1}{c|}{}                                   & \multicolumn{1}{c|}{}                                   & \multicolumn{1}{c|}{}                                   & \multicolumn{1}{c|}{}                                   & -                                         \\ \hline
\multicolumn{1}{|c|}{\cite{zhengBIMGPTPromptBasedVirtual}}           & 2023                  & \multicolumn{1}{c|}{\cellcolor[HTML]{CBCEFB}\checkmark} & \multicolumn{1}{c|}{}                                   & \multicolumn{1}{c|}{}                                   & \multicolumn{1}{c|}{}                                   &                                      & \multicolumn{1}{c|}{\cellcolor[HTML]{FFCCC9}\checkmark} & \multicolumn{1}{c|}{}                                   &                                    & \multicolumn{1}{c|}{\cellcolor[HTML]{98FB98}\checkmark} & \multicolumn{1}{c|}{}                                   & \multicolumn{1}{c|}{}                                   & \multicolumn{1}{c|}{}                                   &                                     & \multicolumn{1}{c|}{}                                   & \multicolumn{1}{c|}{}                                   & \multicolumn{1}{c|}{}                                   & \multicolumn{1}{c|}{}                                   & \cellcolor[HTML]{C0C0C0}Accuracy          \\ \hline
\multicolumn{1}{|c|}{\cite{mishra-etal-2023-llm}}                                           & 2023                  & \multicolumn{1}{c|}{}                                   & \multicolumn{1}{c|}{\cellcolor[HTML]{CBCEFB}\checkmark} & \multicolumn{1}{c|}{}                                   & \multicolumn{1}{c|}{}                                   &                                      & \multicolumn{1}{c|}{}                                   & \multicolumn{1}{c|}{}                                   & \cellcolor[HTML]{FFCCC9}\checkmark & \multicolumn{1}{c|}{}                                   & \multicolumn{1}{c|}{}                                   & \multicolumn{1}{c|}{}                                   & \multicolumn{1}{c|}{\cellcolor[HTML]{98FB98}\checkmark} &                                     & \multicolumn{1}{c|}{\cellcolor[HTML]{C0C0C0}\checkmark} & \multicolumn{1}{c|}{}                                   & \multicolumn{1}{c|}{}                                   & \multicolumn{1}{c|}{}                                   &                                           \\ \hline
\multicolumn{1}{|c|}{\cite{Xu2023ArgumentativeSE}}                                           & 2023                  & \multicolumn{1}{c|}{}                                   & \multicolumn{1}{c|}{\cellcolor[HTML]{CBCEFB}\checkmark} & \multicolumn{1}{c|}{\cellcolor[HTML]{CBCEFB}\checkmark} & \multicolumn{1}{c|}{}                                   &                                      & \multicolumn{1}{c|}{\cellcolor[HTML]{FFCCC9}\checkmark} & \multicolumn{1}{c|}{}                                   &                                    & \multicolumn{1}{c|}{}                                   & \multicolumn{1}{c|}{}                                   & \multicolumn{1}{c|}{}                                   & \multicolumn{1}{c|}{}                                   & \cellcolor[HTML]{98FB98}Legal       & \multicolumn{1}{c|}{\cellcolor[HTML]{C0C0C0}\checkmark} & \multicolumn{1}{c|}{\cellcolor[HTML]{C0C0C0}\checkmark} & \multicolumn{1}{c|}{\cellcolor[HTML]{C0C0C0}\checkmark} & \multicolumn{1}{c|}{\cellcolor[HTML]{C0C0C0}\checkmark} &                                           \\ \hline
\multicolumn{1}{|c|}{\cite{xu-etal-2023-inheritsumm}}                                           & 2023                  & \multicolumn{1}{c|}{}                                   & \multicolumn{1}{c|}{\cellcolor[HTML]{CBCEFB}\checkmark} & \multicolumn{1}{c|}{}                                   & \multicolumn{1}{c|}{}                                   &                                      & \multicolumn{1}{c|}{}                                   & \multicolumn{1}{c|}{}                                   & \cellcolor[HTML]{FFCCC9}\checkmark & \multicolumn{1}{c|}{\cellcolor[HTML]{98FB98}\checkmark} & \multicolumn{1}{c|}{}                                   & \multicolumn{1}{c|}{}                                   & \multicolumn{1}{c|}{}                                   &                                     & \multicolumn{1}{c|}{\cellcolor[HTML]{C0C0C0}\checkmark} & \multicolumn{1}{c|}{}                                   & \multicolumn{1}{c|}{}                                   & \multicolumn{1}{c|}{}                                   &                                           \\ \hline
\multicolumn{1}{|c|}{\cite{pham-etal-2023-select}}                                           & 2023                  & \multicolumn{1}{c|}{}                                   & \multicolumn{1}{c|}{\cellcolor[HTML]{CBCEFB}\checkmark} & \multicolumn{1}{c|}{}                                   & \multicolumn{1}{c|}{}                                   &                                      & \multicolumn{1}{c|}{}                                   & \multicolumn{1}{c|}{}                                   & \cellcolor[HTML]{FFCCC9}\checkmark & \multicolumn{1}{c|}{}                                   & \multicolumn{1}{c|}{}                                   & \multicolumn{1}{c|}{}                                   & \multicolumn{1}{c|}{\cellcolor[HTML]{98FB98}\checkmark} &                                     & \multicolumn{1}{c|}{\cellcolor[HTML]{C0C0C0}\checkmark} & \multicolumn{1}{c|}{}                                   & \multicolumn{1}{c|}{}                                   & \multicolumn{1}{c|}{}                                   &                                           \\ \hline
\multicolumn{1}{|c|}{\cite{sclar-etal-2022-referee}}                                           & 2022                  & \multicolumn{1}{c|}{\cellcolor[HTML]{CBCEFB}\checkmark} & \multicolumn{1}{c|}{}                                   & \multicolumn{1}{c|}{}                                   & \multicolumn{1}{c|}{}                                   &                                      & \multicolumn{1}{c|}{}                                   & \multicolumn{1}{c|}{}                                   & \cellcolor[HTML]{FFCCC9}\checkmark & \multicolumn{1}{c|}{}                                   & \multicolumn{1}{c|}{}                                   & \multicolumn{1}{c|}{}                                   & \multicolumn{1}{c|}{}                                   & \cellcolor[HTML]{98FB98}Reference   & \multicolumn{1}{c|}{\cellcolor[HTML]{C0C0C0}\checkmark} & \multicolumn{1}{c|}{}                                   & \multicolumn{1}{c|}{}                                   & \multicolumn{1}{c|}{\cellcolor[HTML]{C0C0C0}\checkmark} &                                           \\ \hline
\multicolumn{1}{|c|}{\cite{asi-etal-2022-end}}                                           & 2022                  & \multicolumn{1}{c|}{\cellcolor[HTML]{CBCEFB}\checkmark} & \multicolumn{1}{c|}{}                                   & \multicolumn{1}{c|}{}                                   & \multicolumn{1}{c|}{}                                   &                                      & \multicolumn{1}{c|}{}                                   & \multicolumn{1}{c|}{}                                   & \cellcolor[HTML]{FFCCC9}\checkmark & \multicolumn{1}{c|}{}                                   & \multicolumn{1}{c|}{}                                   & \multicolumn{1}{c|}{}                                   & \multicolumn{1}{c|}{}                                   & \cellcolor[HTML]{98FB98}Sales calls & \multicolumn{1}{c|}{\cellcolor[HTML]{C0C0C0}\checkmark} & \multicolumn{1}{c|}{}                                   & \multicolumn{1}{c|}{\cellcolor[HTML]{C0C0C0}\checkmark} & \multicolumn{1}{c|}{}                                   &                                           \\ \hline
\multicolumn{1}{|c|}{\cite{xu-etal-2022-narrate}}                                           & 2022                  & \multicolumn{1}{c|}{}                                   & \multicolumn{1}{c|}{}                                   & \multicolumn{1}{c|}{}                                   & \multicolumn{1}{c|}{}                                   & \cellcolor[HTML]{CBCEFB}InstructGPT  & \multicolumn{1}{c|}{\cellcolor[HTML]{FFCCC9}\checkmark} & \multicolumn{1}{c|}{}                                   &                                    & \multicolumn{1}{c|}{}                                   & \multicolumn{1}{c|}{}                                   & \multicolumn{1}{c|}{}                                   & \multicolumn{1}{c|}{\cellcolor[HTML]{98FB98}\checkmark} &                                     & \multicolumn{1}{c|}{\cellcolor[HTML]{C0C0C0}\checkmark} & \multicolumn{1}{c|}{}                                   & \multicolumn{1}{c|}{}                                   & \multicolumn{1}{c|}{}                                   & \cellcolor[HTML]{C0C0C0}FactCC            \\ \hline
\end{tabular}
}
    }
    \end{center}
\end{sidewaystable*}

\begin{table*}[]
\caption{Pros and cons of the LLM-based methods.}
\label{tab:llm_methods}
\begin{center}
    \setlength{\tabcolsep}{2mm}{
    \resizebox{\textwidth}{!}{
\begin{tabularx}{\textwidth}{llm{2cm}m{4cm}m{4cm}}
\hline\hline
\multicolumn{1}{l}{}                & Method                         & Ref.                                                                                                                                                                    & Pros                                                       & Cons                                                                         \\ \hline
\multirow[t]{3}{*}{Prompt Engineering} & Template Engineering           & \cite{liu2023,Zhao2023ASO,shin2021constrained,narayan2021,zhou2023}                                                                               & Predictable output, easy to control structure.             & Limited flexibility, may not capture complex relationships.                  \\
                                    & Chain of Thought               & \cite{wang2023a,adams2023,zhang-etal-2024-comprehensive}                                                                                                                & Can improve reasoning capabilities by breaking down tasks. & Increases complexity and computational cost.                                 \\
                                    & Agent Interactions             & \cite{xiao2023,asi2022end,de-raedt-etal-2023-idas,Li2024ChatCiteLA}                                                                                                     & Enables dynamic and adaptive behavior through interaction. & Requires sophisticated coordination mechanisms; harder to scale.             \\ \hline
RAG & - & \cite{edgeLocalGlobalGraph2024,parvez-etal-2021-retrieval-augmented,zhou2023,Manathunga2023RetrievalAG,Liu2024TowardsAR}                                                & Can use existing knowledge bases effectively.              & Reliability depends on retrieval accuracy; may introduce bias.               \\ \hline
\multirow[t]{2}{*}{Fine-tuning}        & Internal parameters            & \cite{heZCodePretrainedLanguage,Veen2023RadAdaptRR,you-etal-2024-uiuc,bai2022traininghelpfulharmlessassistant,huang2024nimplementationdetailsrlhf,nath2024leveragingdomainknowledgeefficient}                                                                                                                     & Tailors model to specific tasks or domains.                & Destroying the original structure and capabilities of LLM.                   \\
                                    & External adapters              & \cite{bravzinskas2022efficient,xuSequenceLevelContrastive2022,guAssembleFoundationModels2022,hu2021loralowrankadaptationlarge,nawander-nerella-2025-datahacks,mullick2024leveragingpowerllmsfinetuning}                                                                           & Less risk of overfitting; can adapt quickly to new tasks.  & May not transfer improvements as effectively as full fine-tuning.            \\ \hline
\multirow[t]{1}{*}{Knowledge distillation} & Offline Distillation                              & \cite{suDistilledGPTSource2024,jiang-etal-2024-trisum,10.1145/3675167,jung2024impossibledistillationlowqualitymodel,xu-etal-2023-inheritsumm,pham-etal-2023-select,sclar-etal-2022-referee,asi2022end} & Reduces model size while maintaining performance.          & Requires a well-trained teacher model; can lose nuances during distillation. \\ \hline\hline
\end{tabularx}
}
    }
\end{center}
\vspace{-2em}
\end{table*}


\textbf{The Generative Summarization Methods} treat summarization as a sequence-to-sequence generation task, leveraging advancements in pre-trained models and large-scale corpora to produce high-quality summaries with enriched linguistic understanding. The methods fall into this category will be introduced in detail as the following:

\textbf{DL-based Models}:
Deep-learning-based ATS methods, including RNN, LSTM, GRU, and Transformer models, are well-suited for sequence-to-sequence tasks, enabling effective text processing and generation. RNN-based models, such as SummaRuNNer \cite{nallapati2017}, excel in extractive summarization with interpretable outputs, while conditional RNNs \cite{chopra2016} leverage convolutional attention to improve summaries. However, RNNs face challenges like gradient issues with long sequences, addressed by LSTMs and GRUs, which introduce gating mechanisms for better information retention and computational efficiency. For example, Entity Summarization with Complementarity and Salience (ESCS) \cite{9718581} uses bi-directional LSTMs to enhance summary salience and complementarity, while GRU-based models like \cite{Zhang2018ExtractiveDS} offer faster training and efficient sentence extraction.

Transformers, with their attention mechanisms, have advanced ATS further by handling long sequences and capturing hierarchical dependencies. Models like \cite{grail2021globalizing} use hierarchical propagation layers to combine local and global attention, while \cite{pang2022long} leverage hierarchical latent structures to capture long-range dependencies and details. Additionally, topic-integrated Transformers \cite{wang2020friendly} enhance semantic clarity without altering the Transformer architecture. While these methods produce coherent and high-quality summaries, they still face challenges with computational efficiency and consistent controllability of output quality.

\textbf{Pre-trained Models}:
Pre-trained models, such as BART and Pegasus, are designed for rapid deployment in tasks like text summarization by leveraging extensive knowledge from large datasets without requiring training from scratch \cite{khandelwal2019sample}. BART \cite{lewis2019bart} is optimized for sequence-to-sequence tasks using a two-step approach: corrupting sentences with a noising function and training the model to reconstruct the original text, combining auto-encoder and auto-regressive pre-training objectives. Pegasus \cite{10.5555/3524938.3525989}, tailored specifically for summarization, masks or removes sentences from the input and generates them in the output, enabling highly-abstractive multi-document summarization \cite{Goodwin2020FlightOT}. Similarly, Socratic pre-training \cite{pagnoni2022socratic} enhances Transformer models by training them to generate and answer context-relevant questions, improving adherence to user queries and relevance in summarization tasks. These models effectively capture complex linguistic structures and contextual information, making them adaptable to diverse tasks and lowering the barrier for applying AI. Their success has paved the way for large language models.

\textbf{Pros and Cons:} Abstractive summarization models generate summaries resembling human-written text, offering greater flexibility and compression compared to extractive methods. However, they are more complex to develop, requiring high-quality datasets, significant computational resources, and extended training time, demanding a balance between cost and efficiency.

\subsection{Hybrid Summarization}
Hybrid summarization combines extractive and abstractive methods, typically by integrating an extractive model with either a shallow abstractive model (Extractive to Shallow Abstractive) or a fully designed abstractive model (Extractive to Abstractive).

\textbf{Extractive to Shallow Abstractive}:
Extractive to Shallow Abstractive (E2SA) methods first extract key sentences using ATS techniques and then apply light abstractive strategies such as information compression, fusion \cite{LLORET2013164}, and synonym replacement \cite{6968629}. \cite{10.1007/978-981-10-5687-1_56} proposed a hybrid approach, generating extractive summaries based on semantic and statistical features, including emotional affinity, and transforming them into abstractive summaries using WordNet, the Lesk algorithm, and a part-of-speech tagger.

\textbf{Extractive to Abstractive}:
Extractive to Abstractive (E2A) methods use extractive summarization to identify key sentences, which are then processed by an abstractive summarization model. For example, EA-LTS \cite{8029339} employs a hybrid sentence similarity measure for extraction and an RNN-based encoder-decoder with pointer and attention mechanisms for abstraction, improving efficiency and accuracy in long-text summarization.

\textbf{Pros and Cons:}
The hybrid ATS model aims to combine the strengths of extractive and abstractive methods to outperform both. However, in practice, the sparsity of extracted summaries often leads to lower-quality outputs compared to purely abstractive approaches.

\section{Large Langue Models (LLMs) based Summarization Methods}\label{model:LLM}
\label{sec:llm_ats}


Large Language Models (LLMs), with billions of parameters, are central to natural language processing due to their extensive training on diverse tasks and large text corpora \cite{Zhao2023ASO}. Typically structured as auto-regressive models like GPT, LLMs excel in applications such as summarization, question answering, and logical reasoning \cite{houlsby2019parameter}.

In ATS, LLMs have achieved results comparable to or exceeding human performance. Studies indicate that their summaries match human-crafted ones in quality and paraphrasing diversity \cite{tang2023a,laskar2023,ravaut2023,basyal2023}. Moreover, when prompted with simple task descriptions, LLMs generate preferred and highly factual summaries, avoiding dataset-specific pitfalls \cite{goyal2023}. This review bridges gaps in the literature by providing a comprehensive overview of LLMs in ATS.

LLM-based methods can be generally categorized as: 1) prompt engineering, 2)retrieval Augmented Generation, 3) fine-tuning, and 4) knowledge distillation. The advantages and limitations of these methods are summarized in Table \ref{tab:llm_methods}.

\begin{figure*}
    \centering
    \includegraphics[width=0.8\linewidth]{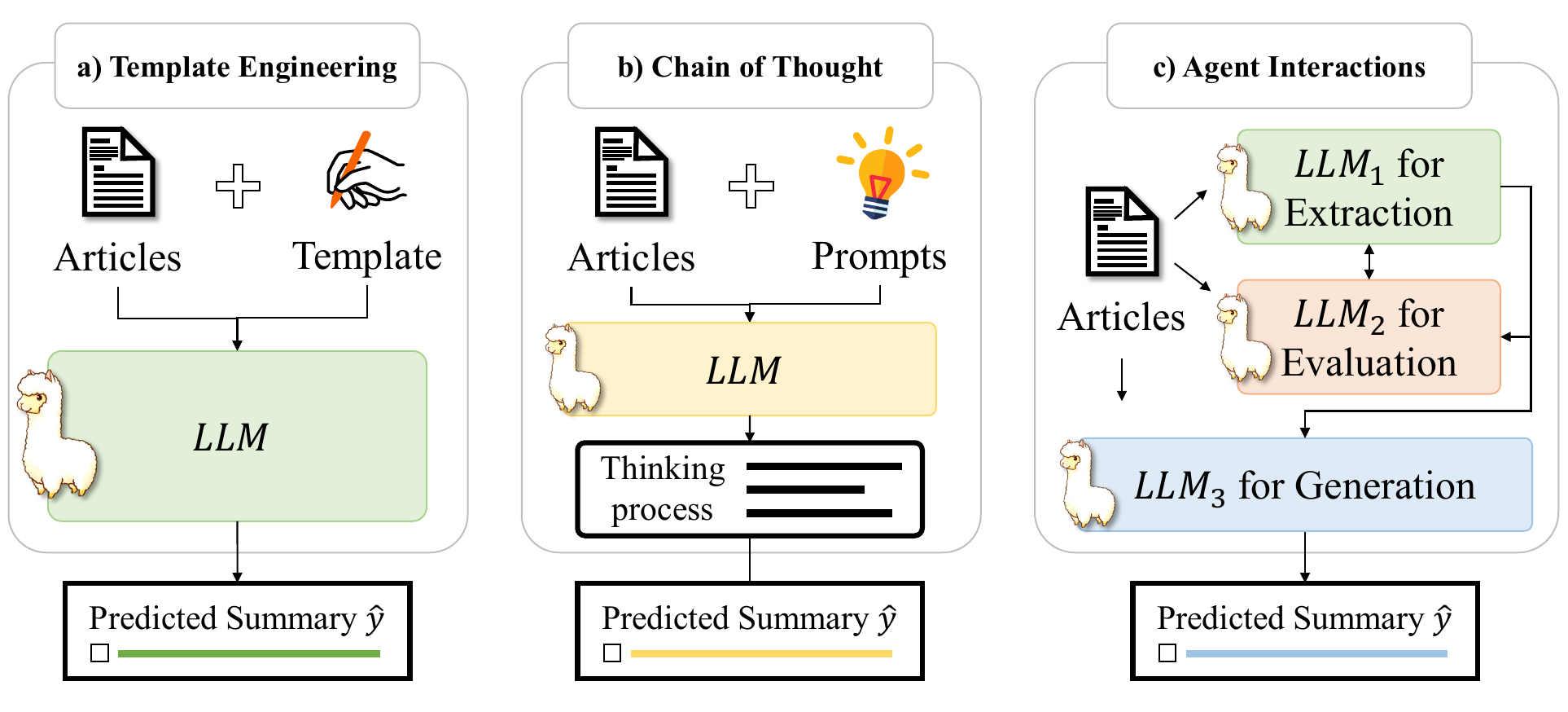}
    \caption{Methodology Overview of Prompt Engineering for LLM-based ATS methods, including: (a) Template-Based Prompting aims to design structured prompt words to guide LLMs in generating effective summaries; (b) Chain-of-Thought (CoT) Prompting incorporates step-by-step reasoning to help generate summaries; (c) Multi-Agent Collaboration involves distinct LLMs for extraction, evaluation, and generation, enabling modular summarization workflows}
    \label{fig:prompt}
\end{figure*}

\subsection{Prompt Engineering for LLM-based ATS}
\label{sec:prompt}
Prompt engineering involves strategically designing prompts to maximize the capabilities of Large Language Models (LLMs) to implement specific tasks. The process can be formalized as:  

\begin{equation}
S = LLM_\theta\left(G(p) + x\right).
\end{equation}  

\noindent In the equation \( S = LLM_\theta(G(p) + x) \), the function \( G \) represents a refinement process that transforms an initial prompt \( p \) into a more effective version using methods such as manual prompt engineering, retrieval of supplementary information, or automatic generation by another LLM. This refined prompt is then combined with the raw input text \( x \) to guide the language model \( LLM_\theta \), where \( \theta \) denotes the model's learned parameters. By enriching the input with a well-designed prompt, this approach helps the model better interpret the task and the context of the input, leading to improved quality in the generated summary \( S \).

The key advantage of prompt engineering is its efficiency, as it minimizes the need for extensive training and can operate effectively with a small set of examples \cite{narayan2021}. We review three specific categories of approaches under the prompt engineering technique in ATS, including Template Engineering, Chain of Thought and Agent Interactions, as illustrated in Figure \ref{fig:prompt}.

\subsubsection{Template Engineering for ATS}
Template engineering in text summarization employs structured linguistic scaffolds to guide LLMs in systematically extracting, condensing, and regenerating textual content while ensuring coherence, contextual relevance, and alignment with user-defined objectives. By integrating explicit instructions, contextual anchors, and output specifications, this approach leverages the semantic comprehension, discourse coherence, and domain adaptability of LLMs to produce summaries that preserve factual accuracy, emphasize key insights, and adhere to application-specific constraints, thereby enabling scalable and intent-aware summarization across diverse domains.
The representative templates are as follows:

\begin{itemize}
    \item $[Input]$ TL;DR: \cite{liu2023}
    \item Article: $[Input]$. Summarize the article in three sentences. Summary: \cite{10.1162/tacl_a_00632}
    \item Please generate a one-sentence summary for the given document.  
    $[Input]$ Try your best to summarize the main content of the given document and generate a short summary in one sentence. Summary: \cite{Zhao2023ASO}
\end{itemize}

In practice, \cite{8455241} proposed a dual-template framework for news summarization by leveraging SPARQL queries on knowledge graphs, distinguishing between a simple template that concatenated subject-predicate-object triplets from document subgraphs and a complex template that incorporated subgraph-level knowledge inference and aggregation to mimic human reasoning. The design emphasized syntactic traversal of directional triplets for basic summarization and hierarchical task decomposition for advanced reasoning. 
To enhance the alignment of LLMs with the requirements of dialogue summarization tasks, \cite{10650513} proposed Baichuan2-Sum, a role-oriented dialogue summarization model based on Baichuan2, which used a role-specific prompt template to address the limitations of small-scale models in capturing multi-role interactions. The design centered on assigning distinct instructions for different dialogue roles, enabling the model to learn role-dependent contextual patterns and generate targeted summaries.

\begin{table}[t]
\caption{Prompt templates for mental health risks summarization extracted from \cite{gyanendro-singh-etal-2024-extracting}.}
\label{t:template_2}
\begin{center}
    \setlength{\tabcolsep}{1mm}{
    \resizebox{\columnwidth}{!}{
\begin{tabularx}{\columnwidth}{X}
\hline\hline
As a mental health assistant, your task is to $[TASK]$ directly from the provided input text to highlight the mental health issues. For the $[TASK]$ task, consider the following aspects:
\begin{itemize}
    \item \textbf{Emotions}: Evaluate expressed emotions, from sadness to intense psychological pain, as they may influence the assigned risk level.
    \item \textbf{Cognitions}: Explore the individual's thoughts and perceptions about suicide, including the level and frequency of suicidal thoughts, intentions of suicide, and any existing plans.
    \item \textbf{Behavior and Motivation}: Evaluate the user's actions related to suicide, such as access to means and concrete plans. Consider their ability to handle difficult/stressful situations and the motivations behind their desire to die.
    \item \textbf{Interpersonal and Social Support}: Investigate the individual's social support or stable relationships, and understand their feelings toward significant others.
    \item \textbf{Mental Health-Related Issues}: Consider psychiatric diagnoses associated with suicide such as schizophrenia, bipolar, anxiety, eating disorder, previous suicidal attempts, and others.
    \item \textbf{Additional Risk Factors}: Consider other factors like socioeconomic and demographic factors, exposure to suicide behavior by others, chronic medical conditions, etc.
\end{itemize}
\\ \hline\hline
\end{tabularx}
}
    }
\end{center}
\vspace{-2em}
\end{table}

For healthcare community question answering, \cite{naik-etal-2024-perspective} introduced a novel perspective-specific answer summarization task, addressing the challenge of synthesizing diverse, multi-perspective responses into coherent summaries. To operationalize this, the authors designed a ``structured prompt template'' that explicitly defined perspective categories (e.g., suggestion), specified summary tone (e.g., advisory), and guided the model to generate summaries starting with predefined phrases (e.g., ``It is suggested''), ensuring alignment with the target perspective.
Furthermore, \cite{gyanendro-singh-etal-2024-extracting} investigated template prompting leveraging to detect and summarize evidence of suicidal ideation in social media, aiming to extract critical mental health-related information. The framework employed structured instruction templates that guided LLMs to focus on six dimensions shown in Table \ref{t:template_2}, including emotions, cognitions, behavior and motivation, interpersonal/social support, mental health issues, and additional risk factors. 
By providing explicit task-oriented instructions with contextualized examples, the study evaluated LLMs' inherent capability to identify key text spans indicating mental health risks.

\begin{table}[t]
\caption{Different prompt templates for ATS extracted from \cite{chrysostomou-etal-2024-investigating}.}
\label{t:template}
\begin{center}
    \setlength{\tabcolsep}{1mm}{
    \resizebox{\columnwidth}{!}{
\begin{tabularx}{\columnwidth}{lX}
\hline\hline
\# & Prompt Template \\
\hline
A & Summarize in a single short paragraph the context below: $[document]$ The summary is: $[summary]$ \\
\hline
B & Summarize in a couple of sentences the document below: $[document]$ The summary is: $[summary]$ \\
\hline
C & Give me a short summary of the below: $[document]$ The summary is: $[summary]$ \\ \hline\hline
\end{tabularx}
}
    }
\end{center}
\vspace{-2em}
\end{table}

To bridge the gap between general-purpose LLMs and domain-specific hardware description languages tasks, \cite{10.1145/3670474.3685966} designed a template-based prompting strategy that generates intermediate descriptive steps: for code generation, problem statements guided stepwise reasoning;  for summarization, the model was prompted to explain each line of code in context using a structured template (e.g., ``\texttt{<code\_context>} \texttt{\textbackslash n} Explain the following line of code given the context above: \texttt{\textbackslash n} \texttt{<code\_line>}''), ensuring alignment with syntactic and functional dependencies. 
To address the sensitivity exhibited by LLMs in prompt design, \cite{chrysostomou-etal-2024-investigating} implemented three distinct prompt templates in Table \ref{t:template} for document summarization, each designed to guide the model in generating summaries through subtly different instructional frameworks. The resulting triplets of summaries per document were aggregated via score averaging to mitigate variability and enhance evaluation reliability.

While manually crafting templates is intuitive and effective for generating reasonable summaries, this approach has its limitations. First, designing effective prompts can be time-consuming and heavily reliant on expertise \cite{shin2021constrained}. Second, even experienced prompt designers may struggle to identify optimal prompts for specific tasks \cite{jiang2020can}. To address these challenges, \cite{narayan2021} proposed leveraging entity chains mentioned in the target summary, as a basis for automating template design.

The effectiveness of these prompts is further amplified by the rapid advancement of LLMs, which possess enhanced capabilities in understanding nuanced semantic relationships, maintaining discourse coherence, and adapting to domain-specific terminology. As models grow in parameter count and training data diversity, the synergy between well-crafted prompts and model architecture ensures that summaries become increasingly precise, context-aware, and aligned with user intent, ultimately transforming text summarization into a more intelligent and scalable solution for both scholarly and industrial applications.

\subsubsection{Chain of Thought for ATS}\label{sec:cot}
Chain of Thought (CoT) is a prompting strategy that structures the model's reasoning process by decomposing complex summarization tasks into explicit, sequential inference steps. This approach guides LLMs to systematically analyze input text through intermediate reasoning, such as identifying key entities, discerning hierarchical relationships, and synthesizing core arguments, before generating a coherent summary. By embedding step-by-step logical derivations, CoT aligns the generative process of the model with human-like cognitive patterns, reducing hallucinations and enhancing factual consistency. This method is particularly effective in multi-step summarization scenarios, such as multi-document synthesis or domain-specific summarization tasks, where structured reasoning ensures outputs remain contextually grounded and aligned with user-defined objectives \cite{wei2022chain}.

A typical paradigm of CoT involves prompting LLMs to first systematically extract key elements from the input text, followed by generating a summary based on these structured intermediate representations. Illustrative Cases from \cite{wang2023a}:

\begin{enumerate}
    \item Article: [\textit{Input}]
    \item What are the important entities in this document? What are the important dates in this document? What events are happening in this document? What is the result of these events? Please answer the above questions:
    \item Let's integrate the above information and summarize the article:
\end{enumerate}

Following this work, \cite{sun2024sourcecodesummarizationera} adapted this paradigm to source code summarization by designing a code-specific CoT framework, the designed CoT included prompts:
\begin{enumerate}
    \item Instruction 1: Input the code snippet and five questions about the code in the format ``Code: \texttt{\textbackslash n} $<code>$ Question: \texttt{\textbackslash n} $<Q1>$ \texttt{\textbackslash n} $<Q2>$ \texttt{\textbackslash n} $<Q3>$ \texttt{\textbackslash n} $<Q4>$ \texttt{\textbackslash n} $<Q5>$ \texttt{\textbackslash n}''
    \item Get LLMs' response to Instruction 1, i.e., Response 1.
    \item Instruction 2: ``Let's integrate the above information and generate a short comment in one sentence for the function.''
    \item Get LLMs' response to Instruction 2, i.e., Response 2. Response 2 contains the comment generated by LLMs for the code snippet.
\end{enumerate}
when asking Instruction 2, Instruction 1 and Response 1 are
paired as history prompts and answers input into the LLM. By systematically addressing these questions through intermediate reasoning steps, the model generates summaries that preserve both syntactic accuracy and semantic coherence in source code contexts.

To balance conciseness and informativeness in summarization, \cite{adams2023} proposed a Chain of Density (CoD) prompting strategy to iteratively refine GPT-4 summaries. The CoT-inspired approach was motivated by the need to systematically address the tradeoff between entity coverage and summary readability, as overly sparse or dense outputs often fail to meet user expectations. The method involved first generating an entity-sparse summary via a standard prompt, then iteratively appending missing salient entities through a structured prompt sequence that maintained fixed output length while enhancing abstraction and fusion. The specific prompts are shown in the following case: 

\begin{tcolorbox}[colback=gray!10,
			colframe=black,
			width=9cm,
			arc=2mm, auto outer arc,
			title={Example of Chain-of-Thought Prompting via ``China of Density Prompting'' extracted from \cite{adams2023}},breakable,]
\label{fig:cod}
Article: ${{ARTICLE}}$

You will generate increasingly concise, entity-dense summaries of the above Article.

Repeat the following 2 steps 5 times.

\textbf{Step 1.} Identify 1-3 informative Entities (``,'' delimited) from the Article which are missing from the previously generated summary. 

\textbf{step 2.} Write a new, denser summary of identical length which covers every entity and detail from the previous summary plus the Missing Entities.

\textbf{Missing Entity is:}
\textbf{- Relevant:} to the main story.
\textbf{- Specific:} descriptive yet concise(5 words or fewer).
\textbf{- Novel:} not in the previous summary.
\textbf{- Faithful:} present in the Article.
\textbf{- Anywhere:} located anywhere in the Article.

\textbf{Guidelines:}
- The first summary should be long(4-5 sentences, ~80 words) yet highly non-specific, containing little information beyond the entities marked as missing,Use overly verbose language and fillers(e.g.,``this article discusses'') to reach ~80 words.

- Make every word count: re-write the previous summary to improve flow and make space for additional entities.

- Make space with fusion, compression, and removal of uninformative phrases like ``the article discusses''.

- The summaries should become highly dense and concise yet self-contained, e.g., easily understood without the Article.

- Missing entities can appear anywhere in the new summary.

- Never drop entities from the previous summary. If space cannot be made, add fewer new entities.

\textbf{Remember, use the exact same number of words for each summary.}

Answer in JSON, The JSON should be a list (length 5) of dictionaries whose keys are ``Missing Entities'' and ``Denser Summary''
\end{tcolorbox}

In addition, 3A-COT framework\cite{zhang20243a} was motivated by human-like reasoning patterns in MDS, leveraging CoT to decompose the task into three structured stages: Attend, Arrange, and Abstract.  First, an Attend-prompt extracted key elements from each document to mitigate redundancy;  second, an Arrange-prompt resolved contradictions and established logical connections between these elements;  finally, an Abstract-prompt synthesized a coherent summary based on the arranged information.  \cite{zhang-etal-2024-comprehensive} addressed the challenges of clustering and abstractive summarization of online news comments by integrating a CoT strategy to enhance LLMs in generating coherent summaries for noisy, ambiguous user-generated content. The CoT approach was motivated by the need to systematize the model's reasoning process when synthesizing dense, semantically diverse comment clusters, ensuring alignment with the structural and contextual nuances of real-world discussions.   

The limitations of CoT in text summarization primarily stem from its computational demands and task-specific constraints: the generation of explicit reasoning steps significantly increases inference costs and delays response times compared to direct output methods. Additionally, CoT raises concerns about loyalty and transparency, as advanced models frequently obscure their reliance on external cues or shortcuts in reasoning explanations, creating a disconnect between internal computations and external justifications that undermines trust in model honesty\cite{chen2025reasoning}.

\subsubsection{Agent Interactions for ATS}
Agent interactions frameworks first deconstruct the summarization task into sub-objectives, then iteratively refine outputs by leveraging internal memory and finally adjust strategies based on feedback from dynamic environments\cite{xi2023}, enhance the autonomy and adaptability of LLMs by embedding task-specific reasoning, memory, and environmental interaction mechanisms.

Typically, \cite{fang2025multillmtextsummarization} proposed a Multi-LLM summarization framework with two distinct strategies: centralized and decentralized. Both strategies involved k LLM agents generating diverse summaries, but differed in evaluation: centralized used a single central agent to select the best summary, while decentralized employed all k agents for evaluation. The implemented algorithm is shown in Algorithm \ref{alg:centralized_summary}.
\cite{mishra2024synfaceditsyntheticimitationedit} introduced an innovative pipeline leveraging LLMs with over 100B parameters, such as GPT-3.5 and GPT-4, to function as synthetic feedback experts capable of generating high-quality edit feedback for enhancing factual consistency in clinical note summarization. The researchers implemented two distinct alignment algorithms to utilize this synthetic feedback, aiming to reduce hallucinations and align weaker, smaller-parameter LLMs (e.g., GPT-2 and Llama 2) with medical facts without requiring additional human annotations.

\begin{algorithm}[t]
    \caption{A case related to Agents Interactions extracted from \cite{fang2025multillmtextsummarization}.}
    \label{alg:centralized_summary}
    \textbf{Require:} ordered set $S = \{S_1, \ldots, S_m\}$ of summaries, set $\mathcal{M} = \{M_1, \ldots, M_k\}$ of $k$ LLMs, a central agent $C \in \mathcal{M}$, max number of conversational rounds $t_{\max}$, initial summarization prompt $P$, evaluation prompt $P_{ec}$for centralized version\\
    \textbf{Ensure:} summary $S^*$ of the text
    \begin{algorithmic}[1]
        \State $S = \textsc{CreateSummary}(S)$
        \For{$i = 1$ to $t_{\max}$} \Comment{conversation rounds}
            \For{model $M_j \in \mathcal{M}$}
                \State $S^{(i)}_j = M_j(P, S)$
            \EndFor
            \State Let $\mathcal{S}_i = \{S^{(i)}_1, S^{(i)}_2, \ldots, S^{(i)}_k\}$
            \State $E^{(i)} = C(P_{ec}, \mathcal{S}_i)$
            \State $\mathbf{r} = \textsc{AggrResults}(E^{(i)})$
            \State $j \leftarrow argmax_{M_j \in \mathcal{M}} r_j$
            \State Set $S^* \leftarrow S^{(i)}_j$
            \If{\textsc{Converged}($\mathbf{r}$)}
                \State \Return $S^*$
            \EndIf
            \State Set $P$ to prompt.
        \EndFor
    \end{algorithmic}
\end{algorithm}

In addition, \cite{xiao2023} proposed a tri-agent pipeline including generator, instructor, and editor, to customize LLM-generated summaries to better meet user expectations. \cite{asi2022end} used agents to address dialogue summarization challenges like lengthy inputs, content verification, limited annotated data, and evaluation, with GPT-3 serving as an offline data annotator to handle dataset scarcity and privacy constraints. \cite{de-raedt-etal-2023-idas} employed clustering-based agents to improve intent detection by abstracting key elements while removing irrelevant information. Similarly, \cite{Li2024ChatCiteLA} introduced ChatCite, an LLM agent that extracts key elements from literature and generates summaries using a Reflective Incremental Mechanism for detailed summarization.

Besides, summarization-oriented LLM can be integrated as an autonomous agent within a multi-agent system framework. For example, \cite{gan2024applicationllmagentsrecruitment} proposed a novel agent framework for automating resume screening, motivated by the need to enhance efficiency and accuracy in recruitment processes through structured decision-making and scalable summarization of candidate profiles. The framework integrated automatic summarization as a core component, leveraging LLM agents to generate concise, graded summaries of resumes while aligning with domain-specific criteria such as skill matching and experience evaluation. 
\cite{NEURIPS2024_f7574461} investigated the integration of summarization agents into interactive systems, motivated by the need to align generated outputs with user preferences through iterative refinement.  The agent dynamically adjusted its output by retrieving and aggregating learned preferences from similar historical contexts, enabling personalized abstractions while preserving factual coherence in response to user-driven refinements.
\cite{NEURIPS2024_d0322637} proposed a planning framework for intelligent agents that utilized an automated summarization agent to extract and structure task-relevant knowledge from expert trajectories, enabling the synthesis of hierarchical state knowledge through comparative analysis and sequential abstraction to train a World Knowledge Model aligned with domain-specific task requirements.

In the context of summarization tasks, agents emerge as autonomous systems capable of integrating reasoning, planning, and tool utilization, distinguishing themselves from direct prompting and chain-of-thought (CoT) approaches. While direct prompting relies on static instructions to elicit model outputs and CoT enhances reasoning by structuring intermediate steps through human-designed prompts, agents operate dynamically, leveraging internal memory, environmental interaction, and adaptive decision-making to iteratively refine summaries. Their strengths lie in handling complex, multi-step workflows, such as cross-referencing external databases or synthesizing heterogeneous data sources, while maintaining contextual coherence and task-specific objectives. However, agents face challenges in interpretability, as their autonomous actions may obscure the rationale behind decisions. Additionally, compared to prompt-based methods, agents often demand higher computational resources and meticulous engineering to balance autonomy with precision, particularly in scenarios requiring strict adherence to factual accuracy or nuanced linguistic patterns. Thus, while agents excel in tasks demanding sustained reasoning and environmental engagement, their broader adoption hinges on addressing scalability, transparency, and robustness in real-world applications.

\subsection{Retrieval Augmented Generation for ATS}

\begin{figure}[t]
    \centering
    \includegraphics[width=0.8\linewidth]{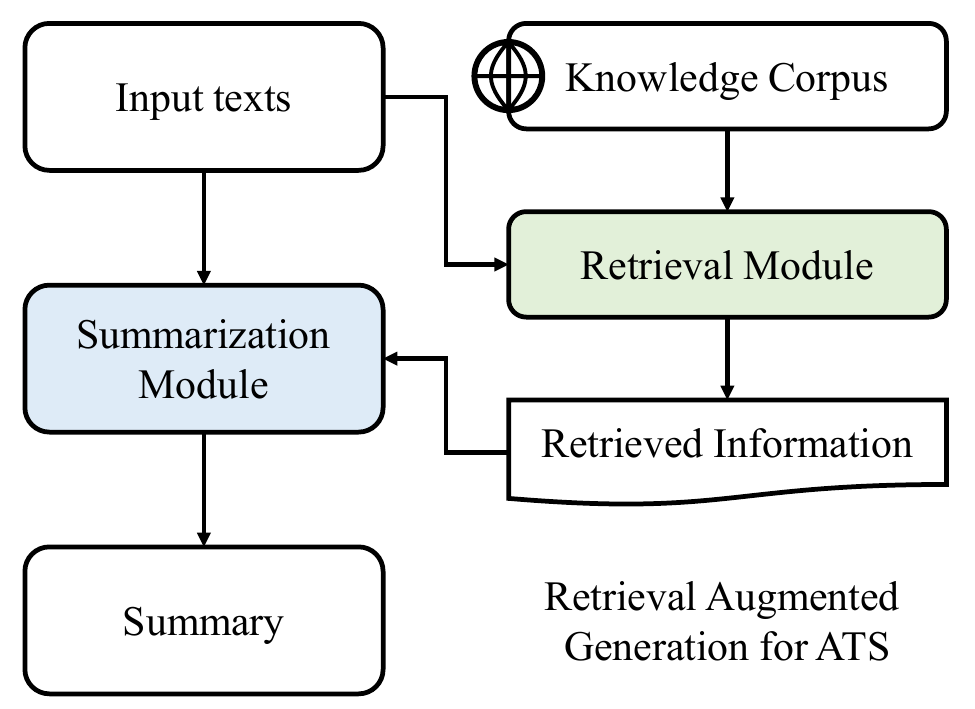}
    \caption{Example from \cite{zhang2024relevantdocumentsknowledgeintensiveapproach}. The RAG-based summarization methods first retrieve contextually relevant information from external knowledge bases using retrieval strategies, then generate abstractive summaries by integrating the retrieved content with the input document to ensure factual coherence and mitigate hallucinations.}
    \label{fig:rag}
    \vspace{-2mm}
\end{figure}

Retrieval Augmented Generation (RAG) enhances the summarization generation process by incorporating externally retrieved documents to supplement the model's internal knowledge \cite{li2022survey}, thereby improving the factual accuracy and coherence of summaries. Typically, as shown in Figure \ref{fig:rag}, this approach involves two stages: first, a retrieval mechanism such as BM25 or dense passage retrieval identifies semantically aligned segments from a pre-indexed knowledge base, and second, a sequence-to-sequence generator fuses these retrieved contexts with the input document to produce summaries that retain salient information while mitigating hallucinations. This dual-stage architecture ensures that the generated output is both linguistically fluent and grounded in verifiable external evidence.

Recent advancements in RAG have demonstrated diverse strategies for integrating external knowledge into generative models. \cite{edgeLocalGlobalGraph2024} introduced Graph RAG, which constructs a graph-based index by extracting structured entities and relationships from text, followed by community detection algorithms to partition the graph into semantically coherent clusters. This framework retrieves community-level summaries derived from knowledge graphs, enabling global context-aware reasoning for complex queries on large-scale datasets. Similarly, \cite{hong2025fgragenhancingqueryfocusedsummarization} proposed Context-Aware Fine-Grained Graph RAG, extended traditional graph-based retrieval by introducing context-aware entity expansion, which broadened the coverage of retrieved entities through dynamic contextual relationships, and query-level fine-grained summarization, which incorporated detailed, query-specific interactions during generation to align summaries with latent user intent. 

In highly specialized domains, where summarization tasks demand precise, domain-specific knowledge and up-to-date factual accuracy, Retrieval-Augmented Generation (RAG) plays a pivotal role by integrating external knowledge bases into the generation process. \cite{suresh2024ragbasedsummarizationagentelectronion} aimed to address the challenges of navigating complex, high-volume experimental data in the Electron Ion Collider domain by developing an RAG-based summarization framework (RAGS4EIC), motivated by the need to streamline information access for researchers while preserving contextual accuracy and traceability. The RAGS4EIC design implemented a two-stage process: first, querying a vector database of structured experimental data to retrieve relevant context, followed by generating concise, citation-enriched summaries using an LLM, with additional flexibility provided by prompt-template-based instruction-tuning to align outputs with user-specific query intent.
In code generation, \cite{parvez-etal-2021-retrieval-augmented} proposed REDCODER, employing dense retrieval to identify semantically similar code snippets or summaries from repositories, while \cite{zhou2023} developed a retrieve-then-summarize pipeline that used sliding window segmentation and template-based abstraction to condense retrieved code fragments into task-specific outputs. 

In the medical field, \cite{Manathunga2023RetrievalAG} designed a hybrid extractive-abstractive summarization approach, retrieving unstructured clinical texts (e.g., electronic health records or research papers) and integrating both direct evidence extraction and context-aware abstraction to generate domain-specific summaries. \cite{Liu2024TowardsAR} introduced LogicSumm and SummRAG, which leverage retrieval from structured medical knowledge bases (e.g., clinical guidelines or pharmacological databases) to enhance logical consistency and factual accuracy in LLM-based reasoning tasks.\cite{ji2024ragrlrclaysumbiolaysummintegratingretrievalaugmented} proposed the RAG-RLRC-LaySum framework to address the challenge of simplifying complex biomedical research for lay audiences, motivated by the need to bridge domain-specific jargon and enhance accessibility through structured knowledge integration. The design incorporated an RAG pipeline enhanced with a reranking mechanism, which aggregated diverse knowledge sources to ensure factual accuracy and contextual relevance in lay summaries, while a Reinforcement Learning for Readability Control component dynamically optimized linguistic simplicity and coherence during generation. 
\cite{zhu2024realmragdrivenenhancementmultimodal} addressed the limitations of existing Electronic Health Records (EHR) models in capturing nuanced medical context by proposing REALM, an RAG-driven framework designed to integrate external knowledge and bridge gaps in clinical predictive tasks. REALM's design combined LLM-based encoding of clinical notes with GRU-driven temporal modeling of EHR sequences, followed by a knowledge alignment module that extracted task-relevant entities from PrimeKG and fused them with multimodal data via an adaptive fusion network, ensuring semantic consistency and reducing hallucinations through contextual grounding in standardized medical knowledge.
\cite{saba2024questionansweringbasedsummarizationelectronic} designed a QA-based RAG framework to address the limitations of traditional neural models in EHR summarization, particularly the challenges of insufficient annotated training data and quadratic computational complexity due to long input lengths. The RAG design leveraged predefined questions identified by subject-matter experts to guide the extraction of targeted answers from EHRs, combining semantic search for relevant content retrieval with LLM-based generation to ensure factual consistency and reduce hallucinations, while avoiding the need for extensive training. 

Retrieval-augmented generation (RAG) for ATS integrates external knowledge retrieval with generative models to enhance factual accuracy and contextual relevance. By dynamically accessing external databases or domain-specific corpora, RAG mitigates the limitations of standalone language models, which may generate hallucinated or outdated information due to fixed training data. This approach ensures summaries are factually grounded, particularly in specialized domains such as legal or scientific texts, where access to authoritative references improves coherence and reliability. Furthermore, RAG introduces interpretability by anchoring summaries to retrievable sources, enabling traceable reasoning. However, its effectiveness is constrained by the quality and coverage of the retrieval system; incomplete or noisy knowledge bases can propagate errors into summaries. Additionally, the dual-stage pipeline—retrieval followed by generation—introduces computational overhead and latency, limiting real-time deployment in resource-constrained settings. The interplay between retrieved content and generative processes may also result in fragmented outputs if contextual integration is suboptimal, undermining fluency. While RAG addresses critical gaps in model knowledge and bias, its practical utility hinges on robust knowledge curation, efficient retrieval mechanisms, and seamless alignment of retrieved information with the summarization objective.

\subsection{Fine-tuning LLMs for ATS}
\label{sec:finetune}
\begin{figure}[t]
    \centering
    \includegraphics[width=0.9\linewidth]{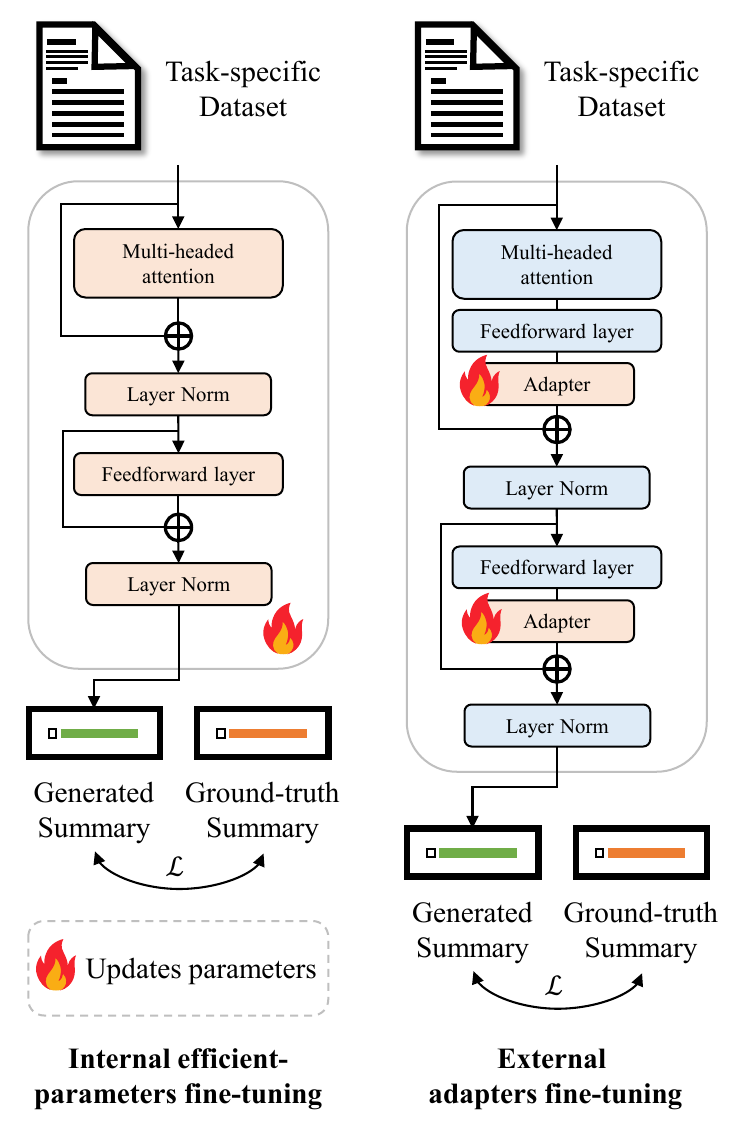}
    \caption{Example from \cite{houlsby2019parameter}. Fine-tuning LLMs for ATS are primarily categorized into internal fine-tuning approaches that selectively update a subset of pre-trained parameters to retain foundational linguistic knowledge and external adapters fine-tuning that employ compact neural modules inserted between transformer layers to encode domain-specific transformations.}
    \label{fig:finetune}
    \vspace{-2mm}
\end{figure}

Fine-tuning is pivotal for tailoring LLMs to specialized domains such as legal systems, medical diagnostics, and scientific research. This process significantly enhances their ability to grasp domain-specific knowledge and adapt to field-specific linguistic conventions. As shown in Figure \ref{fig:finetune}, the approaches typically fall into two categories: (1) Internal parameter‑efficient fine‑tuning, which strategically activates and updates only a subset of the LLM's internal parameters while preserving its pre-trained core capabilities; and (2) External adapters fine-tuning, where compact neural modules are inserted between transformer layers to learn domain-specific transformations\cite{houlsby2019parameter}. By focusing computational resources on these targeted adaptation mechanisms, fine-tuning enables LLMs to effectively capture ATS-relevant features such as semantic hierarchies, discourse structures, and task-specific terminology, while maintaining general language understanding capabilities.

\subsubsection{Internal efficient-parameters fine-tuning}

This approach involves freezing most of the LLM's parameters and updating only a few layers, such as the embedding or output layers. 
\cite{you-etal-2024-uiuc} addressed the challenge of summarizing lengthy biomedical literature for lay audiences by proposing an extract-then-summarize framework leveraging LLMs. The researchers implemented a fine-tuned GPT-3.5 model within this framework to enhance the system's ability to generate accessible summaries tailored to non-expert readers.
\cite{heZCodePretrainedLanguage} enhanced model efficiency by replacing the self-attention layers in the encoder with disentangled attention layers, representing each word with two vectors to encode its content and position. Similarly, \cite{Veen2023RadAdaptRR} explored efficient fine-tuning strategies for large language models in radiology report summarization, utilizing domain-specific pre-training while updating only 0.32\% of the model's parameters. 

Reinforcement Learning from Human Feedback (RLHF) \cite{bai2022traininghelpfulharmlessassistant} offers a powerful framework for aligning LLMs with human-centric evaluation criteria in the context of ATS. By constructing reward models through annotated datasets that encode preferences for conciseness, factual accuracy, and coherence, RLHF enables iterative policy optimization to refine summary generation. This approach addresses critical challenges in abstractive summarization, such as information compression while preserving key semantic elements, elimination of redundant expressions, and maintenance of factual consistency across multi-sentence outputs. 
\cite{huang2024nimplementationdetailsrlhf} systematically replicated and extended the RLHF-based scaling behaviors reported in OpenAI's TL;DR summarization work by constructing a novel RLHF pipeline from scratch, meticulously enumerating over 20 critical implementation details, and documenting key insights throughout the reproduction process.
\cite{nath2024leveragingdomainknowledgeefficient} proposed a novel methodology to reduce the dependency on extensive human preference annotations in RLHF by infusing domain-specific knowledge into the reward model, thereby enabling task-specific alignment with minimal annotated data while maintaining interpretability and adaptability to diverse human value preferences.

\subsubsection{External adapters fine-tuning}
Training external adapters allows models to learn the distribution of target data without modifying the original parameters of the LLM. \cite{bravzinskas2022efficient} proposed an efficient few-shot method using adapters, where the adapters were pre-trained on a large corpus and fine-tuned on a smaller human-annotated dataset. Similarly, \cite{xuSequenceLevelContrastive2022} introduced a contrastive learning approach for supervised ATS, aiming to maximize the similarity between a document, its gold-standard summary, and model-generated summaries. Additionally, \cite{guAssembleFoundationModels2022} combined encoder and decoder foundation models into a single model, AdaMo, and implemented adaptive knowledge transfer techniques such as continuous pretraining and intermediate fine-tuning, along with task-specific designs for sequence-to-sequence learning.

LoRA (Low-Rank Adaptation)\cite{hu2021loralowrankadaptationlarge} has been applied in ATS to efficiently fine-tune LLMs by introducing low-rank matrices that adapt pre-trained weights without modifying the original architecture, significantly reducing computational and memory costs. This method enables models to maintain high summarization performance while being optimized for specific tasks or domains, such as biomedical or technical text, with minimal additional parameters.
The DataHacks team developed a perspective-aware summarization approach\cite{nawander-nerella-2025-datahacks} for healthcare community question-answering forums by fine-tuning LLMs with various LoRA configurations to balance performance and computational efficiency. Their method involved identifying and classifying perspective-specific spans, followed by generating summaries based on distinct perspectives using adapted prompt strategies. \cite{mullick2024leveragingpowerllmsfinetuning} investigated the efficacy of fine-tuning open-source LLMs for aspect-based summarization by adapting pre-trained models such as Llama2, Mistral, Gemma, and Aya to domain-specific tasks. They employed supervised fine-tuning using prompt-completion pairs from the OASUM dataset, combined with parameter-efficient techniques like Quantized Low-Rank Adaptation (QLoRA) and PEFT to enhance training efficiency and model specialization for aspect-focused information extraction.

Fine-tuning LLMs for ATS offers significant advantages, including enhanced domain-specific adaptation to specialized contexts (e.g., biomedical or legal texts) by capturing nuanced linguistic patterns and task requirements. Techniques like LoRA enable parameter-efficient training, reducing computational costs while maintaining performance, and mitigating the need for extensive manual annotation through knowledge transfer from pre-training. However, this approach faces challenges such as overfitting to limited training data in low-resource domains, high computational demands for training and inference, and potential inheritance of biases or factual inaccuracies from training datasets. Domain mismatches between pre-training objectives and fine-tuning tasks may further degrade performance when source texts differ structurally or semantically from the target domain. Additionally, the ``black-box'' nature of LLMs complicates interpretability, posing barriers for high-stakes applications where transparency is critical. These trade-offs highlight the need for balanced strategies to optimize domain-specific performance while addressing computational, ethical, and interpretability constraints in ATS deployment.

\subsection{Knowledge Distillation from LLMs for ATS}
A knowledge distillation system consists of three key components: knowledge, a distillation algorithm, and a teacher-student architecture \cite{gou2021knowledge}. In this process, a smaller student model is supervised by a larger teacher model \cite{hinton2015distilling}. For ATS, this typically involves an offline distillation process where an LLM serves as a fixed teacher model, with its outputs providing additional supervision to train a smaller student model. This approach enables the student model to inherit the LLM's robust summarization capabilities, making it particularly useful in scenarios with limited computational resources or strict data privacy requirements. As shown in Table \ref{table:kd}, \cite{jiang-etal-2024-trisum} improved the quality of aspect-triple rationales and summaries through a dual scoring system, followed by a curriculum-learning strategy to train a smaller local model on ATS tasks. Similarly, \cite{10.1145/3675167} addressed low-resource cross-lingual summarization using mBART for incremental training and reinforcement learning to optimize discrete prompts. \cite{pham-etal-2023-select} proposed a three-step method for adapting smaller models to summarize forum discussions by sampling from a large corpus, retrieving annotated prompts, and filtering low-quality data.

\begin{table}[t]
\centering
\caption{Knowledge distillation ATS methods and their teacher/student models.}
\begin{center}
    \setlength{\tabcolsep}{2mm}{
\resizebox{\columnwidth}{!}{
\begin{tabular}{llllll}
\hline\hline
Ref.                                                 & Year & Domain   & T-model & S-model                   & Parameters \\ \hline
\cite{suDistilledGPTSource2024}                      & 2024 & Code     & GPT-3.5 & jam                       & 350M       \\
\cite{jiang-etal-2024-trisum}                        & 2024 & General  & GPT-3.5 & BART-Large                & 509M       \\
\cite{jung2024impossibledistillationlowqualitymodel} & 2023 & General  & GPT-3.5 & T5-large                  & 770M       \\
\cite{xu2023a}                                       & 2023 & General  & GPT-3.5 & ZCode++                   & 800M       \\
\cite{pham-etal-2023-select}                         & 2023 & General  & GPT-3.5 & Finetuned BART & 406.3M     \\
\cite{sclar2022}                                     & 2022 & General  & GPT-3   & GPT2-Large                & 774M       \\
\cite{asi2022end}                                    & 2022 & Dialogue & GPT-3   & BART-Large                & 509M+      \\  \hline\hline
\end{tabular}
}
\label{table:kd}
}
    \end{center}
\vspace{-2em}
\end{table}

Knowledge distillation from LLMs for ATS aims to transfer summarization capabilities from complex teacher models to smaller student models, enabling efficient deployment while retaining key strengths like factual consistency and coherence. This approach reduces computational costs and allows real-time applications on resource-limited devices, leveraging soft label distillation to mimic the teacher's nuanced outputs. However, distilled models inherit biases or errors from the teacher, struggle with multi-step reasoning tasks, and often underperform in  summarization compared to their larger counterparts. While cost-effective and scalable, their effectiveness depends on careful mitigation of inherited flaws and domain-specific fine-tuning, requiring a balance between efficiency and accuracy.

\textbf{Pros and Cons:} Large Language Models (LLMs) have transformed NLP tasks by simplifying training processes and enabling efficient summarization through prompt-based methods. However, they come with challenges. Their outputs can sometimes be inconsistent, leading to variability in performance. Additionally, even minor adjustments to prompt wording can greatly influence the quality of summaries, emphasizing the importance of refining prompting techniques. In domain-specific scenarios requiring extra training, the associated costs can become considerably higher than traditional approaches.

\begin{table*}[ht]
\centering
\caption{Description and pros/cons of overlap-based and similarity-based evaluation metircs.}
\label{tab:metric}
\setlength{\tabcolsep}{6pt}
\begin{center}
    \setlength{\tabcolsep}{2mm}{
    \resizebox{\textwidth}{!}{
\begin{tabularx}{\textwidth}{llXp{4cm}p{4cm}}
\hline\hline
Ref. & Name & Description & Pros & Cons \\
\midrule
\cite{lin2004rouge} & ROUGE & Measures n-gram overlap and longest common subsequences; prioritizes recall. 
& Simple, widely used. & Ignores fluency, may miss paraphrases. \\
\cite{papineni2001} & BLEU & Evaluates translation quality using precision and brevity penalty.
& Considers n-gram precision. & Penalizes brevity harshly, may miss semantic equivalence. \\
\cite{banerjee2005meteor} & METEOR & Uses WordNet to handle synonyms and paraphrases; distinguishes word types.
& Captures semantic equivalence. & Complex setup, potentially slower. \\
\cite{10.1121/1.2016299} & PPL & Estimates sentence probability normalized by length.
& Simple computation. & Doesn't consider context beyond word sequence. \\
\cite{zhang2020} & BERTScore & Computes token similarity; correlates with human judgments.
& High correlation with human judgment. & Requires pre-trained model. \\
\cite{NEURIPS2021_e4d2b6e6} & BARTScore & Uses pre-trained seq2seq models for evaluation.
& Comprehensive evaluation. & Requires large pre-trained models. \\
\cite{sellam-etal-2020-bleurt} & BLEURT & Models human judgment with strong performance on limited data.
& Strong performance on limited data. & Requires training on human ratings. \\
\cite{niu-etal-2021-unsupervised} & BERT-iBLEU & Evaluates paraphrasing with semantic closeness and IDF weighting.
& Encourages semantic closeness. & May reduce novel phrasing. \\
\cite{kryscinski-etal-2020-evaluating} & FactCC & Verifies factual consistency via span prediction.
& Verifies facts. & Supervision required for training. \\
\cite{laban-etal-2022-summac} & SummaC & Enhances inconsistency detection via sentence.
& Detects inconsistencies. & Complexity in setup. \\\hline\hline
\end{tabularx}
}
    }
    \end{center}
\vspace{-2em}
\end{table*}

\section{Evaluation Metrics}
\label{sec:evaluation}
Evaluating the quality of summaries is essential in summarization research \cite{ermakova2019}. Stable and consistent assessment methods are crucial for advancing the field, and this section introduces the evaluation metrics detailed in Table \ref{table:LLM}. Previous studies emphasize multi-dimensional evaluation, with \cite{gehrmann2018bottom} identifying relevance, factual consistency, conciseness, and semantic coherence as key metrics, where relevance assesses how well the summary captures the main ideas. Similarly, \cite{peyrard2019} proposed redundancy, relevance, and informativeness as core measures. Based on these insights, this paper categorizes evaluation methods into three groups: overlap-based, similarity-based, and LLM-based metrics, with non-LLM metrics summarized in Table \ref{tab:metric}.

\subsection{Metrics based on Term Overlap}
Overlap-based evaluation is one of the most widely used methods for assessing summarization quality. It measures the matching of words between candidate summaries (\(C\)) and reference summaries (\(S\)) and uses metrics such as precision (\(P\)), recall (\(R\)), and F-score (\(F\)) to quantify overlap \cite{gholamrezazadeh2009comprehensive}:
\begin{equation}
\begin{aligned}
    P =\frac{\left | S \cap C \right |}{C}, 
    R &=\frac{\left | S \cap C \right |}{S},
    F =\frac{\left (\beta + 1 \right ) \times P \times R}{\beta^{2} \times P + R} \\
    s.t. \ \beta^{2} &=\frac{1-\alpha}{\alpha}, \alpha \in \left [ 0,1 \right ] 
    \label{eq:overlap}
\end{aligned}
\end{equation}
However, individual word-overlap methods are limited because they do not account for the order or context of words in the candidate summaries. To address these shortcomings, several improvements to these metrics have been proposed.

\textbf{ROUGE} (Recall-Oriented Understudy for Gisting Evaluation) \cite{lin2004rouge} is a widely used metric for ATS evaluation, measuring overlap between candidate summaries (\(C\)) and reference summaries (\(S\)). 
ROUGE-N calculates n-gram overlap as:
\begin{equation}
\begin{aligned}
    ROUGE\text{-}N=\frac{\sum_{S}\sum_{gram_n \in S}Count_{match}(gram_n)}{\sum_{S}\sum_{gram_n \in S}Count(gram_n)} 
    \label{eq:rougen}
\end{aligned}
\end{equation}
where \(n\) is the n-gram length, typically \(n=1/2/3\), and \(gram_n\) is the number of matching n-grams in \(C\) and \(S\). ROUGE-L evaluates the longest common subsequence (LCS) between summaries, calculating precision, recall, and F-score. ROUGE-W adds positional weights to address LCS limitations, and ROUGE-S uses skip-bigram co-occurrences to allow gaps. While ROUGE is stable and reliable across sample sizes \cite{lin2004looking}, it primarily emphasizes recall and does not directly assess fluency or conciseness \cite{kohEmpiricalSurveyLong2023}.

\textbf{BLEU} (Bilingual Evaluation Understudy) \cite{papineni2001} evaluates summarization using precision, measuring the overlap between candidate (\(C\)) and reference (\(S\)) summaries. BLEU precision is calculated as:  
\begin{equation}
\begin{aligned}
    BLEU\text{-}P=\frac{\sum_{C}\sum_{gram_n \in C}Count_{clip}(gram_n)}{\sum_{C}\sum_{gram_n \in C}Count(gram_n)} 
    \label{eq:bleup}
\end{aligned}
\end{equation}  
Clipped counts cap candidate word counts at their maximum reference count. To penalize summaries shorter than the reference, BLEU introduces a brevity penalty (\(BP\)):  
\begin{equation}
    BP=
    \begin{cases}
        1 & \text{if } c > r \\
        e^{1-r/c} & \text{if } c \le r
    \end{cases}
    \label{eq:BP}
\end{equation}  
The BLEU score is then computed as:  
\begin{equation}
    BLEU=BP \cdot exp\left(\sum_{n=1}^{N}w_n \cdot logBLEU\text{-}P\right)
    \label{eq:BLEU}
\end{equation}  
where \(N\) is the maximum n-gram order, and \(w_n\) is typically set to \(1/N\). While BLEU effectively measures word overlap, it does not consider grammatical diversity or expressive variations.

\textbf{METEOR} (Metric for Evaluation of Translation with Explicit ORdering) \cite{banerjee2005meteor} was developed to overcome the rigidity of BLEU, which relies on exact n-gram matches. METEOR evaluates summaries by aligning words between a candidate and reference summary using flexible matching techniques, including synonyms, stemming, and paraphrases, facilitated by resources like WordNet. This allows METEOR to account for linguistic variations and semantic similarities that BLEU cannot capture. Additionally, METEOR differentiates between function words (e.g., ``the'', ``and'') and content words (e.g., ``run'', ``house''), giving more weight to content words to better reflect meaning. By incorporating these features, METEOR provides a more nuanced and flexible evaluation metric for summarization tasks.

\subsection{Similarity-based ATS Metrics}

Similarity-based metrics assess summary quality by measuring the semantic similarity between the candidate summaries and either the reference summaries or the source document. These metrics can be broadly grouped into two types: reference-based metrics, which compare the candidate summary against ground-truth references (e.g., BERTScore, BARTScore, BLEURT), and source-based metrics, which assess consistency or entailment between the candidate and the source document (e.g., FactCC, SummaC).

\textbf{BERTScore} \cite{zhang2020} evaluates text quality by computing a similarity score between each token in the candidate summary and each token in the reference summary, using contextual embeddings from pre-trained language models like BERT. These approaches capture semantic meaning and contextual relationships, offering a more flexible and effective evaluation compared to traditional n-gram-based metrics.

\textbf{BARTScore} \cite{NEURIPS2021_e4d2b6e6} evaluates text quality by computing the generation likelihood of a candidate summary conditioned on a reference summary or source document, using a pre-trained sequence-to-sequence language model like BART. By treating evaluation as a text generation task, BARTScore leverages the full generative capacity of large models to capture semantic adequacy and fluency. This approach enables more nuanced assessment of summary quality, especially in abstractive scenarios, and has been shown to correlate well with human judgments, outperforming traditional n-gram and embedding-based metrics.

\textbf{BLEURT} \cite{sellam-etal-2020-bleurt} builds on BERT embeddings by incorporating additional fine-tuning on human-annotated data and synthetic examples. This fine-tuning process allows BLEURT to explicitly model text quality judgments, aligning more effectively with task-specific requirements such as fluency, coherence, and relevance. Compared to BERTScore, BLEURT provides a more comprehensive evaluation by integrating various aspects of text quality.

\textbf{BERT-iBLEU} \cite{niu-etal-2021-unsupervised} evaluates paraphrasing by balancing semantic similarity and surface-form diversity. It promotes semantic closeness using BERTScore and discourages surface-level similarity with a penalty based on self-BLEU. The metric is defined as:

\begin{equation}
\begin{scriptsize}
\begin{aligned}
\text{BERT-}i\text{BLEU} &= \left(\frac{\beta * \text{BERT-}score^{-1}+1.0 *(1-self\text{-BLEU})^{-1}}{\beta+1.0}\right)^{-1} \\
self\text{-BLEU} &=\text{BLEU}(source, candidate)
\end{aligned}
\end{scriptsize}
\end{equation}

\noindent where \(\beta\) controls the trade-off between semantic similarity and surface-form dissimilarity, and self-BLEU measures similarity between the source and candidate texts. As BERT-iBLEU is reference-free, it can be used both as an evaluation metric for paraphrase quality and as a criterion for re-ranking candidates during task adaptation and self-supervision.

\textbf{FactCC} \cite{kryscinski-etal-2020-evaluating} takes a source document and a summary statement as input and outputs the fact consistency label for that summary statement. FactCC is a weakly-supervised model proposed to verify factual consistency in summaries by predicting consistency, extracting supporting spans, and identifying inconsistent spans, which outperforms strongly supervised models and aids human verification. By modeling different types of fact consistency errors, the method achieves improved performance consistency evaluation.

\textbf{SummaC} \cite{laban-etal-2022-summac} evaluates the factual consistency of summaries by leveraging Natural Language Inference (NLI) models. It compares summary sentences with segments of the source document to determine entailment, contradiction, or neutrality. Sentence-level NLI scores are aggregated to produce a document-level consistency score, as the overall score for the generated summaries. SummaC effectively bridges the gap between traditional overlap-based metrics and semantic understanding in summarization evaluation.

\subsection{LLM-based ATS Metrics}
Traditional evaluation metrics for assessing summary quality primarily focus on specific dimensions such as lexical overlap (e.g., ROUGE) or semantic similarity (e.g., BERTScore). However, these approaches exhibit notable limitations in capturing the multifaceted nature of summarization tasks. By contrast, large language models (LLMs) demonstrate a capacity for more nuanced and comprehensive analyses through prompt engineering that leverages their contextual understanding and reasoning capabilities. 
Inspired by \cite{li-etal-2024-leveraging-large}, existing LLM-based ATS metrics can be systematically categorized into four following groups:

\subsubsection{Score-based LLM Evaluation Metrics}
Score-based evaluation metric prompts LLMs to generate continuous quality scores for summaries, typically on a predefined scale. For example, LLMs might evaluate fluency, coherence, or relevance directly by assigning numerical ratings based on the input text and reference (if provided). 

\textbf{G-EVAL} \cite{liu-etal-2023-g} introduced a human-aligned reference-free framework that leverages GPT-4 with chain-of-thought reasoning and form-filling paradigms to assess generated summaries, addressing limitations of traditional reference-based metrics in capturing creativity and diversity. 

\textbf{ICE} \cite{jainMultiDimensionalEvaluationText2023} leverages few‑shot in‑context learning. By exposing GPT‑3 to four sets of manually scored examples, the model is prompted to output dimension‑specific scores (ranging from 0 to 1) for target summaries, enabling the assessment of coherence, relevance, fluency, and consistency without additional training.

\textbf{SumAutoEval} \cite{yuan2024evaluatesummarizationfinegranularityauto} decomposed summarization quality into four granular, entity-centric dimensions: Completeness, Correctness, Alignment, and Readability, by extracting fine-grained entities from both the ground-truth and candidate summaries and then prompting an LLM with a suite of consistency-checked prompts to verify each entity's coverage, factual accuracy, positional relevance, and linguistic fluency, finally aggregating the entity-level judgments into interpretable, human-aligned scalar scores.

\textbf{OP-I-PROMPT} \cite{siledar-etal-2024-one} distilled the evaluation of opinion summaries into a single, dimension-independent prompt that instructed an LLM to score each summary on seven discrete aspects: fluency, coherence, relevance, faithfulness, aspect coverage, sentiment consistency, and specificity.

\textbf{CREAM} \cite{gong2024creamcomparisonbasedreferencefreeeloranked} distills completeness, conciseness, and faithfulness into a single reference‑free, comparison‑based prompt for GPT‑4, instructing it to extract key facts from paired meeting summaries and score each summary on these three aspects without relying on gold references or source transcripts.

\textbf{Talec} \cite{zhang2024talecteachllmevaluate} were operationalized through a set of ten binary labels: five unacceptable labels (score = 0) and five acceptable labels (score = 1), that jointly determined a final score of 0, 1, or 2 (full).  These labels assessed whether a response failed to meet requirements, contained incorrect or unrelated content, refused to answer, exhibited untranslated text, confusing structure, external links, stiffness, repetition, subject imprecision, or incompleteness.

Score-based evaluation metrics leverage the reasoning capabilities of large language models (LLMs) to generate continuous quality scores for summaries, typically on a predefined scale. This approach enables direct assessment of dimensions such as fluency, coherence, and relevance by assigning numerical ratings based on input text and reference summaries (if available). The primary strength of this method lies in its quantitative precision, allowing for straightforward comparisons across models or outputs. However, the reliance on predefined criteria introduces subjectivity, as the interpretation of ``high fluency'' or ``strong relevance'' may vary across LLMs or evaluators. Additionally, this method often depends heavily on high-quality reference summaries, which are not always accessible or representative of diverse user preferences, potentially limiting its applicability in real-world scenarios.

\subsubsection{Probability-based LLM Evaluate Metrics}
Probability-based evaluation metrics compute the likelihood of generating a summary given a source text or reference. This method assumes that higher-generation probabilities correlate with better quality, reflecting semantic alignment and fluency. 

\textbf{GPTScore} \cite{fu-etal-2024-gptscore} prompted LLM evaluator to calculate the generation probability of generated text with definitions of evaluation tasks and aspects that leveraged the zero-shot reasoning capabilities of LLMs to assess generated summaries through natural language instructions.

\textbf{ProbDiff} \cite{xia-etal-2024-language} proposed a self‑evaluation metric that quantifies an LLM's competence by computing, for each query, the negative log‑probability difference between its initial response and a K‑step refined response. Under the empirically supported assumption that stronger models exhibit flatter probability distributions, a smaller discrepancy indicates higher confidence and better performance, enabling model evaluation without external annotations or additional training.

\textbf{BWRS} \cite{gao-etal-2024-bayesian} proposed a Bayesian calibration framework that adjusts win‑rate estimates from imperfect LLM evaluators by modeling both observed outcomes and evaluator accuracy as Beta‑Bernoulli processes. By drawing samples from the posterior distributions of evaluator true positive or negative rates and observed win counts, BWRS derives an unbiased estimate of the true model win rate. This calibration both corrects evaluator bias and quantifies uncertainty, enabling more reliable pairwise comparisons without extensive human labeling.

Probability-based evaluation metrics assess summary quality by computing the likelihood of generating a summary given a source text or reference, under the assumption that higher-generation probabilities correlate with better alignment and fluency. This method capitalizes on the pretraining of LLMs on vast corpora, enabling objective probabilistic assessments grounded in statistical patterns. A key advantage is its efficiency, as it requires minimal computational overhead compared to human annotation. However, this approach risks conflating fluency with factual accuracy, as a grammatically correct but semantically incorrect summary may receive a high probability score. Furthermore, the method may prioritize common patterns in training data over novel or user-specific summarization needs, introducing biases that compromise its reliability in diverse contexts.

\subsubsection{LLM-based Pairwise Comparison Metrics}
Pairwise comparison metrics compare two summaries to determine which one is superior in terms of quality, coherence, or relevance. This method is particularly useful for ranking models or identifying strengths/weaknesses in competing outputs. 

\textbf{BOOOOKSCORE} \cite{changBooookScoreSystematicExploration2023}  constructed a reference-free coherence metric by distilling human-annotated coherence-error taxonomies into a GPT-4 driven, few-shot prompt that classified each summary sentence against eight error types and reported the proportion of unconfused sentences, thus enabling scalable, LLM-agnostic evaluation of book-length summarization systems.

\textbf{Calibrated Confidence Score} \cite{virk2025calibrationlargelanguagemodels} computes the geometric mean of per-token generation probabilities and then applies Platt scaling using human‑similarity labels (derived from BERTScore) to produce a calibrated confidence score that aligns model confidence with human judgments.

\textbf{PRE} \cite{10.1145/3627673.3679677} introduced the Peer Review Evaluator, which quantifies alignment between automated and human judgments by computing agreement rate, Kendall's tau, and Spearman correlation across summarization and non-factoid question‑answering tasks.

\textbf{FineSurE} \cite{song-etal-2024-finesure} decomposed summarization evaluation into three fine-grained dimensions: faithfulness, completeness, and conciseness, by prompting a large language model (1) to classify each summary sentence into nine factuality-error types and (2) to align a set of human- or LLM-extracted keyfacts with the corresponding summary sentences, then aggregating sentence- and keyfact-level judgments into interpretable percentage scores that correlated strongly with human annotations.

\textbf{ACUEval} \cite{wan-etal-2024-acueval} decomposed each summary into atomic content units and validated them against the source document via an LLM-driven two-step pipeline, yielding interpretable, fine-grained faithfulness scores that improved balanced accuracy over prior metrics by 3\% and enabled actionable error correction.

Pairwise comparison metrics evaluate two summaries to determine relative superiority in terms of quality, coherence, or relevance, mimicking human judgments by ranking outputs based on their strengths. This method is particularly effective for model benchmarking and identifying strengths/weaknesses in competing outputs, as it aligns with how humans intuitively compare alternatives. Its flexibility allows application even without reference summaries, relying solely on the LLM's contextual reasoning. Nevertheless, this approach lacks an absolute benchmark, making it challenging to quantify overall performance improvements. Additionally, the absence of explicit criteria for comparison (e.g., prioritizing factual accuracy over conciseness) can lead to inconsistent evaluations, especially when summaries differ significantly in multiple dimensions.

\subsubsection{Hybrid Evaluation Metrics}
Hybrid evaluation metrics use multiple metrics or LLMs to assess summaries from different perspectives. This reduces bias and captures a wider range of quality dimensions. 

\textbf{Humanlike} \cite{gaoHumanlikeSummarizationEvaluation2023} examined four human-inspired evaluation protocols, including Likert-scale rating, pairwise comparison, Pyramid content-unit matching, and binary factuality verification, through which ChatGPT assessed summarization quality on five benchmark corpora. By computing Spearman correlations and accuracy against expert annotations, the authors demonstrated that ChatGPT's judgments, particularly when prompted with concise task descriptions, surpassed traditional surface-level metrics (ROUGE, BERTScore, BARTScore) and rivaled human inter-annotator agreement, yielding a cost-effective, reproducible alternative for summary evaluation.

\textbf{DRPE} \cite{wuLargeLanguageModels2023} leveraged a diverse set of LLM-based role-players, statically defined for objective dimensions such as coherence and grammar and dynamically generated for subjective dimensions such as interestingness,to conduct pairwise comparisons between machine-generated and reference summaries, aggregated the resulting votes via batch prompting, and was empirically shown to correlate more strongly with human judgments than conventional overlap- or embedding-based measures.

\textbf{FENICE} \cite{scirè2024fenicefactualityevaluationsummarization} introduced an interpretable and computationally efficient factuality metric for summarization evaluation that extracted atomic claims from summaries via an LLM and distilled them into a T5-base generator, aligned each claim with document spans at sentence, paragraph and document granularity through a DeBERTa-v3-large NLI model, refined the entailment scores with coreference resolution,  and achieved state-of-the-art balanced accuracy on the AGGREFACT benchmark while surpassing baselines on a newly annotated long-form summarization dataset.

\textbf{DEBATE} \cite{kim-etal-2024-debate} (Devil's Advocate-Based Assessment and Text Evaluation), a multi-agent LLM framework that instantiated a Commander–Scorer–Critic architecture in which a Critic was explicitly prompted to adopt the role of Devil's Advocate and iteratively challenged the Scorer's judgements. By enabling structured debates capped at a pre-defined number of rounds, DEBATE alleviated single-agent biases, yielded scores that exhibited significantly higher correlations with human ratings on the SummEval and Topical-Chat meta-evaluation benchmarks, and established a new state-of-the-art among reference-free, LLM-based NLG evaluators.

Hybrid evaluation metrics integrate multiple methods, such as score-based, probability-based, and pairwise comparisons, or combine outputs from different LLMs to assess summaries from diverse perspectives. This approach mitigates the limitations of individual metrics by aggregating their strengths, such as reducing subjectivity in scoring or capturing both absolute and relative quality dimensions. Its adaptability allows customization for specific tasks, such as emphasizing factual accuracy in news summarization or conciseness in technical documents. However, hybrid methods introduce complexity in design and interpretation, as aggregating diverse outputs (e.g., scores, probabilities, rankings) may obscure insights into specific quality aspects. Moreover, the resource intensity of integrating multiple models or metrics increases computational costs, necessitating careful calibration to balance comprehensiveness and efficiency.

\section{Applications based on ATS}
\label{sec:applications}
The ultimate objective of Automated Text Summarization (ATS) is to enhance the efficiency of information retrieval and analysis in real-world scenarios. ATS has applications in various domains, including the summarization of news articles, scientific papers, and other areas that involve substantial reading efforts.

\textbf{News Summarization}: represents the most extensively investigated domain within ATS, largely owing to the richness and maturity of available datasets, coupled with the significant demand for its application across various industries.
\cite{barzilay2005} introduced sentence fusion, a text-to-text generation technique that merges phrases containing similar information into a single sentence for summary generation. Due to the real-time nature of news, \cite{mckeown2002} presented Newsblaster, a system that is based a clustering methods for crawled news articles by topic, and thus produces summaries based on topic clusters. In addition, based on the chronological nature of the forward and backward relations of news articles, \cite{ghalandari2020} proposes a timeline-based summarization method, 

\textbf{Novel Summarization}: focuses on condensing novel articles, texts characterized by their extended length. \cite{ladhak2021} suggested calculating the alignment score using ROUGE and METEOR metrics to identify the most suitable sentences for crafting summaries. \cite{ladhak2021} developed two techniques for creating fictional character descriptions, deriving from the articles' attributes and centered on the dependency relationships among pivotal terms.

\textbf{Scientific Paper Summarization}: presents challenges including managing citation relationships, specialized structural parsing, and generating accurate proper names. CGSum\cite{an2021enhancing} is a model for summarization based on citation graphs, capable of integrating information from both the primary paper and its references using graph attention networks\cite{velivckovic2017graph}. SAPGraph \cite{qi2022sapgraph} offers an extractive summarization framework for scientific papers, utilizing a structure-aware heterogeneous graph. This framework represents the document as a graph with three types of nodes and edges, drawing on structural information from facets and knowledge to model the document effectively.

\textbf{Blog Summarization}: is crucial for accessing real-time information, with platforms like Twitter and Facebook hosting millions of highly pertinent and timely messages for users. \cite{shen2013participant} introduced a participant-based event summarization approach that expands the Twitter event stream through a mixture model, identifying significant sub-events tied to individual participants. \cite{inouye2011comparing} formulated a methodology to globally and locally model temporal context, proposing an innovative unsupervised summarization framework that incorporates social-temporal context for Twitter data. \cite{inouye2011comparing} examined various algorithms for the extractive summarization of microblog posts, presenting two algorithms that generate summaries by selecting a subset of posts from a specified set.

\textbf{Dialogue Summarization}, encompassing summarization of meetings, chats, and emails, is an increasingly pertinent task to deliver condensed information to users. \cite{kumarMeetingSummarizationSurvey2022} identified that the challenge in dialogue summarization lies in the heterogeneity arising from multiple participants' varied language styles and roles. \cite{mehdad2014} introduced a query-based dialogue summarization system that selects and extracts conversational discourse based on the overall content and specific phrase query information. \cite{ganesh2020} developed a Zero-shot approach to conversation summarization, employing discourse relations to structure the conversation and leveraging an existing document summarization model to craft the final summary. 

\textbf{Medical Summarization}: Summarization in the medical field holds substantial clinical significance, as it has the potential to expedite departmental workflows (e.g., in radiology), diminish redundant human labor, and enhance clinical communication \cite{kahn2009}. \cite{sotudeh2020} crafted a universal framework for evaluating the factual accuracy of summaries, utilizing an information extraction module to conduct automated fact-checking on the citations within generated summaries. \cite{adler2012} devised a novel implicit-based text exploration system, tailoring it to the healthcare sector. This system allows users to navigate the results of their queries more deeply by traversing from one proposition to another, guided by a network of implicit relations outlined in an implicit graph. \cite{zhang2020a} introduced a model founded on seq2seq that integrates essential clinical terms into the summarization process, demonstrating marked improvements in the MIMIC-CXR open clinical dataset.

\section{Future Directions in Large Language Model-Based Summarization}
\label{sec:future_direction}
Despite the remarkable progress of Large Language Models (LLMs) in Automatic Text Summarization (ATS), several limitations remain in current research. These include challenges in adapting general-purpose LLMs to domain-specific summarization tasks, handling long documents under the token length constraints of LLMs, and improving the interpretability of internal LLM mechanisms. In this section, we outline key future research directions in LLM-based ATS, grounded in the limitations identified throughout this survey.

\textbf{Mitigating Hallucination and Enhancing Trust in LLMs for ATS: }
One of the key challenges in applying Large Language Models (LLMs), not only in Automatic Text Summarization (ATS) but across a wide range of NLP tasks, is the issue of hallucination, where the generated content is factually incorrect or not grounded in the source text. In the context of ATS, this results in summaries that may include fabricated details, misrepresentations, or unsupported claims, thereby compromising the reliability and trustworthiness of the output\cite{maynez-etal-2020-faithfulness,pagnoni-etal-2021-understanding,cao-etal-2022-hallucinated}. This problem is particularly pronounced in abstractive summarization, where LLMs generate novel sentences without strict reliance on the input content.

To address hallucination in ATS, future research could potentially explore three key directions. First, hallucination detection and evaluation require significant advancement. Conventional metrics such as ROUGE and BLEU rely heavily on lexical overlap and are inadequate for identifying factual inconsistencies. While newer metrics like BERTScore provide semantic similarity measures, they still fall short in assessing factual alignment. There is a pressing need for factuality-aware evaluation metrics that can reliably quantify the trustworthiness and source alignment of generated summaries. Second, grounded and retrieval-augmented summarization offers a potential direction to mitigate hallucination by conditioning generation on sentence-level evidence or extracted facts from the input. Techniques such as Retrieval-Augmented Generation (RAG) or hybrid extractive-abstractive frameworks can serve to anchor LLM outputs in verifiable content. Third, emerging work on reinforcement learning for factual alignment presents a compelling direction, where trust-oriented reward signals, based on external fact-checkers or entailment models, can guide LLMs toward producing summaries that are not only fluent but also faithful to the source.

\textbf{Scalable Long-Document Summarization with LLMs: }
A critical limitation of current Large Language Model (LLM)-based Automatic Text Summarization (ATS) systems is their restricted ability to handle long-form documents due to input token constraints. Popular models such as GPT-3 and PaLM typically support context windows ranging from 2K to 32K tokens, which significantly limits their applicability to documents like legal rulings, scientific publications, or technical manuals that require reasoning across extended discourse structures \cite{Brown2020LanguageMA, 10.5555/3648699.3648939}. When input texts exceed these boundaries, truncation can lead to the loss of essential information, while naive chunking strategies often disrupt semantic coherence and degrade summary quality.

To address these challenges, one promising research direction is the development of long-context modeling strategies. These include memory-augmented transformers, recurrence-based architectures, and retrieval-augmented generation frameworks, which aim to handle longer sequences efficiently without linear increases in computational complexity \cite{ainslie-etal-2020-etc, guo-etal-2022-longt5}. In parallel, chunk-wise summarization frameworks, which segment documents into coherent units and incorporate cross-segment reasoning or hierarchical attention mechanisms, can help maintain contextual continuity and improve overall summary cohesion. Exploring these directions will be essential for advancing the scalability and robustness of LLM-based summarization systems in real-world, document-intensive scenarios.

\textbf{Improving Interpretability in LLM-Based ATS: }
Despite the fluency and human-likeness of summaries generated by LLM-based ATS systems, the underlying generation process often remains a black box, offering limited transparency into how specific content is derived from the source text. This lack of interpretability poses a critical limitation, particularly in applications where accuracy and accountability are essential. Users may hesitate to trust or adopt ATS systems unless they can clearly verify the provenance, contextual relevance, and justification of each piece of generated information.

To improve interpretability in LLM-based ATS, future research should explore two potential directions grounded in recent advancements in large language model techniques. First, prompt-based interpretability, including Chain-of-Thought (CoT) prompting and structured prompt engineering, offers a lightweight yet effective way to make the generation process more transparent. By encouraging models to generate intermediate reasoning steps or explicitly cite relevant portions of the input, these approaches help users understand the logical flow behind the summary content without requiring model retraining. Second, retrieval-augmented summarization (RAG) can improve interpretability by linking generated summaries to specific, verifiable source passages retrieved at inference time. These directions collectively move toward ATS systems that not only produce high-quality summaries but also provide clear, verifiable reasoning aligned with user expectations and real-world accountability standards.

\textbf{Domain-Specific Summarization with LLMs: }
While Large Language Models (LLMs) have demonstrated good capabilities in generic summarization tasks, their performance often degrades in domain-specific scenarios such as biomedical literature, legal documents, financial reports, or technical manuals. These domains present unique challenges, including specialized terminology, complex discourse structures, and high factual precision requirements, which general-purpose models trained on open-domain corpora are inappropriate to handle. As a result, LLM-generated summaries in these settings may lack accuracy, omit critical content, or fail to meet the precision and clarity expected by domain experts.

To address these challenges, future research could potentially explore three directions. First, domain-adaptive fine-tuning remains a widely used and effective approach, where LLMs are further trained on in-domain corpora to internalize specialized knowledge. Notable examples based on the pre-trained models include BioBERT for biomedical texts \cite{lee2020biobert}, Legal-BERT for legal documents \cite{chalkidis2020legal}, and SciBERT for scientific articles \cite{beltagy2019scibert}. To reduce the computational burden of full model retraining, parameter-efficient tuning methods such as LoRA \cite{hu2021loralowrankadaptationlarge} offer a promising alternative by enabling lightweight adaptation with minimal additional parameters. Second, prompt-based adaptation strategies, such as in-context learning or few-shot prompting using domain-specific exemplars, can guide LLMs toward stylistically appropriate and contextually relevant summaries without modifying model weights \cite{sanh2021multitask}. Third, integrating domain knowledge from structured resources (e.g., ontologies, taxonomies, or controlled vocabularies) through retrieval-augmented or knowledge-grounded generation frameworks \cite{liu2022kat} can further enhance the factuality and informativeness of the output. These directions collectively aim to build robust summarization systems that are better aligned with the demands of specialized domains and capable of producing trustworthy, high-utility summaries for professional use.

\section{Conclusion}

In this survey, we present a comprehensive review of Automatic Text Summarization (ATS) techniques, focusing on the evolution of both conventional methods and Large Language Model (LLM)-based approaches. This work makes the following key contributions: (1) providing an \textbf{Up-to-date Survey on ATS}, reviewing the latest developments in traditional, inflexible summarization methods (i.e., extractive and abstractive approaches) alongside LLM-based methods that offer paradigm flexibility in summarization; (2) presenting an in-depth analysis of \textbf{LLM-based Summarization Methods}, highlighting the latest works and research directions, including how in-context learning, prompt engineering, and few-shot learning have reshaped the ATS field; and (3) proposing a \textbf{Novel Retrieval Algorithm} that leverages LLMs to efficiently search for and organize relevant research papers, with potential applications beyond ATS.

For future directions, LLMs present significant advantages in terms of flexibility, generative quality, and reduced dependence on large labeled datasets. However, they also pose challenges such as ``hallucination'' in generated content, limitations in domain-specific summarization, efficiency concerns, and a lack of explainability. Accordingly, future research may focus on (1) optimizing prompt design to enhance LLM performance across a variety of ATS tasks; (2) developing more robust fine-tuning techniques to improve LLMs' ability to handle specialized domains, such as medical or legal texts; and (3) addressing issues related to consistency, factual accuracy, and interpretability in LLM-generated summaries through improved evaluation methods. Tackling these challenges will help solidify LLMs' pivotal role in advancing ATS.


\appendix
\section{Annotation Questionnaire Design}
\label{app:1}
\subsection{A.1 \texttt{is\_LLM} Classification Task}
\textbf{Question:} Determine whether the paper is related to Large Language Models (LLMs) based on its title, abstract, and main content. If ``Yes'', provide a brief rationale; if ``No'', explain the unrelated domain.  

\textbf{Judgment Criteria:}  
\begin{enumerate}
    \item[0.] \textbf{Yes:} Core content involves LLM architecture design, training methods, performance optimization, or application scenarios (e.g., GPT, BERT, Transformer).
    \item \textbf{No:} Focuses on traditional machine learning, computer vision, small-scale NLP models (e.g., RNN, CNN), or non-model-related fields (e.g., theoretical analysis, hardware acceleration).
\end{enumerate}

\textbf{Examples:}  
\begin{itemize}
    \item Yes: The paper proposes an improved LLaMA-3 architecture.
    \item No: The paper studies traditional SVM applications in ATS.
\end{itemize}

\subsection{A.2 \texttt{is\_dataset} Classification Task}
\textbf{Question:} Judge whether the paper is related to dataset construction. If ``Yes'', describe the dataset's purpose and characteristics; if ``No'', explain the reason. 

\textbf{Judgment Criteria:}  
\begin{enumerate}
    \item[0.] \textbf{Yes:} The paper explicitly constructs and releases a new dataset or details the extension/improvement of an existing dataset (e.g., annotation methods, data sources).
    \item \textbf{No:} The paper only uses publicly available datasets without involving dataset design or improvement.
\end{enumerate}

\textbf{Examples:}  
- Yes: The paper introduces the VSoLSCSum dataset with annotation tools and benchmarks. 
- No: The paper employs the Multi-News dataset for training but does not discuss dataset design. 

\subsection{A.3 \texttt{Type} Classification Task}
\textbf{Question:} Select the paper's task type from the following categories and provide justification: 
\begin{enumerate}
    \item[0.] \textbf{Evaluation:} Systematic comparison of model/method performance.
    \item \textbf{Survey:} Survey of summarization research progress, challenges, or applications.
    \item \textbf{Model:} Proposes a novel architecture or improves an existing model.
    \item \textbf{Others:} Other types, please indicate the specific category.
\end{enumerate}

\textbf{Examples:}  
\begin{itemize}
    \item Evaluation: The paper evaluates GPT-3's summarization performance via human preference judgments.
    \item Model: The paper proposes a sparse attention mechanism for LLMs.
\end{itemize}

\subsection{A.4 \texttt{Methodology} Classification Task}
\textbf{Question:} Select the technique(s) involved in the paper (multiple choices allowed) and briefly describe their application:  
\begin{enumerate}
    \item[0.] \textbf{Prompt Engineering:} Use prompts to guide model outputs.
    \item \textbf{Fine-tuning:} Train pre-trained models on specific tasks.
    \item \textbf{Distillation:} Compress large models into smaller ones via knowledge distillation.
    \item \textbf{Others:} Specify the method.
\end{enumerate}

\textbf{Examples:}  
\begin{itemize}
    \item Prompt Engineering: The paper designs a prompt template to enhance model generalization.
    \item Distillation: The paper distills GPT-3's knowledge into a smaller T5 model.
\end{itemize}

\subsection{A.5 \texttt{Domain} Classification Task}
\textbf{Question:} Classify the paper's domain, selecting from the following categories: News, Novel, Scientific Paper, Blog, Dialogue, Medical, or Other. Provide a rationale based on the title, abstract, and main content.  

\textbf{Classification Criteria:}  
\begin{enumerate}
    \item[0.] \textbf{News:} Summarization of journalistic articles or current events.
    \item \textbf{Novel:} Summarization of fictional literary works.
    \item \textbf{Scientific Paper:} Summarization of academic research articles.
    \item \textbf{Blog:} Summarization of personal or professional blog posts.
    \item \textbf{Dialogue:} Summarization of conversation-based texts (e.g., interviews, meetings).
    \item \textbf{Medical:} Summarization of clinical notes, medical reports, or health-related content.
    \item \textbf{Other:} Specify the domain if not listed above.
\end{enumerate}

\textbf{Examples:}  
\begin{itemize}
    \item \textbf{Scientific Paper:} The paper presents a model for arxiv papers summarization.
    \item \textbf{Medical:} The study proposes a model for condensing discharge summaries in electronic health records.
\end{itemize}

\subsection{Annotation Guidelines}
\begin{itemize}
    \item Base judgments on the paper's title and abstract.  
    \item Document controversies in the ``Remarks'' section if ambiguities arise.  
    \item Preserve original rationales (e.g., cite key sentences from the abstract).  
\end{itemize}

\footnotesize
\bibliography{IEEEabrv,mybibfile_clean}
\bibliographystyle{elsarticle-num-names}


\section{Author Biographies}

\noindent\textbf{Yang Zhang}\\[0.5em]
\begin{center}
  \includegraphics[width=1in,height=1.25in,clip,keepaspectratio]{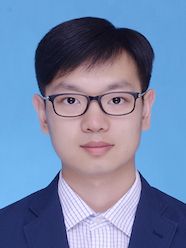}
\end{center}
Yang Zhang is an assistant professor with the Southwestern University of Finance and Economics, China. He received his Ph.D. from the Graduate School of Informatics, Kyoto University in 2022. Before that, he received his Bachelor of Engineering from the University of New South Wales and his Master of Economics and Finance from Sydney University. He joined Southwestern University of Finance and Economics, China in 2023 as a lecturer. His research interests include text mining, text recommendation, and NLP for financial technology.

\vspace{1em}

\noindent\textbf{Hanlei Jin}\\[0.5em]
\begin{center}
  \includegraphics[height=1.25in,clip,keepaspectratio]{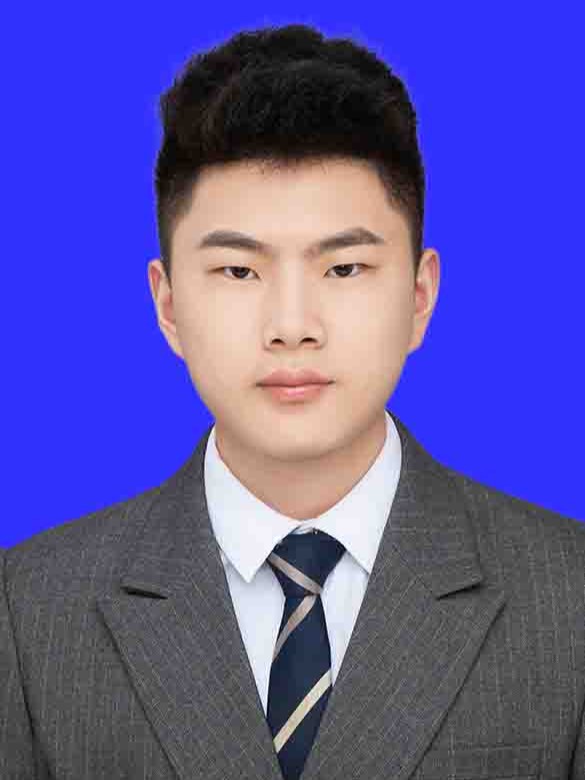}
\end{center}
Hanlei Jin is currently working toward the Ph.D. in Management Science and Engineering at Southwestern University of Finance and Economics, Chengdu, China. He holds a B.S. degree in Management from the same institution. His research interests include text mining, text generation, and financial intelligence.

\vspace{1em}

\noindent\textbf{Dan Meng}\\[0.5em]
\begin{center}
  \includegraphics[width=1in,height=1.25in,clip,keepaspectratio]{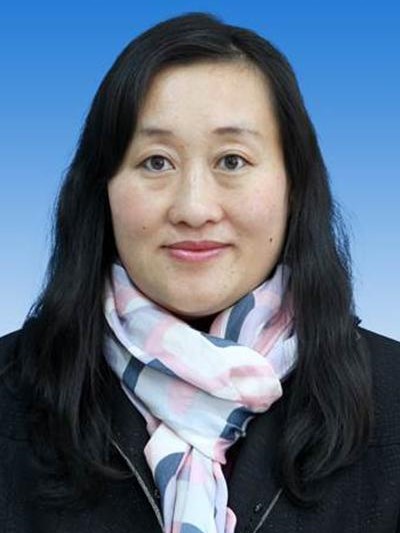}
\end{center}
Dan Meng is a professor with the Southwestern University of Finance and Economics, Chengdu, China. Previously, she received her Ph.D. from Southwest Jiaotong University, Chengdu, China. Her research interests include intelligent finance, intelligent decision making, and uncertainty information processing.

\vspace{1em}

\noindent\textbf{Jun Wang}\\[0.5em]
\begin{center}
  \includegraphics[width=1in,height=1.25in,clip,keepaspectratio]{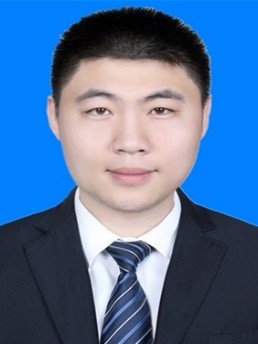}
\end{center}
Jun Wang is a professor with the Southwestern University of Finance and Economics, China. Prior to his current appointment, he was a researcher at the Memorial University of Newfoundland in St. John's, Canada, and he was awarded the National Scholarship in 2017. His research interests include NLP, social media, social network analysis, financial analysis, and business intelligence.

\vspace{1em}

\noindent\textbf{Jinghua Tan}\\[0.5em]
\begin{center}
  \includegraphics[width=1in,height=1.25in,clip,keepaspectratio]{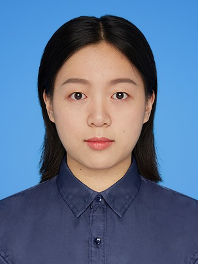}
\end{center}
(Member, IEEE) Jinghua Tan received the B.S., M.S., and Ph.D. degrees from the Southwestern University of Finance and Economics, Chengdu, China. She is an associate professor with Sichuan Agricultural University and was a visiting scholar at the Memorial University of Newfoundland in St. John's, Canada in 2019. Her research interests include data mining and financial intelligence.

\end{document}